\documentclass[lettersize,journal]{IEEEtran}
\usepackage{amsmath,amsfonts}
\usepackage{algorithmic}
\usepackage{algorithm}
\usepackage{array}
\usepackage[caption=false,font=normalsize,labelfont=sf,textfont=sf]{subfig}
\usepackage{textcomp}
\usepackage{stfloats}
\usepackage{url}
\usepackage{verbatim}
\usepackage{graphicx}
\usepackage{cite}
\usepackage[colorlinks=True]{hyperref}
\usepackage{upgreek}
\usepackage{wasysym}
\usepackage{xcolor}
\usepackage{multirow}
\usepackage{multicol}
\usepackage{makecell}
\usepackage{booktabs}
\usepackage{amssymb}
\usepackage{enumitem}

\usepackage{pifont}
\usepackage{amsthm}
\newtheorem{myRmk}{Remark}
\newtheorem{myDef}{Definition}

\newtheorem{myLem}{Lemma}
\begin{document}

\title{Adaptive 3D Convolution for Remote Sensing Image Fusion}

\author{Siran Peng, 
	    Xiangyu Zhu,~\IEEEmembership{Senior Member,~IEEE,}
	    Shang-Qi Deng,
	    Liang-Jian Deng,~\IEEEmembership{Senior Member,~IEEE,}
	    and Zhen Lei,~\IEEEmembership{Fellow,~IEEE}
\thanks{Siran Peng and Xiangyu Zhu are with the State Key Laboratory of Multimodal Artificial Intelligence Systems, Institute of Automation, Chinese Academy of Sciences (CASIA), Beijing 100190, China; the School of Artificial Intelligence, University of Chinese Academy of Sciences (UCAS), Beijing 100049, China. (Emails: pengsiran2023@ia.ac.cn, xiangyu.zhu@ia.ac.cn)}
\thanks{Shangqi Deng is with the State Key Laboratory of Human-Machine Hybrid Augmented Intelligence, and Institute of Artificial Intelligence and Robotics, Xi'an Jiaotong University, Xi'an 710049, No.28 Xianning West Road, China. (Email: shangqideng@stu.xjtu.edu.cn)}
\thanks{Liang-Jian Deng is with the School of Mathematical Sciences/Multi-Hazard Early Warning Key Laboratory of Sichuan Province, University of Electronic Science and Technology of China (UESTC), Chengdu, Sichuan 611731, China. (Emails: liangjian.deng@uestc.edu.cn)}
\thanks{Zhen Lei is with the State Key Laboratory of Multimodal Artificial Intelligence Systems, Institute of Automation, Chinese Academy of Sciences (CASIA), Beijing 100190, China; the School of Artificial Intelligence, University of Chinese Academy of Sciences (UCAS), Beijing 100049, China; the Centre for Artificial Intelligence and Robotics, Hong Kong Institute of Science and Innovation, Chinese Academy of Sciences, Hong Kong, China. (Email: zhen.lei@ia.ac.cn)}
\thanks{Corresponding author: Liang-Jian Deng.}
}



\maketitle

\begingroup
\renewcommand\thefootnote{}
\footnotetext{
Accepted by IEEE Transactions on Image Processing (TIP), 2026.
DOI: 10.1109/TIP.2026.3689418.
}
\endgroup

\begin{abstract}
Remote sensing image fusion aims to create a high-resolution multi/hyper-spectral image from a high-resolution image with limited spectral information and a low-resolution image with abundant spectral data. Recently, deep learning (DL) techniques have shown significant effectiveness in this area. Most DL-based methods approach image fusion as a 2D problem by encoding spectral information into feature map channels. However, our research suggests that this strategy introduces notable spectral distortions. In contrast, some methods consider spectral data as an additional dimension, utilizing standard 3D convolutions to preserve spectral information. Nevertheless, in a standard 3D convolutional layer, the same set of kernels is applied across all input regions, which we have found to be sub-optimal for image fusion. Furthermore, standard 3D convolutions necessitate substantial computational resources. To address these challenges, we propose a novel convolutional paradigm called Adaptive 3D Convolution (Ada3D) for remote sensing image fusion. Ada3D applies a unique set of 3D kernels to each input voxel, enabling the capture of fine-grained details. These adaptive kernels are generated through a two-step process: (\emph{i}) spatial and spectral kernels are derived from their respective image sources; (\emph{ii}) these two types of kernels are then combined to form content-aware 3D kernels that effectively integrate spatial and spectral information. Additionally, adaptive biases are introduced to enhance the convolutional outcome at the voxel level. Furthermore, we incorporate the group convolution technique to reduce computational complexity. As a result, Ada3D offers full adaptivity in an efficient manner. Evaluation results across five datasets demonstrate that our method achieves state-of-the-art (SOTA) performance, underscoring the superiority of Ada3D. The code is available at {\url{https://github.com/PSRben/Ada3D}}.

\end{abstract}


\begin{figure}[t]
	\begin{center}
		\begin{minipage}{1\linewidth}
			{\includegraphics[width=0.9\linewidth]{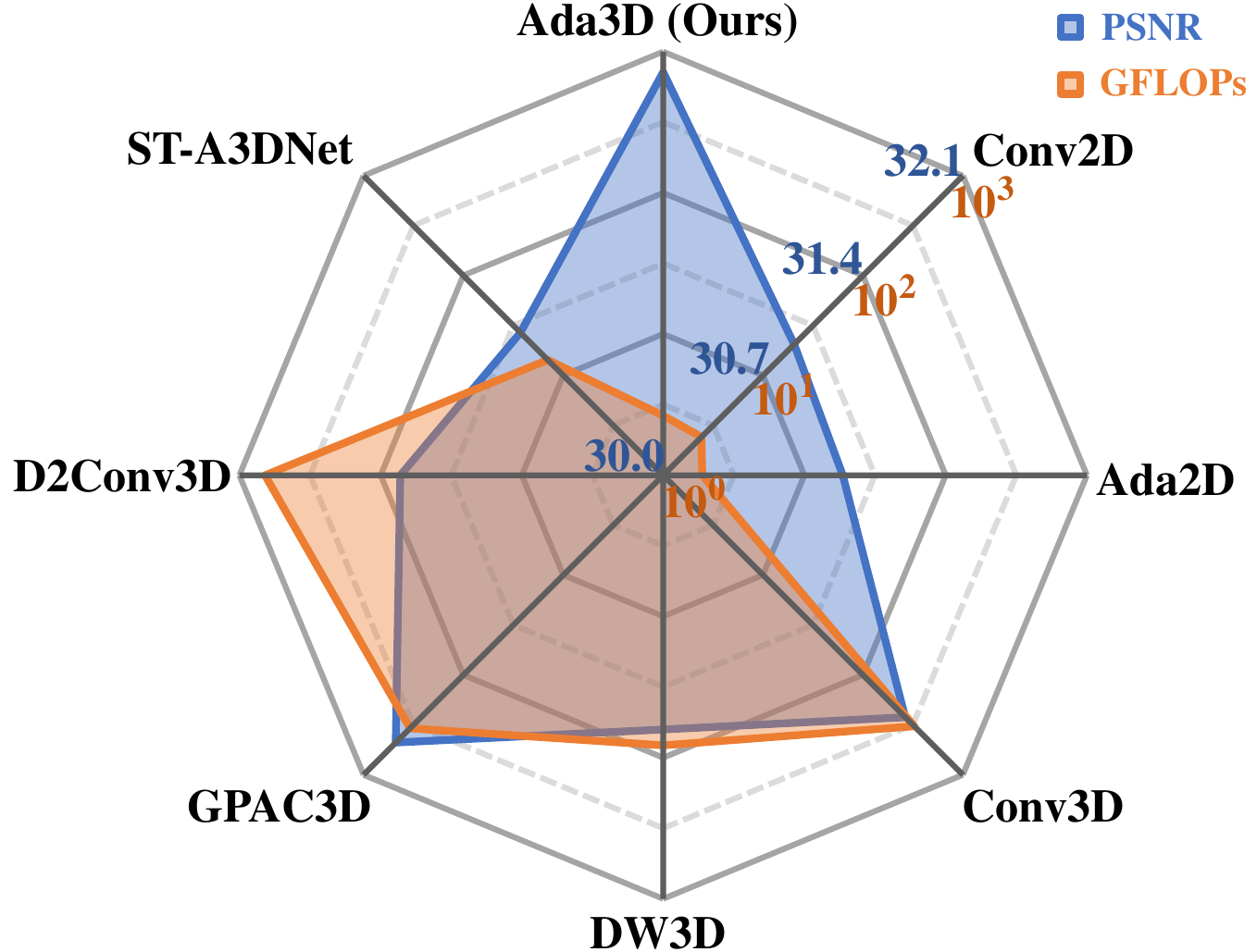}}
			\centering
		\end{minipage}
	\end{center}
    \vspace{-3pt}
	\caption{
	Comparison of Ada3D with several convolutional paradigms, including standard 2D convolution (Conv2D), adaptive 2D convolution (Ada2D), standard 3D convolution (Conv3D), depth-wise 3D convolution (DW3D), grouped 3D PAC (GPAC3D) \cite{su2019pixeladaptive}, D2Conv3D \cite{Schmidt_2022_WACV}, and ST-A3DNet \cite{10.1145/3510829}. For fairness, all methods have an equal number of parameters. Quantitative evaluation outcomes on the WDC dataset demonstrates the superior effectiveness and efficiency of Ada3D. For detailed results, please refer to Table~\ref{abl5}. \label{titlefigure} 
}
\end{figure}

\section{Introduction}
Satellite sensors often face challenges in obtaining high-resolution multi/hyper-spectral images due to hardware constraints. To overcome this, they can simultaneously acquire a high-resolution image with limited spectral information alongside a low-resolution image rich in spectral data. The objective of remote sensing image fusion is to integrate these two types of images, producing a high-resolution output enriched with spectral details. This study focuses on two remote sensing image fusion applications: pansharpening \cite{9245579} and hyper-spectral pansharpening \cite{7284770}. Pansharpening combines a high-resolution panchromatic (PAN) image with a low-resolution multi-spectral (LRMS) image to create a high-resolution multi-spectral (HRMS) result. Hyper-spectral pansharpening extends this technique to hyper-spectral images, generating a high-resolution hyper-spectral (HRHS) image from a PAN image and a low-resolution hyper-spectral (LRHS) image.

Traditional pansharpening and hyper-spectral pansharpening studies can be generally divided into three main categories: component substitution (CS) methods \cite{aiazzi2006mtf, 2010A, 2019Robust}, multi-resolution analysis (MRA) approaches \cite{Liu2000Smoothing, 6616569, vivone2018full}, and variational optimization (VO) techniques \cite{palsson2013new, zhang2015pan, 8444767, 9501252, 9722977}. CS-based methods project the LRMS/LRHS image into a transformed domain, where spatial information is considered as an independent component. By replacing this component with the PAN image, they produce the desired HRMS/HRHS output. While recognized for their simplicity, low computational demands, and high spatial fidelity, these methods introduce significant spectral distortions. MRA-based approaches leverage an MRA framework to incorporate spatial details from the PAN image into the LRMS/LRHS image, creating an HRMS/HRHS result. These approaches excel in preserving spectral characteristics but may experience notable spatial distortions. In contrast, VO-based techniques focus on revealing the intrinsic relationships between two distinct image types. They typically utilize various forms of prior information to develop optimization models that merge spatial and spectral data. Despite careful design, VO-based techniques often struggle to produce satisfactory fusion results.

In recent years, deep learning (DL) has emerged as the premier solution for tackling image fusion challenges in the remote sensing field \cite{2016Pansharpening, 8237455, 8731649, 2020Deep, 2020Detail, 9761261, 10137388, 10142023, 10.1145/3581783.3612084, 10243544, 10415854}. 
Inspired by advancements in other low-level vision tasks \cite{dong2014learning, 9133311, 9913829, 10387229}, most DL-based methods approach image fusion as a 2D problem, encoding spectral information into feature map channels. However, as we show mathematically in Section \ref{s32}, this can lead to spectral distortions. Alternatively, some methods approach image fusion as a 3D problem, dedicating an additional dimension to spectral data. Due to the limited variety of foundational DL models in the 3D domain, these methods typically rely on standard 3D convolutions for feature learning. In a standard convolutional layer, the same set of kernels is applied across different regions of various inputs, making the process \emph{content-agnostic} \cite{su2019pixeladaptive}. Section \ref{s33} demonstrates that this is sub-optimal for image fusion. Moreover, the increased dimensionality of 3D convolutions substantially raises computational complexity.

Adaptive convolution marks a significant advancement in overcoming the content-agnostic limitations of standard convolution \cite{2016Dynamic,2019Adaptive,2021Decoupled}. By utilizing distinct kernel sets for different input regions, adaptive convolution enhances the extraction of detailed information. Recently, this technique has been applied to remote sensing image fusion, demonstrating considerable promise \cite{jin2022aaai,ijcai2022p179,Duan_2024_CVPR}. However, its development has largely been confined to the 2D domain, with limited exploration in 3D. Existing approaches like D2Conv3D \cite{Schmidt_2022_WACV} and ST-A3DNet \cite{10.1145/3510829} still depend on globally shared kernels, preventing them from being considered fully adaptive 3D convolution techniques.

The aforementioned situation motivates us to propose Adaptive 3D Convolution (Ada3D), an innovative convolutional paradigm for remote sensing image fusion. Ada3D assigns a unique set of 3D kernels to each input voxel, allowing for the capture of fine-grained details. These adaptive kernels are generated through an intuitive two-step process: (\emph{i}) spatial kernels are derived from the PAN image, while spectral kernels are extracted from the LRMS/LRHS image; (\emph{ii}) these two types of kernels are then combined to form content-aware 3D kernels, effectively integrating both spatial and spectral information. In parallel, adaptive biases are computed using a similar strategy to enhance the convolutional outcome on a voxel-by-voxel basis. Additionally, we leverage group convolution to reduce computational complexity. Our experimental results show that Ada3D outperforms other 3D convolutional paradigms while maintaining FLOPs comparable to 2D methods, as depicted in Fig.~\ref{titlefigure}. Compared to existing approaches that are not fully adaptive, Ada3D establishes a new and efficient paradigm for 3D convolution, achieving voxel-level adaptivity by dynamically generating unique kernels and biases for each location. In conclusion, the \emph{\textbf{contributions}} of this paper are as follows:

\begin{enumerate}
	\item We mathematically illustrate that treating image fusion purely as a 2D problem can result in substantial spectral distortions (Section \ref{s32}). Furthermore, we prove that the kernel-sharing approach in standard convolution is sub-optimal for image fusion applications (Section \ref{s33}).
	\item We introduce Ada3D, a novel 3D convolutional paradigm for remote sensing image fusion. In Ada3D, each voxel is processed with a unique set of adaptive kernels. These kernels are derived from two image sources, effectively integrating both spatial and spectral information. We also generate adaptive biases to enhance the convolutional outcome on a voxel-wise basis. Additionally, the group convolution technique is employed to reduce complexity.
	\item We evaluate our method on five datasets, where Ada3D achieves state-of-the-art (SOTA) performance in qualitative and quantitative assessments. Notably, with the same number of parameters, Ada3D surpasses other 3D convolution paradigms in both effectiveness and efficiency.
\end{enumerate}

\section{Related Works \& Motivations}
\label{s2}
\subsection{DL Methods for Remote Sensing Image Fusion}
Over recent years, DL-based methods have gained dominance in various remote sensing image fusion applications. By leveraging the advantages of neural networks, these methods significantly outperform traditional approaches. Most DL-based studies treat image fusion as a 2D problem, encoding spectral information within feature map channels. Noteworthy contributions to this area include PNN \cite{2016Pansharpening}, PanNet \cite{8237455}, and U2Net \cite{10.1145/3581783.3612084}. PNN, in particular, stands out as a groundbreaking accomplishment for introducing DL into remote sensing image fusion. Utilizing three standard 2D convolutional layers, PNN attains SOTA performance at the time of its publication. PanNet applies high-pass filters to extract edge information and employs residual network blocks (ResBlocks) \cite{He_2016_CVPR} to merge spatial and spectral features. U2Net, on the other hand, integrates modified cross-attention modules within two U-shaped networks, achieving exceptional performance in the pansharpening task. However, these methods often introduce spectral distortions as they alter the representation of spectral information. To tackle this issue, some studies treat image fusion as a 3D problem, incorporating an additional dimension for spectral data. This strategy draws inspiration from successes in the wider domain of remote sensing image processing, where 3D convolutions have demonstrated remarkable effectiveness \cite{firat2023multiscale,firat2023hybrid,firat20233d,li2024dbanet,10398878}. Building on this foundation, notable recent advancements in image fusion include 3DSSR-Net \cite{rs14174250} and MSAC-Net \cite{zhang2022msac}. The former utilizes stacked standard 3D convolutional layers to extract features while preserving spectral integrity. The latter proposes an innovative gating mechanism to improve the performance of standard 3D convolutions. Due to their increased dimensionality, these methods are burdened by a significant volume of floating-point operations (FLOPs), which hampers their broader adoption.

\begin{figure*}[t]
	\begin{center}
		\begin{minipage}{1\linewidth}
			{\includegraphics[width=0.96\linewidth]{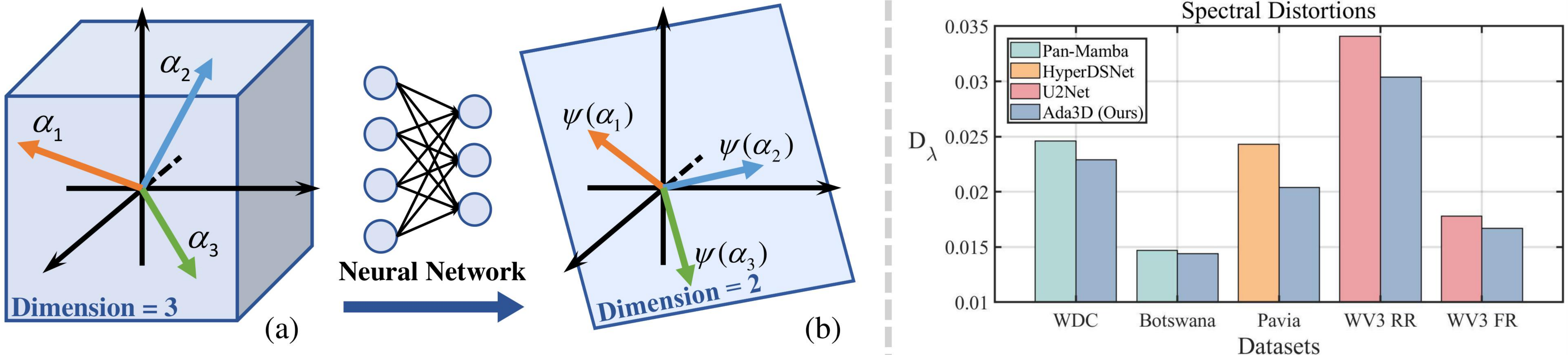}}
			\centering
		\end{minipage}
	\end{center}
	\caption{{\textbf{Left}}: A visual representation of how 2D modeling introduces spectral distortions. (a) The vector space of spectral data, where the number of spectral bands $L$ is set to 3 for simplicity. (b) The vector space of spectral information, which is encoded into feature map channels via a convolutional layer. This scenario illustrates the case where ${\rm{rank}}(\boldsymbol{A})=2<L$, resulting in a loss of spectral information. \textbf{Right}: A comparison of spectral distortions between Ada3D and SOTA 2D methods for image fusion, including Pan-Mamba \cite{he2024pan}, HyperDSNet \cite{9870551}, and U2Net \cite{10.1145/3581783.3612084}. Spectral distortions are quantified utilizing the ${{\rm{D}}_{\lambda}}$ \cite{6998089} index, with lower ${{\rm{D}}_{\lambda}}$ values indicating fewer spectral distortions. Notably, Ada3D exhibits the lowest spectral distortions across all tested datasets.\label{2d3d}}
\end{figure*}

\subsection{Adaptive Convolution}
Standard convolution employs the same set of kernels across all regions of different inputs, leading to sub-optimal feature extraction and reduced flexibility \cite{su2019pixeladaptive}. In contrast, adaptive convolution assigns unique sets of kernels to each area of every input, allowing for more effective and dynamic feature extraction. The pioneering work in this field is DFN \cite{2016Dynamic}, which creates a separate network branch to generate adaptive 2D kernels from image input. Building on this, LS-DFN \cite{2018Dynamic} leverages the attention mechanism to produce kernels with enhanced content information. Later achievements, such as PAC \cite{su2019pixeladaptive} and DDF \cite{2021Decoupled}, further highlight the promise of this novel convolutional paradigm. Adaptive convolution has recently been applied to remote sensing image fusion, with notable contributions such as LAGNet \cite{jin2022aaai} and ADKNet \cite{ijcai2022p179}. The former introduces an innovative bias production technique and utilizes stacked adaptive convolutional layers to effectively merge spatial and spectral features. The latter proposes a two-branch kernel generator for information integration, demonstrating outstanding generalization capabilities. However, the methods discussed above are confined to 2D space, with only a few efforts made to explore the potential of adaptive convolution in 3D contexts. Existing approaches, such as D2Conv3D \cite{Schmidt_2022_WACV} and ST-A3DNet \cite{10.1145/3510829}, continue to rely on globally shared kernels, which prevents them from being classified as genuine adaptive 3D convolution methods. 

\subsection{Motivations}
Most DL-based methods frame image fusion as a 2D vision problem, encoding spectral information within feature map channels. However, our mathematical analysis reveals that this approach can cause significant spectral distortions. Alternatively, some methods introduce an extra dimension for spectral data, treating image fusion as a 3D problem. These methods typically use standard 3D convolutions for feature learning, which we have found to be sub-optimal for image fusion applications. Moreover, standard 3D convolutions require substantial computational resources. By assigning unique sets of kernels to various input regions, adaptive convolution offers a promising alternative to standard convolution. 
However, its development has predominantly been confined to 2D space, with limited exploration in 3D contexts. Therefore, we propose Ada3D, an innovative convolutional paradigm for remote sensing image fusion. Ada3D allocates a unique set of 3D kernels to each input voxel. These adaptive kernels consist of spatial and spectral components derived from the PAN and LRMS/LRHS images, facilitating the integration of different information. Additionally, we leverage the group convolution technique to greatly reduce complexity, bringing Ada3D’s FLOPs to a level comparable to that of 2D convolution.

\section{Mathematical Analysis}
\label{s3}
In this section, we first introduce some basic theories from linear algebra. Leveraging these theories, we mathematically illustrate the limitations of 2D modeling and standard convolution in remote sensing image fusion. This analysis serves as the theoretical foundation for the development of Ada3D. Throughout the rest of this paper, indices and sizes are denoted by $i$ and $H$, vectors by $\boldsymbol{x}$, matrices by $\boldsymbol{X}$, and tensors by $\mathcal{X}$.

\subsection{Basic Theories}
\label{s31}
\begin{myDef}
\label{d1}
Given an $M$th-order tensor $\mathcal{X}\in\mathbb{R}^{I_1\times I_2\times \cdots \times I_M}$, its mode-$k$ unfolding can be represented by a matrix $\boldsymbol{X}\in\mathbb{R}^{I_k\times(I_1I_2\cdots I_{k-1}I_{k+1}\cdots I_{M})}$. In this unfolding, the element at position $(i_1, i_2, \cdots, i_M)$ in $\mathcal{X}$ is mapped to the element at position $(i_k, j)$ in $\boldsymbol{X}$, where the index $j$ is computed as:
\begin{equation}
j=1+\sum\limits_{\substack{n=1\\n\neq k}}^{M}(i_n-1)\prod\limits_{\substack{m=1\\m\neq k}}^{n-1}I_m.
\end{equation}
\end{myDef}

\begin{myDef}	
\label{d2} 
Let $\varphi$ be a linear map from an $n$-dimensional vector space $V$ to an $m$-dimensional vector space $U$. Suppose $(\boldsymbol{\alpha}_1, \boldsymbol{\alpha}_2, \cdots, \boldsymbol{\alpha}_n)$ is a basis for $V$ and $(\boldsymbol{\beta}_1, \boldsymbol{\beta}_2, \cdots, \boldsymbol{\beta}_m)$ is a basis for $U$. For each basis vector $\boldsymbol{\alpha}_i$ in $V$, its image under $\varphi$, denoted by $\varphi(\boldsymbol{\alpha}_i)$, can be written as a unique linear combination of the basis vectors in $U$. Explicitly, we have:
\begin{equation}
\label{de1}
\footnotesize
\setlength{\arraycolsep}{0.5pt}
(\varphi(\boldsymbol{\alpha}_1),\varphi(\boldsymbol{\alpha}_2),\cdots,\varphi(\boldsymbol{\alpha}_n)) = (\boldsymbol{\beta}_1, \boldsymbol{\beta}_2, \cdots, \boldsymbol{\beta}_m)
\begin{pmatrix}
	a_{11} & a_{12} & \cdots & a_{1n} \\ 
	a_{21} & a_{22} & \cdots & a_{2n} \\ 
	\vdots & \vdots &        & \vdots \\ 
	a_{m1} & a_{m2} & \cdots & a_{mn}
\end{pmatrix}.
\end{equation}
Here, we define the $m\times n$ matrix on the right as $\boldsymbol{A}$, which represents the linear map $\varphi$ with respect to two given bases. Let $\boldsymbol{\lambda}={(\lambda_1,\lambda_2,\cdots,\lambda_n)}^\intercal$ be a coordinate vector based on $(\boldsymbol{\alpha}_1, \boldsymbol{\alpha}_2, \cdots, \boldsymbol{\alpha}_n)$. The corresponding coordinate vector $\boldsymbol{\mu}={(\mu_1,\mu_2,\cdots,\mu_m)}^\intercal$, expressed in terms of $(\boldsymbol{\beta}_1, \boldsymbol{\beta}_2, \cdots, \boldsymbol{\beta}_m)$ after applying the linear map, is given by:
\begin{equation}
	\label{de2}
	\boldsymbol{\mu} = \boldsymbol{A}\boldsymbol{\lambda}.
\end{equation}
\end{myDef}

\begin{myDef}
	\label{d3}
	Let $\boldsymbol{A}$ be an $m\times n$ matrix. The rank of $\boldsymbol{A}$, denoted as ${\rm{rank}}(\boldsymbol{A})$, is defined as the maximum number of linearly independent columns (or cows) in the matrix. Thus, ${\rm{rank}}(\boldsymbol{A})\leq {\rm{min}}(m, n)$. If $U$ is the vector space spanned by the column vectors of $\boldsymbol{A}$, then the dimension of $U$ is ${\rm{rank}}(\boldsymbol{A})$.
\end{myDef}

\begin{myDef}
\label{d4} 
Let $\boldsymbol{A}$ be an $m\times n$ coefficient matrix, $\boldsymbol{x}$ be an $n$-dimensional vector of variables, and $\boldsymbol{b}$ be an $m$-dimensional vector of constants. Then, the non-homogeneous system of linear equations can be expressed in the form $\boldsymbol{A}\boldsymbol{x}=\boldsymbol{b}$. Let $\boldsymbol{\overline{A}}=(\boldsymbol{A}|\boldsymbol{b})$ be the augmented matrix. The number of solutions $N$ under different conditions is represented as follows:
\begin{equation}
N=
\begin{cases}
	1, & {\rm{rank}}(\boldsymbol{A})={\rm{rank}}(\boldsymbol{\overline{A}})=n\\
	\infty, &{\rm{rank}}(\boldsymbol{A})={\rm{rank}}(\boldsymbol{\overline{A}})<n\\
	0, &{\rm{rank}}(\boldsymbol{A})\neq{\rm{rank}}(\boldsymbol{\overline{A}})
\end{cases}
.
\end{equation}
\end{myDef}

\subsection{Comparison between 2D \& 3D Modeling}
\label{s32}
Most DL-based methods treat image fusion as a 2D problem. They typically encode spectral information into feature map channels using a standard 2D convolutional layer. Let's denote the spectral data as $\mathcal{X}\in\mathbb{R}^{H\times W\times L}$ (with $H$, $W$, and $L$ representing the height, width, and number of spectral bands, respectively), the resulting feature map as $\mathcal{Y}\in\mathbb{R}^{H\times W\times C}$ (with $C$ indicating the number of channels in the feature map), and the convolutional layer as $f(\cdot)$. Then, the encoding process of spectral information can be described as follows:
\begin{equation}
	\label{pe1}
	\mathcal{Y}=f(\mathcal{X}).
\end{equation}
If the kernel size is $1\times 1$, we perform mode-$3$ unfolding on $\mathcal{X}$ and $\mathcal{Y}$ to express the function $f(\cdot)$ mathematically. This results in matrices $\boldsymbol{X}\in\mathbb{R}^{L\times(HW)}$ and $\boldsymbol{Y}\in\mathbb{R}^{C\times(HW)}$. Given that each spectral band corresponds to a distinct wavelength range, the columns of $\boldsymbol{X}$ can be considered as coordinate vectors with respect to a basis $(\boldsymbol{\alpha}_1,\boldsymbol{\alpha}_2,\cdots,\boldsymbol{\alpha}_L)$, where $\boldsymbol{\alpha}_i\in\mathbb{R}^{L}$. Additionally, the columns of $\boldsymbol{Y}$ can be viewed as coordinate vectors in terms of the standard basis $(\boldsymbol{e}_1,\boldsymbol{e}_2,\cdots,\boldsymbol{e}_C)$, where $\boldsymbol{e}_i\in\mathbb{R}^{C}$. Then, Equation \ref{pe1} can be rewritten as:
\begin{equation}
	\label{pe2}
	\boldsymbol{Y}=\boldsymbol{A}\boldsymbol{X},
\end{equation}
where $\boldsymbol{A}\in\mathbb{R}^{C\times L}$ is the weight matrix of $f(\cdot)$. According to Definition \ref{d2}, $\boldsymbol{A}$ can be interpreted as the matrix representation of the linear map $\psi$, which is the abstract form of the function $f(\cdot)$. $\psi$ maps spectral information into feature map channels with respect to $(\boldsymbol{\alpha}_1,\boldsymbol{\alpha}_2,\cdots,\boldsymbol{\alpha}_L)$ and $(\boldsymbol{e}_1,\boldsymbol{e}_2,\cdots,\boldsymbol{e}_C)$. Therefore, $\psi(\boldsymbol{\alpha}_{i})$ is precisely the $i$-th column of $\boldsymbol{A}$. As stated in Definition \ref{d3}, the dimension of the vector space spanned by $(\psi(\boldsymbol{\alpha}_{1}), \psi(\boldsymbol{\alpha}_{2}), \cdots, \psi(\boldsymbol{\alpha}_{L}))$ is equal to ${\rm{rank}}(\boldsymbol{A})$. When ${\rm{rank}}(\boldsymbol{A})$ is smaller than $L$, the dimension of the vector space spanned by $(\boldsymbol{\alpha}_1,\boldsymbol{\alpha}_2,\cdots,\boldsymbol{\alpha}_L)$, spectral information loss occurs. This happens because a lower-dimensional vector space possesses a reduced capacity to encapsulate the information contained in complex data. Additionally, a detailed explanation for the scenario where the kernel size is $k\times k$ is provided in the supplementary material. We now present the following remark.

\begin{myRmk}
\label{th1}
Approaching image fusion as a 2D problem may cause significant spectral distortions, particularly when the feature map has fewer channels than the spectral bands.
\end{myRmk}

In the 3D modeling of image fusion, spectral data is stored in an additional dimension, maximizing the representation capability of spectral information. Experimental outcomes, as shown on the right side of Fig.~\ref{2d3d}, indicate that 3D modeling consistently reduces spectral distortions compared to 2D modeling. Furthermore, the left side of the figure visually demonstrates how 2D modeling causes spectral distortions.

\begin{figure}[t]
	\begin{center}
		\begin{minipage}{1\linewidth}
			{\includegraphics[width=1\linewidth]{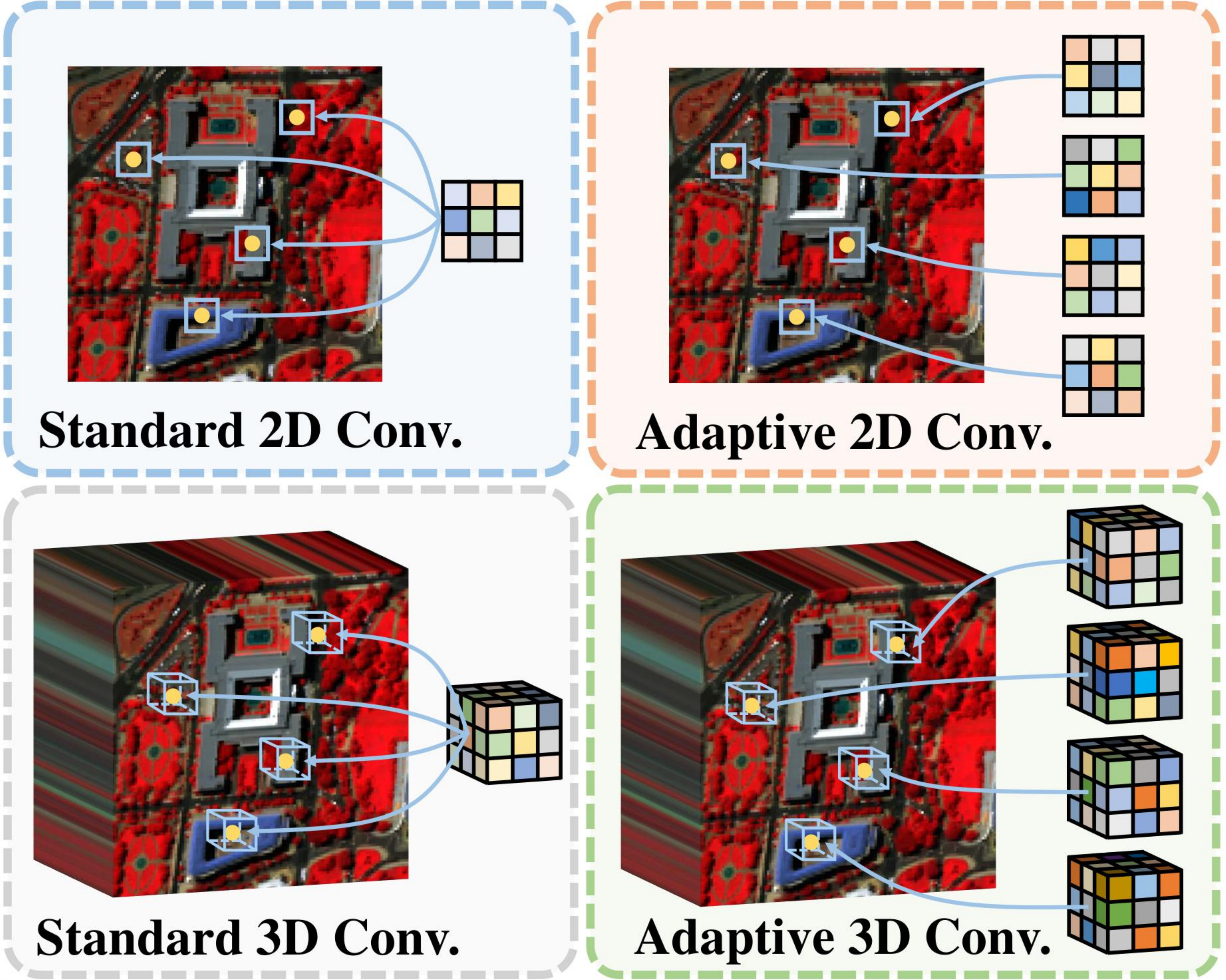}}
			\centering
		\end{minipage}
	\end{center}
	\caption{Graphical illustration of standard 2D convolution, adaptive 2D convolution, standard 3D convolution, and adaptive 3D convolution.\label{stdada}}
\end{figure}

\subsection{Comparison between Standard \& Adaptive Convolutions}
\label{s33}
Given that image fusion aims to reconstruct a high-resolution image enriched with spectral information, the convolutional output should accurately reflect the physical properties of the scene rather than being an abstract representation of the input data. Consequently, we assume that each convolutional layer has a specific, intended output that aligns with this goal. In this context, we examine a simplified 2D convolution model, where the input data is represented as $\boldsymbol{\tilde{A}}\in\mathbb{R}^{H\times W}$ and the target output as $\boldsymbol{B}\in\mathbb{R}^{H\times W}$. In a 2D convolutional layer with a kernel size of $k\times k$, the value of $\boldsymbol{B}{(i, j)}$ is calculated as the weighted sum of the elements within a $k\times k$ neighborhood centered at $\boldsymbol{\tilde{A}}{(i, j)}$. To mathematically express this operation, we begin by converting these neighborhoods into vectors. This conversion is achieved by stacking the elements of each neighborhood in a row-wise manner. The resulting vectors are then organized into an $H\times W\times k^2$ tensor, denoted as $\mathcal{A}$. Next, we perform mode-$3$ unfolding on $\mathcal{A}$ and transpose the resulting matrix, yielding $\boldsymbol{A}\in\mathbb{R}^{(HW)\times k^2}$. Additionally, we flatten $\boldsymbol{B}$ row-wise to obtain $\boldsymbol{b}\in\mathbb{R}^{HW}$. 

In a standard convolutional layer, the same set of kernels is applied across all regions of the input. For simplicity, we denote the kernel weights as a vector $\boldsymbol{x}\in\mathbb{R}^{k^2}$. Thus, the standard 2D convolution can be mathematically expressed as:
\begin{equation}
\label{pe3}
\boldsymbol{A}\boldsymbol{x}=\boldsymbol{b}.
\end{equation}
According to Definition \ref{d4}, Equation \ref{pe3} can be viewed as a non-homogeneous system, with $\boldsymbol{A}$ denoting the coefficient matrix, $\boldsymbol{x}$ the vector of variables, and $\boldsymbol{b}$ the vector of constants. Let $\boldsymbol{\overline{A}}=(\boldsymbol{A}|\boldsymbol{b})\in\mathbb{R}^{(HW)\times(k^2+1)}$ be the augmented matrix. When ${\rm{rank}}(\boldsymbol{A})\neq{\rm{rank}}(\boldsymbol{\overline{A}})$, the system has no solution, indicating that the standard convolution cannot yield the desired output.

In an adaptive convolutional layer, each input region is processed with a unique set of kernels. We denote the weights of these adaptive kernels as $\boldsymbol{X}\in\mathbb{R}^{(HW)\times k^2}$. The simplified adaptive convolution can be mathematically expressed as:
\begin{equation}
\label{pe4}
\boldsymbol{A}{(i, \cdot)}{\boldsymbol{X}{(i, \cdot)}}^\intercal=\boldsymbol{b}{(i)},\quad i=1, 2, \cdots, HW.
\end{equation}
If we consider $\boldsymbol{A}{(i, \cdot)}$ and $\boldsymbol{\overline{A}}{(i, \cdot)}$ as row matrices, we have ${\rm{rank}}(\boldsymbol{A}{(i, \cdot)})={\rm{rank}}(\boldsymbol{\overline{A}}{(i, \cdot)})\leq k^2$. This relationship suggests that solutions exist theoretically, indicating that the adaptive convolution can generate the desired output.

For comprehensive explanations of the remaining cases, please refer to the supplementary material. Additionally, Fig.~\ref{stdada} provides a graphical illustration of standard and adaptive convolutions in both 2D and 3D domains for better understanding. In conclusion, we present the following remark. 
\begin{myRmk}
\label{th2}
The kernel-sharing strategy of standard convolution limits its information processing capabilities compared to adaptive convolution, making it sub-optimal for image fusion.
\end{myRmk}

\begin{figure*}[t]
	\begin{center}
		\begin{minipage}{1\linewidth}
			{\includegraphics[width=0.94\linewidth]{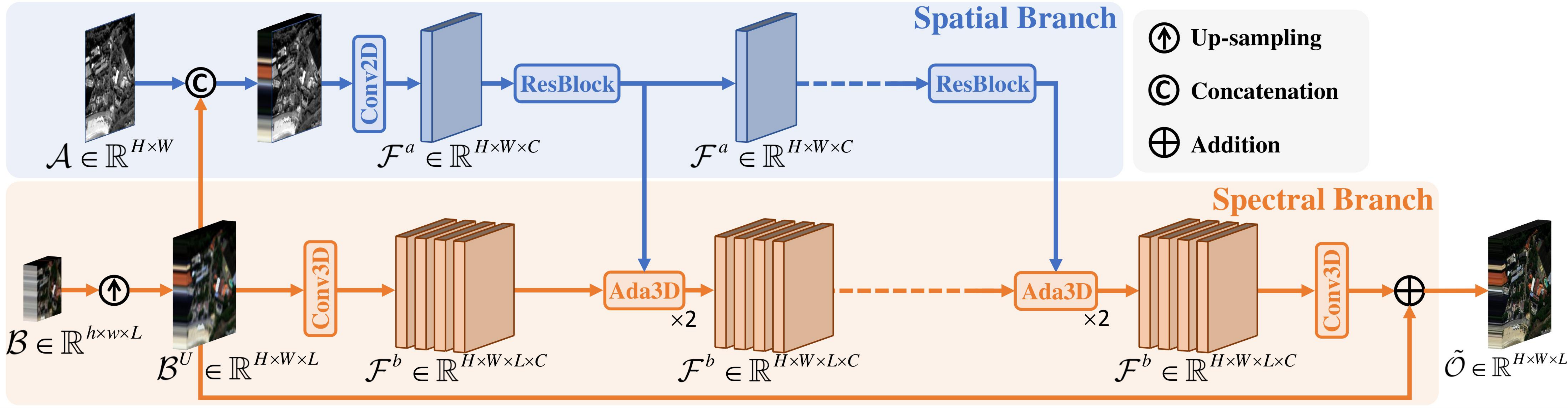}}
			\centering
		\end{minipage}
	\end{center}
    \vspace{-3pt}
	\caption{Graphical illustration of our network architecture, which consists of two main sections: a spatial branch and a spectral branch. The former focuses on extracting spatial details, while the latter captures spectral characteristics and progressively achieves information integration using Ada3D blocks. \label{pip}}
\end{figure*}

\section{Methodology}
\label{s4}
\subsection{Notations}
\label{s41}
The PAN image is represented as $\mathcal{A} \in \mathbb{R}^{H \times W}$, where $H$ and $W$ denote its height and width, respectively. Additionally, the LRMS/LRHS image is represented as $\mathcal{B} \in \mathbb{R}^{h \times w\times L}$, with $h=H/4$ and $w=W/4$, and $L$ indicating the number of spectral bands. Furthermore, the up-sampled LRMS/LRHS image, the generated HRMS/HRHS image, and the ground-truth (GT) image are denoted as ${\mathcal{B}}^U \in \mathbb{R}^{H \times W\times L}$, $\mathcal{\tilde{O}} \in \mathbb{R}^{H \times W\times L}$, and $\mathcal{{O}} \in \mathbb{R}^{H \times W\times L}$, respectively. In general, our network takes $\mathcal{A}$ and $\mathcal{B}$ as inputs to produce an output $\mathcal{\tilde{O}}$, which is supervised by $\mathcal{{O}}$. The network comprises two main sections: a spatial branch and a spectral branch. The former generates spatial feature maps denoted by ${\mathcal{F}}^{a} \in \mathbb{R}^{H \times W \times C}$, where $C$ represents the number of channels. In addition, the latter produces spectral feature maps denoted as ${\mathcal{F}}^{b} \in \mathbb{R}^{H \times W \times L \times C}$. Moreover, the adaptive 3D kernels have a size of $k \times k \times k$.

\subsection{Network Architecture}
\label{s42}
Since the primary focus of this study is not on the network architecture itself, we adopt a relatively simple design, as illustrated in Fig.~\ref{pip}, to validate the effectiveness of Ada3D. This architecture consists of two main sections: a spatial branch and a spectral branch. The former is responsible for extracting spatial details, while the latter captures spectral characteristics and progressively performs image fusion.

In the spatial branch, we start by concatenating $\mathcal{A}$ and $\mathcal{B}^U$ to integrate spatial information. Subsequently, a standard  $3 \times 3$ convolutional layer is applied to adjust the number of channels, producing the spatial feature map $\mathcal{F}^{a}$. Following that, we utilize a series of ResBlocks to extract spatial features. 

In the spectral branch, we first up-sample $\mathcal{B}$ using either Bicubic or PixelShuffle to enhance its spatial resolution, resulting in $\mathcal{B}^U$. Next, we apply a standard $3\times 3\times 3$ convolutional layer to generate a feature vector for each voxel in $\mathcal{B}^U$, producing $\mathcal{F}^{b}$. Subsequently, a series of Ada3D blocks are employed to effectively merge spectral and spatial information. We then utilize another standard $3\times 3\times 3$ convolutional layer to combine the feature vector values of each voxel. The result is added to $\mathcal{B}^{U}$, yielding the desired output, $\mathcal{\tilde{O}}$.

\subsection{Kernel Generator}
\begin{figure*}[t]
	\begin{center}
		\begin{minipage}{1\linewidth}
			{\includegraphics[width=0.94\linewidth]{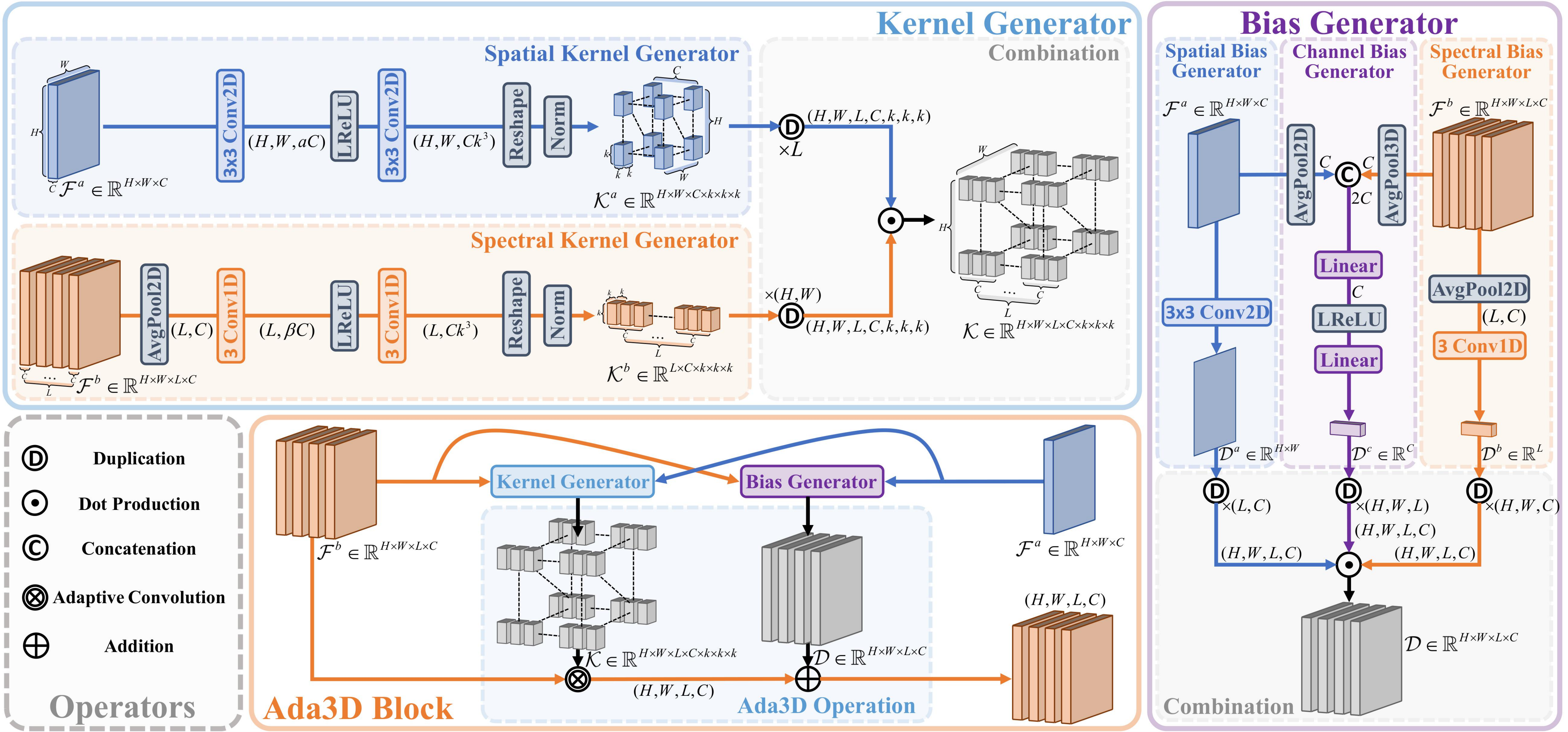}}
			\centering
		\end{minipage}
	\end{center}
    \vspace{-2pt}
	\caption{The Ada3D block. Taking $\mathcal{F}^a$ and $\mathcal{F}^b$ as inputs, the kernel and bias generators produce adaptive kernels $\mathcal{K}$ and biases $\mathcal{D}$. The acquired $\mathcal{K}$ and $\mathcal{D}$ are then applied to $\mathcal{F}^b$, facilitating the integration of spatial and spectral information. The 1D and 2D convolutional layers use kernel sizes of $3$ and $3\times 3$, both with a stride and padding of 1. Normalization is applied over each $k\times k\times k$ kernel field, and the number of hidden channels is determined by $\alpha$ and $\beta$.\label{ada3d}}
\end{figure*}

The kernel generator is meticulously designed for the efficient production of adaptive 3D kernels that integrate both spatial and spectral information. As illustrated in Fig.~\ref{ada3d}, it consists of three components: a spatial kernel generator, a spectral kernel generator, and a combination section. 

The spatial kernel generator focuses on creating spatial kernels from $\mathcal{A}$. To achieve this, we first employ two $3\times 3$ convolutional layers to extract spatial details and expand the channel number of $\mathcal{F}^{a}$, resulting in a feature map of size $H\times W\times (Ck^3)$. Notably, the channel count is $Ck^3$ rather than $C^2k^3$, indicating that we employ the group convolution technique with the group number set to $C$. Subsequently, this feature map is reshaped into a sixth-order tensor, which represents the spatial kernels, denoted as $\mathcal{K}^{a} \in \mathbb{R}^{H\times W\times C\times k\times k\times k}$. Essentially, $\mathcal{K}^{a}$ can be interpreted as $HW$ sets of 3D kernels, with each set corresponding to a unique spatial position.

The spectral kernel generator aims to produce spectral kernels from $\mathcal{B}$. To begin with, we apply 2D global average pooling to remove spatial information from $\mathcal{F}^b$, generating a matrix of size $L\times C$. Subsequently, we utilize two consecutive 1D convolutional layers with a kernel size of $3$ to capture spectral information and increase the channel number to $Ck^3$. The resulting matrix is then reshaped into a fifth-order tensor, denoted as $\mathcal{K}^{b} \in \mathbb{R}^{L\times C\times k\times k\times k}$, which serves as the spectral kernels. In essence, $\mathcal{K}^{b}$ can be interpreted as $L$ sets of 3D kernels, each associated with a distinct spectral band.

\begin{table*}[t]
	\centering\renewcommand\arraystretch{1.1}\setlength{\tabcolsep}{7pt}
	\belowrulesep=0pt\aboverulesep=0pt
	\caption{Comparison of parameters and FLOPs across various convolutional paradigms. The number of output channels is set to $C$.\label{flops}}
	\begin{tabular}{c|cccc}
		\toprule
		{Convolutions} & {Standard 3D} & {Depth-wise 3D} & {PAC3D} & {Ada3D (Ours)} \\  
		\midrule
		{Params} & $C^2k^3$ & $Ck^3+C^2$ & $27C^3k^3$ & $3(3\alpha +\beta)C^2(k^3+1)$ \\
		{FLOPs} & $2HWLC^2k^3$ & $2HWLCk^3+2HWLC^2$ & $54HWLC^3k^3$ & $3HWLCk^3+18\alpha HWC^2(k^3+1)+6\beta LC^2(k^3+1)$  \\
		{FLOPs / Params} & $2HWL$ & $2HWL$ & $2HWL$ & $\approx(HWL/C+6\alpha HW+2\beta L)/(3\alpha+\beta)$ \\	
		\bottomrule
	\end{tabular}
\end{table*}

In the combination section, we start by duplicating $\mathcal{K}^{a}$ and $\mathcal{K}^b$ to match the size of $H\times W\times L\times C\times k\times k\times k$. After that, we use a dot-product operation to sufficiently merge these two types of kernels, thereby creating adaptive 3D kernels represented as $\mathcal{K} \in \mathbb{R}^{H\times W\times L\times C\times k\times k\times k}$.

\subsection{Bias Generator}
The bias generator produces adaptive biases that enhance the convolutional output. As illustrated in Fig.~\ref{ada3d}, it consists of four key components: a spatial bias generator, a spectral bias generator, a channel bias generator, and a combination section. Taking $\mathcal{F}^a$ as input, the spatial bias generator creates spatial biases denoted as $\mathcal{D}^{a} \in \mathbb{R}^{H\times W}$. The spectral bias generator uses $\mathcal{F}^b$ as input to produce spectral biases, represented as $\mathcal{D}^{b} \in \mathbb{R}^{L}$. In the channel bias generator, $\mathcal{F}^a$ and $\mathcal{F}^b$ are concatenated to form channel biases, denoted as $\mathcal{D}^{c} \in \mathbb{R}^{C}$. Finally, in the combination section, $\mathcal{D}^{a}$, $\mathcal{D}^{b}$, and $\mathcal{D}^{c}$ are integrated, similar to the process used in the kernel generator, resulting in adaptive biases represented as $\mathcal{D} \in \mathbb{R}^{H\times W\times L\times C}$.

\subsection{Ada3D Block}
The Ada3D block, shown in Fig.~\ref{ada3d}, is specifically designed to overcome the limitations of standard 3D convolution and achieve sufficient information integration. Given inputs $\mathcal{F}^{a}$ and $\mathcal{F}^{b}$, the Ada3D block first produces $\mathcal{K}$ and $\mathcal{D}$ using kernel and bias generators. These are then applied to process $\mathcal{F}^b$ through the Ada3D operation. Let $\boldsymbol{D}\in\mathbb{R}^{C\times (HWL)}$, $\boldsymbol{F}\in\mathbb{R}^{C\times (HWL)}$, and $\boldsymbol{\tilde{F}}\in\mathbb{R}^{C\times (HWL)}$ denote the mode-$4$ unfolding of $\mathcal{D}$, $\mathcal{F}^b$, and the resulting feature map, respectively. The Ada3D operation is mathematically expressed as:
\begin{equation}
\footnotesize
\label{eada}
{\boldsymbol{\tilde{F}}{(\cdot,i)}} = {\boldsymbol{D}{(\cdot,i)}} + \sum_{j \in \Omega (i)}\mathcal{K}({p}_i,\cdot,{p}_{i}-{p}_{j})\odot{\boldsymbol{F}}{(\cdot,j)},\quad i=1, 2, \cdots, HWL.
\end{equation}
Here, $\Omega (i)$ is a collection of 1D positional indices that locate elements within the $k\times k\times k$ convolution window centered around the $i$-th voxel of an $H\times W\times L$ tensor. Additionally, ${p}_{i}$ and ${p}_{j}$ denote the 3D positional indices of the $i$-th and $j$-th voxels. ${{p}_{i}-{p}_{j}}$ is a 3D positional offset that locates a weight within the $k\times k\times k$ kernel. Consequently, $\mathcal{K}({p}_i,\cdot,{p}_{i}-{p}_{j})\in\mathbb{R}^{C}$ represents a weighting vector in $\mathcal{K}$. Furthermore, $\odot$ defines the dot-product operation. Equation \ref{eada} illustrates that within an Ada3D block, each input voxel is convoluted with a distinct set of 3D kernels and refined by a particular set of biases. 

\begin{table*}[t]	
	\centering\renewcommand\arraystretch{1.}\setlength{\tabcolsep}{3.1pt}
	\belowrulesep=0pt\aboverulesep=0pt
	\caption{Quantitative evaluation results on testing samples of the WDC, Botswana, and Pavia datasets, which belong to the hyper-spectral pansharpening task. The best results are highlighted in \textbf{bold}, while the second-best results are \underline{underlined}. Methods listed above the dividing line represent traditional approaches, whereas those below it are DL-based techniques.}
	\begin{tabular}{l|c|ccccc|ccccc|ccccc}
		\toprule
		
		\multirow{2}{*}{\textbf{Methods}} & 
		\multirow{2}{*}{\textbf{Params}} & 
		\multicolumn{5}{c|}{\textbf{WDC}} & 
		\multicolumn{5}{c|}{\textbf{Botswana}} &
		\multicolumn{5}{c}{\textbf{Pavia}}
		\\
		\cmidrule(lr){3-7}\cmidrule(lr){8-12}\cmidrule(lr){13-17}
		&\multicolumn{1}{c|}{} 
		&\multicolumn{1}{c}{PSNR} 
		&\multicolumn{1}{c}{CC} 
		&\multicolumn{1}{c}{SSIM} 
		&\multicolumn{1}{c}{SAM} 
		&\multicolumn{1}{c|}{ERGAS}
		&\multicolumn{1}{c}{PSNR} 
		&\multicolumn{1}{c}{CC} 
		&\multicolumn{1}{c}{SSIM} 
		&\multicolumn{1}{c}{SAM} 
		&\multicolumn{1}{c|}{ERGAS} 
		&\multicolumn{1}{c}{PSNR} 
		&\multicolumn{1}{c}{CC} 
		&\multicolumn{1}{c}{SSIM} 
		&\multicolumn{1}{c}{SAM} 
		&\multicolumn{1}{c}{ERGAS}  
		\\
		
		\midrule
		
		\textbf{GLP} \cite{aiazzi2006mtf} & $-$ 
		& 27.946 & 0.934 & 0.761 & 6.546 & 5.110 
		& 32.559 & 0.951 & 0.837 & 1.383 & 1.207 
		& 31.944 & 0.935 & 0.749 & 6.099 & 4.909
		\\  
		\textbf{CNMF} \cite{6049465} & $-$ 
		& 24.604 & 0.890 & 0.678 & 8.441 & 6.682 
		& 30.220 & 0.917 & 0.788 & 1.934 & 1.718 
		& 31.184 & 0.894 & 0.659 & 6.953 & 6.263 
		\\ 
		\textbf{Hysure} \cite{7000523} & $-$ 
		& 25.598 & 0.913 & 0.718 & 7.254 & 5.834 
		& 30.610 & 0.928 & 0.796 & 1.747 & 1.595 
		& 32.208 & 0.921 & 0.730 & 6.240 & 5.474 
		\\ 
		\midrule
		\textbf{HyperPNN} \cite{8731649} & 0.13-0.14M
		& 29.258 & 0.945 & 0.860 & 4.051 & 5.749 
		& 33.114 & 0.961 & 0.873 & 1.366 & 1.195 
		& 33.394 & 0.963 & 0.827 & 4.566 & 3.750 
		\\ 
		\textbf{HSpeNet1} \cite{9200718} & 0.18-0.19M 
		& 29.634 & 0.960 & 0.870 & 4.039 & 4.266 
		& 31.746 & 0.942 & 0.844 & 1.456 & 1.663 
		& 33.612 & 0.964 & 0.824 & 4.690 & 3.721 
		\\ 
		\textbf{HSpeNet2} \cite{9200718} & 0.11-0.13M 
		& 29.700 & 0.961 & 0.872 & 4.009 & 4.261 
		& 32.575 & 0.953 & 0.849 & 1.400 & 1.348 
		& 33.472 & 0.962 & 0.819 & 4.642 & 3.818 
		\\ 
		\textbf{FusionNet} \cite{2020Detail} & 0.21-0.26M 
		& 29.696 & 0.959 & 0.866 & 3.917 & 4.339 
		& 32.506 & 0.952 & 0.850 & 1.397 & 1.367 
		& {34.739} & {0.969} & 0.847 & 4.462 & 3.446 
		\\ 
		\textbf{HyperDSNet} \cite{9870551} & 0.18-0.31M 
		& {30.232} & {0.964} & {0.875} & {4.102} & {3.943} 
		& {33.538} & {0.964} & {0.876} & {1.305} & {1.126} 
		& 34.376 & {0.969} & {0.849} & {4.295} & {3.434}
		\\ 
		\textbf{FPFNet} \cite{10298274} & 3.00-3.06M 
		& 30.291 & 0.957 & 0.855 & 4.440 & 4.250 
		& 33.451 & 0.962 & 0.871 & 1.369 & 1.135 
		& 33.581 & 0.959 & 0.825 & 4.627 & 3.931 
		\\   
		\textbf{Pan-Mamba} \cite{he2024pan} & 0.53-0.59M 
		& {30.794} & \underline{0.968} & {0.884} & {3.632} & \underline{3.693} 
		& \underline{33.901} & \underline{0.965} & \underline{0.882} & \underline{1.286} & \underline{1.089} 
		& 34.057 & 0.967 & {0.857} & 4.508 & 3.689 
		\\    
		\textbf{ADWM} \cite{Huang_2025_CVPR} & 0.50-0.62M
		& 29.960 & 0.959 & 0.871 & 3.909 & 4.155
		& 32.500 & 0.959 & 0.864 & 1.419 & 1.201
		& \underline{34.974} & \underline{0.972} & \underline{0.862} & 4.403 & 3.335 
		\\    
	\textbf{DFCFN} \cite{10835747} & 0.50-0.57M
		& \underline{31.301} & {0.967} & \underline{0.885} & \underline{3.502} & 4.038
		& 33.489 & 0.964 & 0.879 & 1.307 & 1.106
		& 34.894 & 0.971 & \underline{0.862} & \underline{4.240} & \underline{3.238}
		\\   
		\textbf{Ada3D (Ours)} & 0.55-0.58M 
		& \textbf{31.998} & \textbf{0.971} & \textbf{0.893} & \textbf{3.461} & \textbf{3.414}  
		& \textbf{34.290} & \textbf{0.969} & \textbf{0.889} & \textbf{1.230} & \textbf{1.034}
		& \textbf{35.752} & \textbf{0.973} & \textbf{0.873} & \textbf{4.038} & \textbf{3.142}
		\\     
		\midrule
		\textbf{Ideal Values} 
		& $-$
		&\multicolumn{1}{c}{\textbf{+$\infty$}}
		&\multicolumn{1}{c}{\textbf{\textbf{1}}}
		&\multicolumn{1}{c}{\textbf{\textbf{1}}}
		&\multicolumn{1}{c}{\textbf{\textbf{0}}}
		&\multicolumn{1}{c|}{\textbf{\textbf{0}}}
		&\multicolumn{1}{c}{\textbf{+$\infty$}}
		&\multicolumn{1}{c}{\textbf{\textbf{1}}}
		&\multicolumn{1}{c}{\textbf{\textbf{1}}}
		&\multicolumn{1}{c}{\textbf{\textbf{0}}}
		&\multicolumn{1}{c|}{\textbf{\textbf{0}}}
		&\multicolumn{1}{c}{\textbf{+$\infty$}}
		&\multicolumn{1}{c}{\textbf{\textbf{1}}}
		&\multicolumn{1}{c}{\textbf{\textbf{1}}}
		&\multicolumn{1}{c}{\textbf{\textbf{0}}}
		&\multicolumn{1}{c}{\textbf{\textbf{0}}}
		\\ 
		\bottomrule
	\end{tabular}
	\label{hsp}	
\end{table*}

\subsection{Parameters \& FLOPs Analysis}
The learnable parameters of an Ada3D block primarily come from its spatial and spectral kernel generators. The former contributes $9\alpha C^2(k^3+1)$ parameters, while the latter adds $3\beta C^2(k^3+1)$ parameters. Consequently, the total parameter count of an Ada3D block is $(9\alpha +3\beta)C^2(k^3+1)$. 
In terms of computational complexity, the spatial kernel generator, spectral kernel generator, combination section, and adaptive convolution operation require $18\alpha HWC^2(k^3+1)$, $6\beta LC^2(k^3+1)$, $HWLCk^3$, and $2HWLCk^3$ FLOPs, respectively. Therefore, the total FLOP count is around $3HWLCk^3+18\alpha HWC^2(k^3+1)+6\beta LC^2(k^3+1)$.
Table~\ref{flops} provides a comparison of Ada3D with standard 3D convolution, depth-wise 3D convolution, and PAC3D in terms of parameters and FLOPs. PAC3D, our proposed 3D extension of PAC \cite{su2019pixeladaptive}, employs a standard $3\times 3\times 3$ convolutional layer to generate adaptive 3D kernels, which leads to considerable computational cost. In practice, the ratio $L/C$ is typically less than $5$. Consequently, as demonstrated in the third row of Table~\ref{flops}, Ada3D achieves lower FLOPs than other 3D convolutional paradigms with an equivalent number of parameters.

\subsection{Loss function}
The loss function for our method is crafted as a weighted sum of two distinct loss terms, as illustrated below:
\begin{equation}
	\begin{cases}
		{\mathcal{L}oss} = \mathcal{L}_1 + {\lambda}_{ergas} \mathcal{L}_{ergas} \\
		{\mathcal{L}_{ergas}} = \frac{1}{M}\sum\limits_{m=1}^M{\rm{ERGAS}}(\mathcal{\tilde{O}}^{\{m\}}, \mathcal{O}^{\{m\}}) \\
		{\rm{ERGAS}}(\mathcal{\tilde{O}}, \mathcal{O}) =\frac{100}{r}\sqrt{\frac{1}{L}\sum\limits_{i=1}^{L}\frac{{\rm{MSE}}({\mathcal{\tilde{O}}}_i,{\mathcal{O}}_i)}{{\mu}^{2}_{{\mathcal{\tilde{O}}}_i}}}
	\end{cases}.
\end{equation} 
Here, $\mathcal{L}_1$ denotes the widely used sum of absolute differences (SAD). ${\lambda}_{ergas}$ is a hyper-parameter associated with $\mathcal{L}_{ergas}$. ERGAS stands for the relative dimensionless global error in synthesis \cite{1997Fusion}, while MSE represents the mean square error. $M$ is the total number of training samples, and $r$ is the up-sampling scale, which is set to $4$ in this paper. Additionally, $\mathcal{\tilde{O}}^{\{m\}}$ and $\mathcal{O}^{\{m\}}$ refer to the $m$-th $\mathcal{\tilde{O}}$ and $\mathcal{O}$ in the training dataset. $\mathcal{\tilde{O}}_i$ and $\mathcal{O}_i$ denote the $i$-th spectral band in $\mathcal{\tilde{O}}$ and $\mathcal{O}$. Furthermore, ${\mu}^2$ represents the root mean square (RMS). In this paper, we utilize the ERGAS loss function to enhance the model's ability to capture and learn spectral information.

\section{Experiments}
\label{s5}

\begin{figure*}[t]
	\begin{center}
		\begin{minipage}[t]{0.95\linewidth}
			\begin{minipage}[t]{0.086\linewidth}
				{\includegraphics[width=1\linewidth]{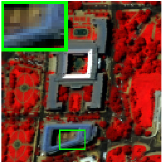}}
				\vspace{2pt}
				{\includegraphics[width=1\linewidth]{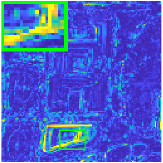}}
				{\includegraphics[width=1\linewidth]{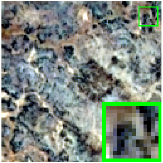}}
				\vspace{2pt}
				{\includegraphics[width=1\linewidth]{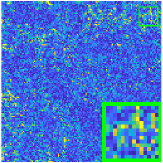}}
				{\includegraphics[width=1\linewidth]{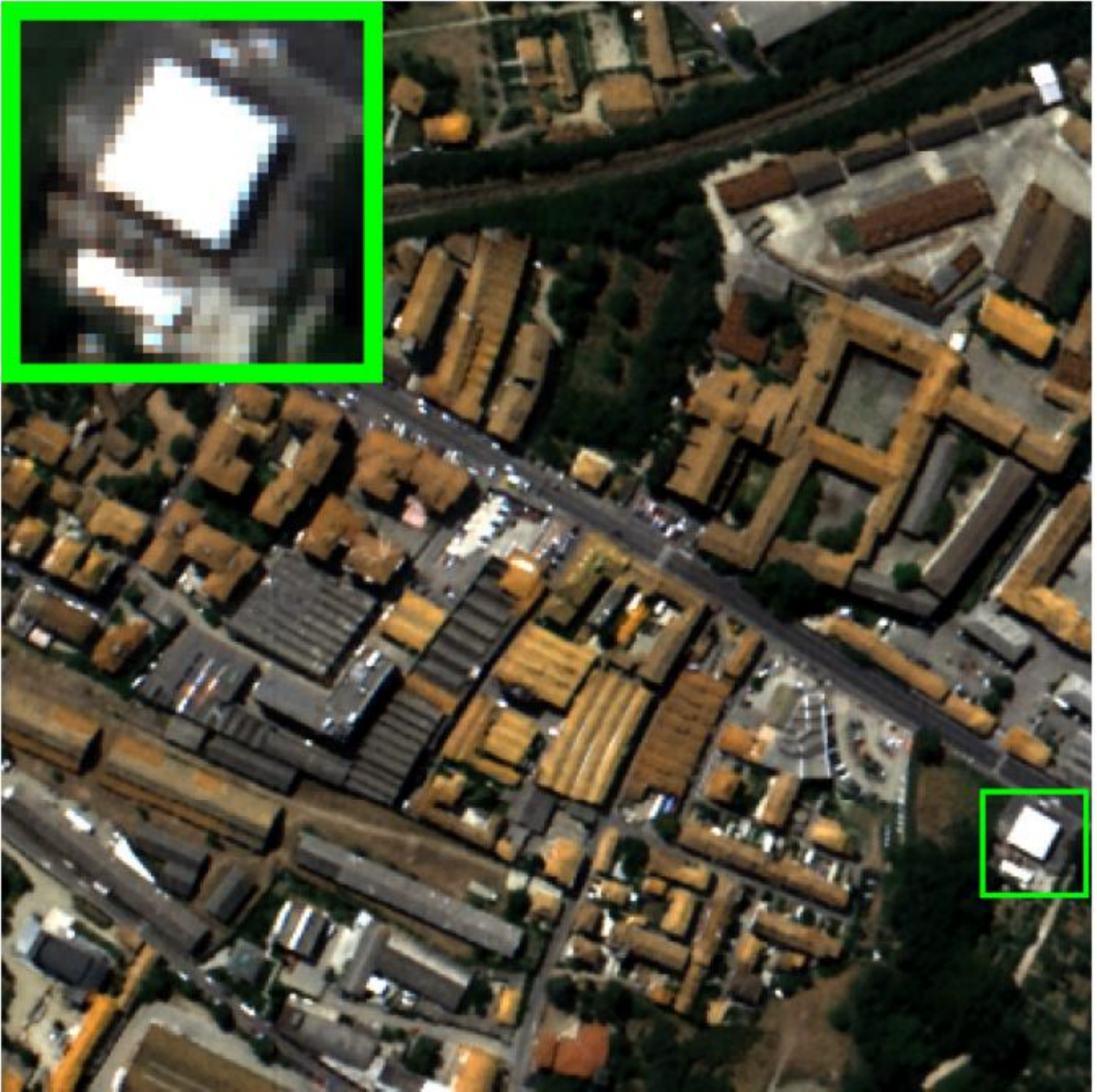}}
				{\includegraphics[width=1\linewidth]{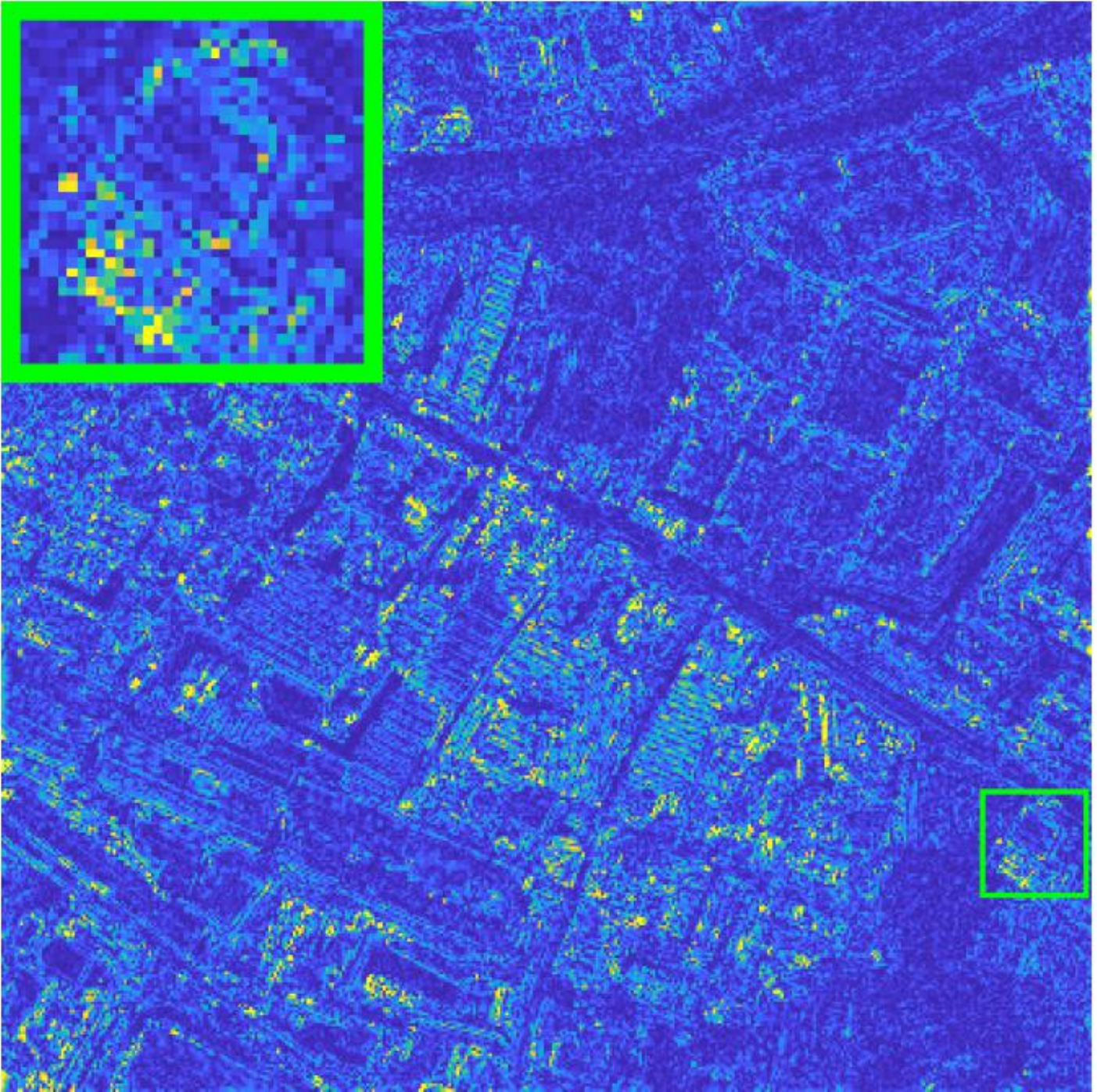}}
				\vspace{2pt}
				\scriptsize{HyperPNN}
				\centering
				
			\end{minipage}
			\begin{minipage}[t]{0.086\linewidth}
				{\includegraphics[width=1\linewidth]{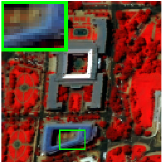}}
				\vspace{2pt}
				{\includegraphics[width=1\linewidth]{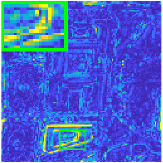}}
				{\includegraphics[width=1\linewidth]{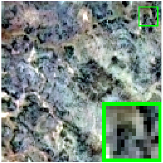}}
				\vspace{2pt}
				{\includegraphics[width=1\linewidth]{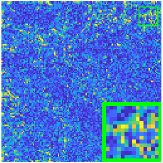}}
				{\includegraphics[width=1\linewidth]{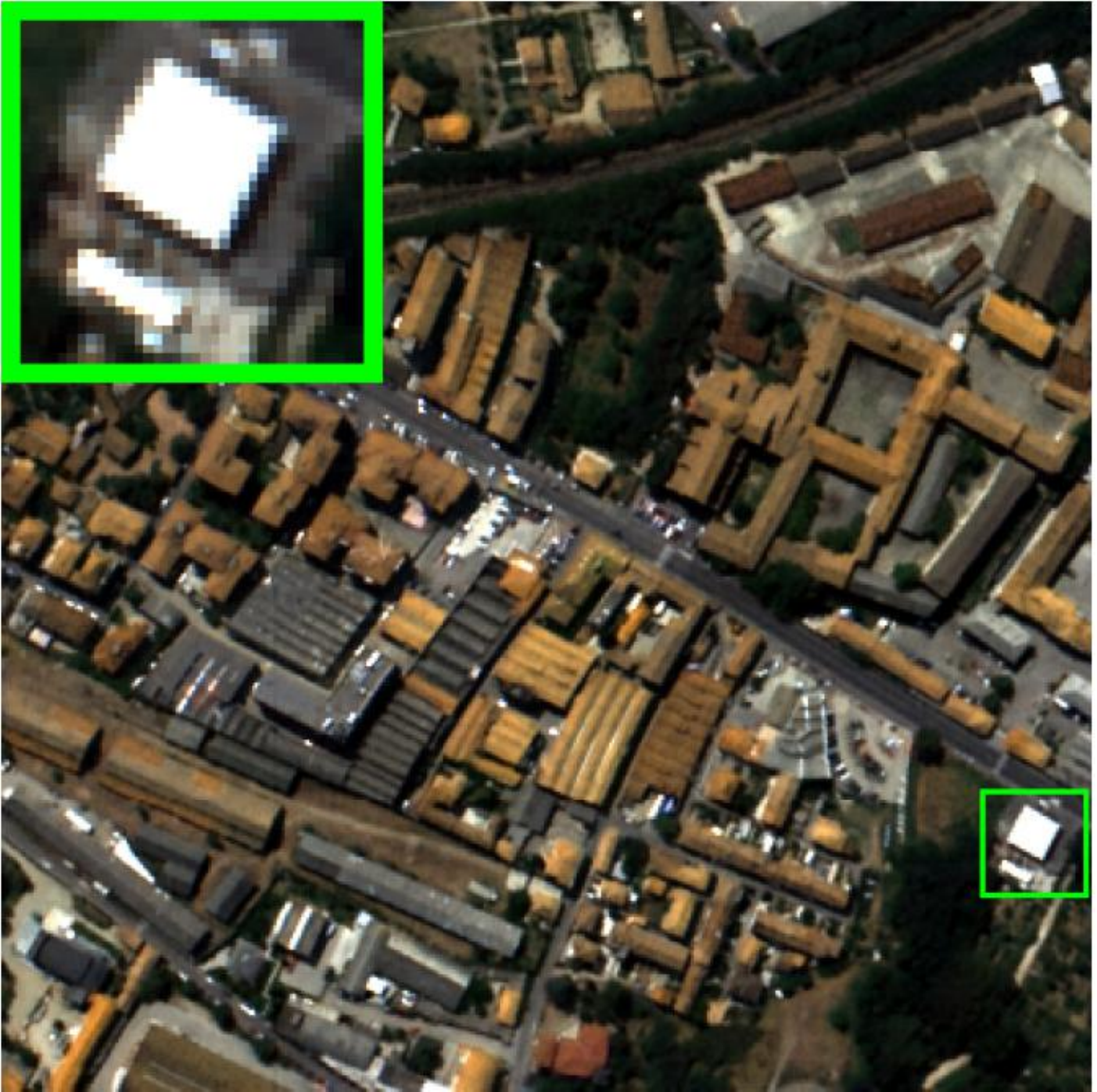}}
				{\includegraphics[width=1\linewidth]{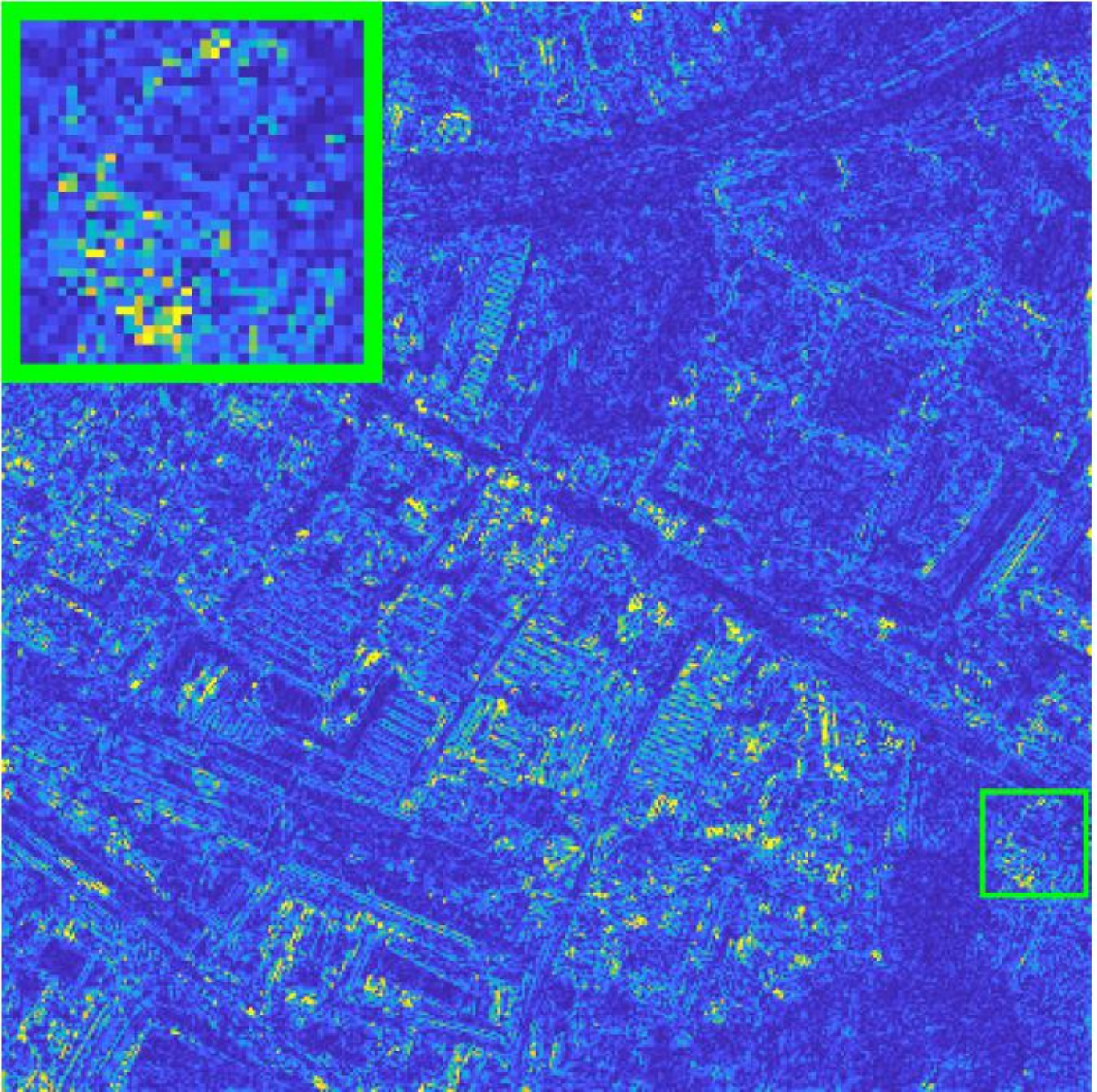}}
				\vspace{2pt}
				\scriptsize{HSpeNet1}
				\centering
				
			\end{minipage}
			\begin{minipage}[t]{0.086\linewidth}
				{\includegraphics[width=1\linewidth]{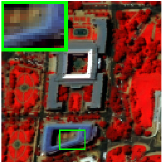}}
				\vspace{2pt}
				{\includegraphics[width=1\linewidth]{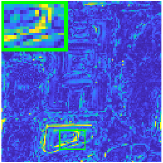}}
				{\includegraphics[width=1\linewidth]{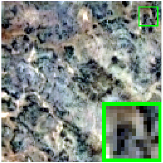}}
				\vspace{2pt}
				{\includegraphics[width=1\linewidth]{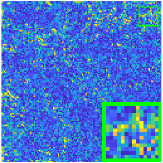}}
				{\includegraphics[width=1\linewidth]{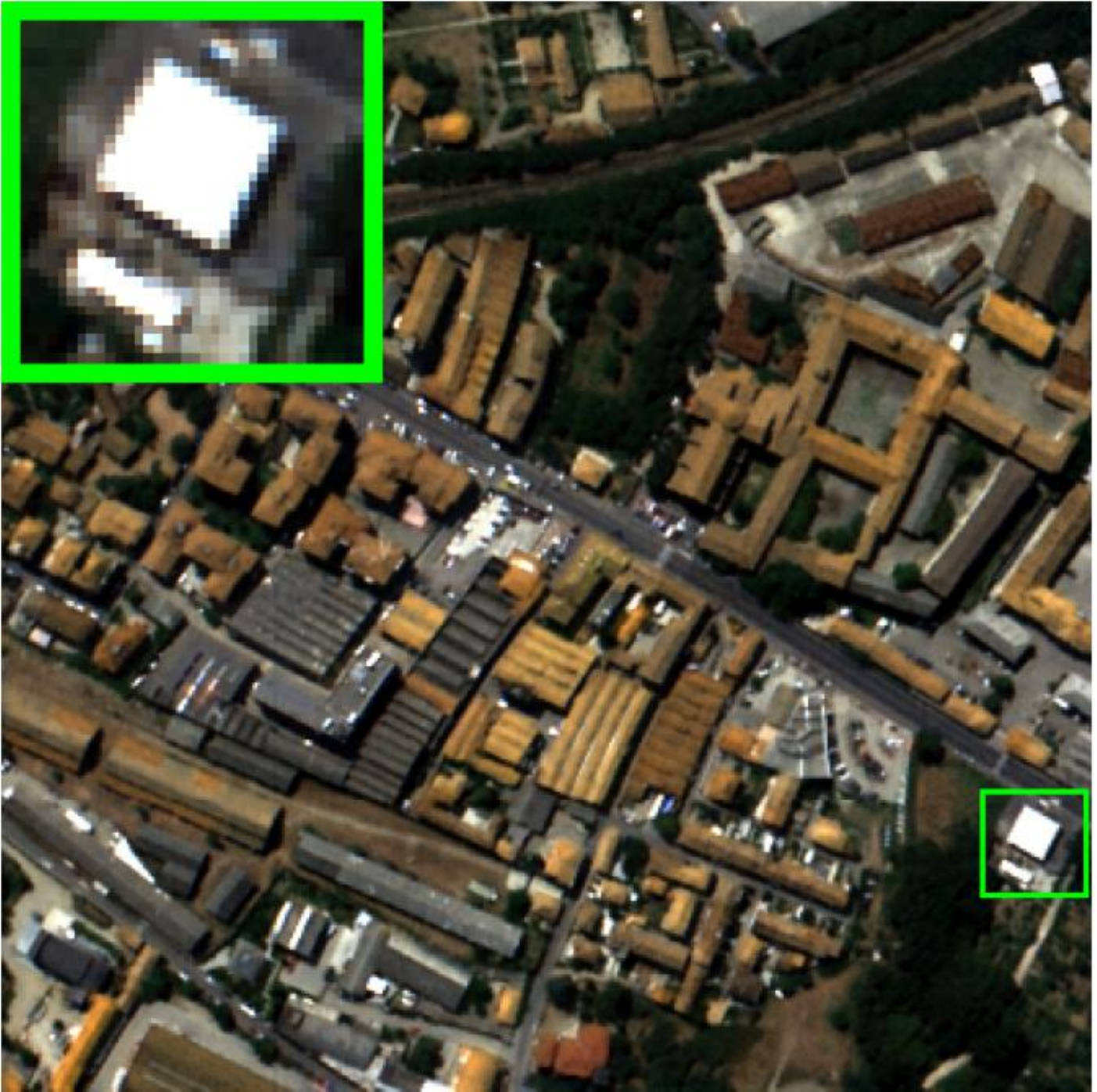}}
				{\includegraphics[width=1\linewidth]{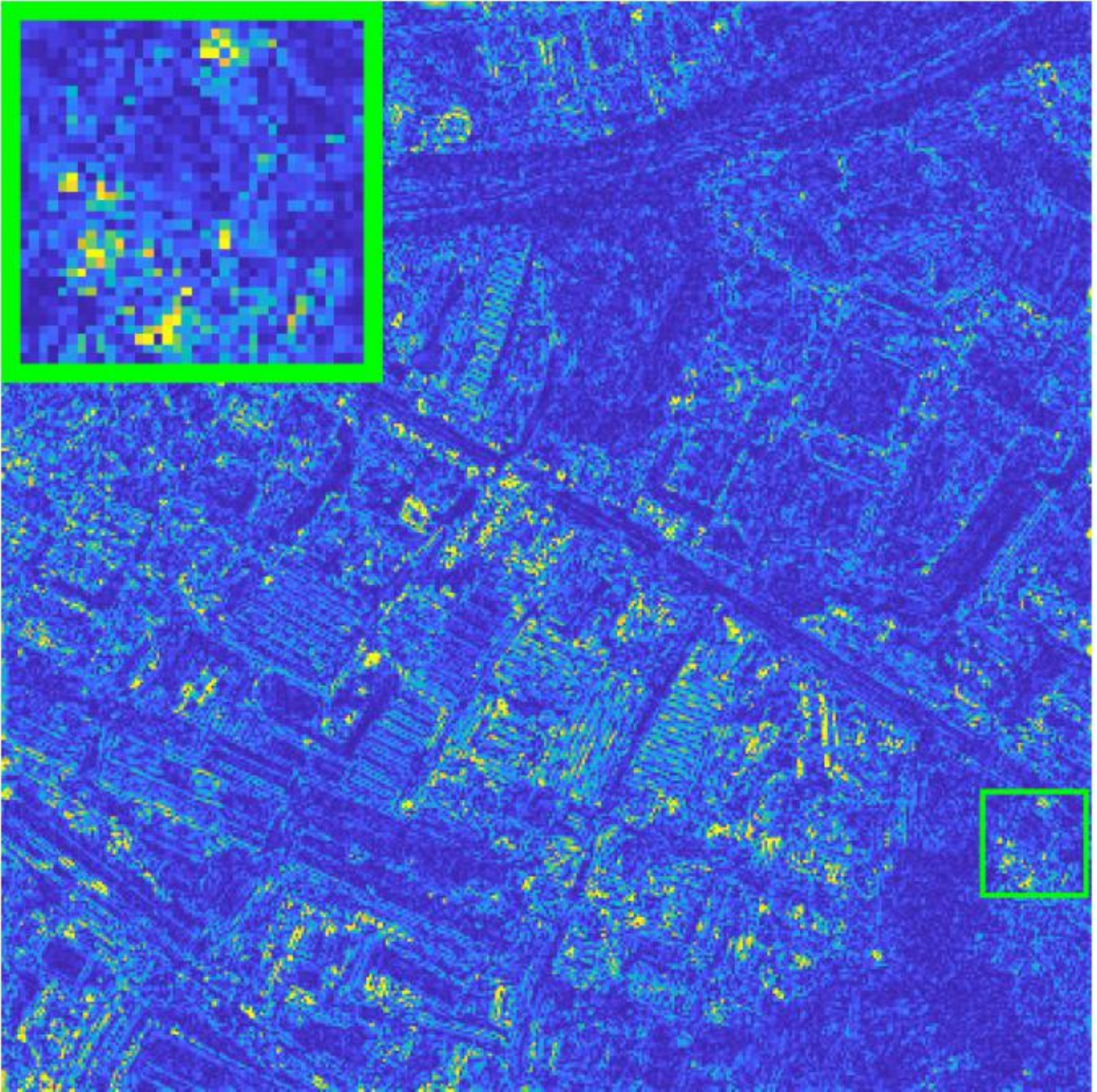}}
				\vspace{2pt}
				\scriptsize{HSpeNet2}
				\centering
				
			\end{minipage}
			\begin{minipage}[t]{0.086\linewidth}
				{\includegraphics[width=1\linewidth]{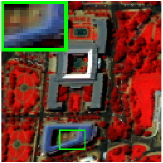}}
				\vspace{2pt}
				{\includegraphics[width=1\linewidth]{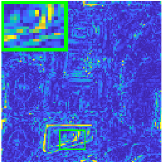}}
				{\includegraphics[width=1\linewidth]{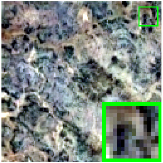}}
				\vspace{2pt}
				{\includegraphics[width=1\linewidth]{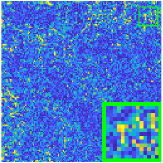}}
				{\includegraphics[width=1\linewidth]{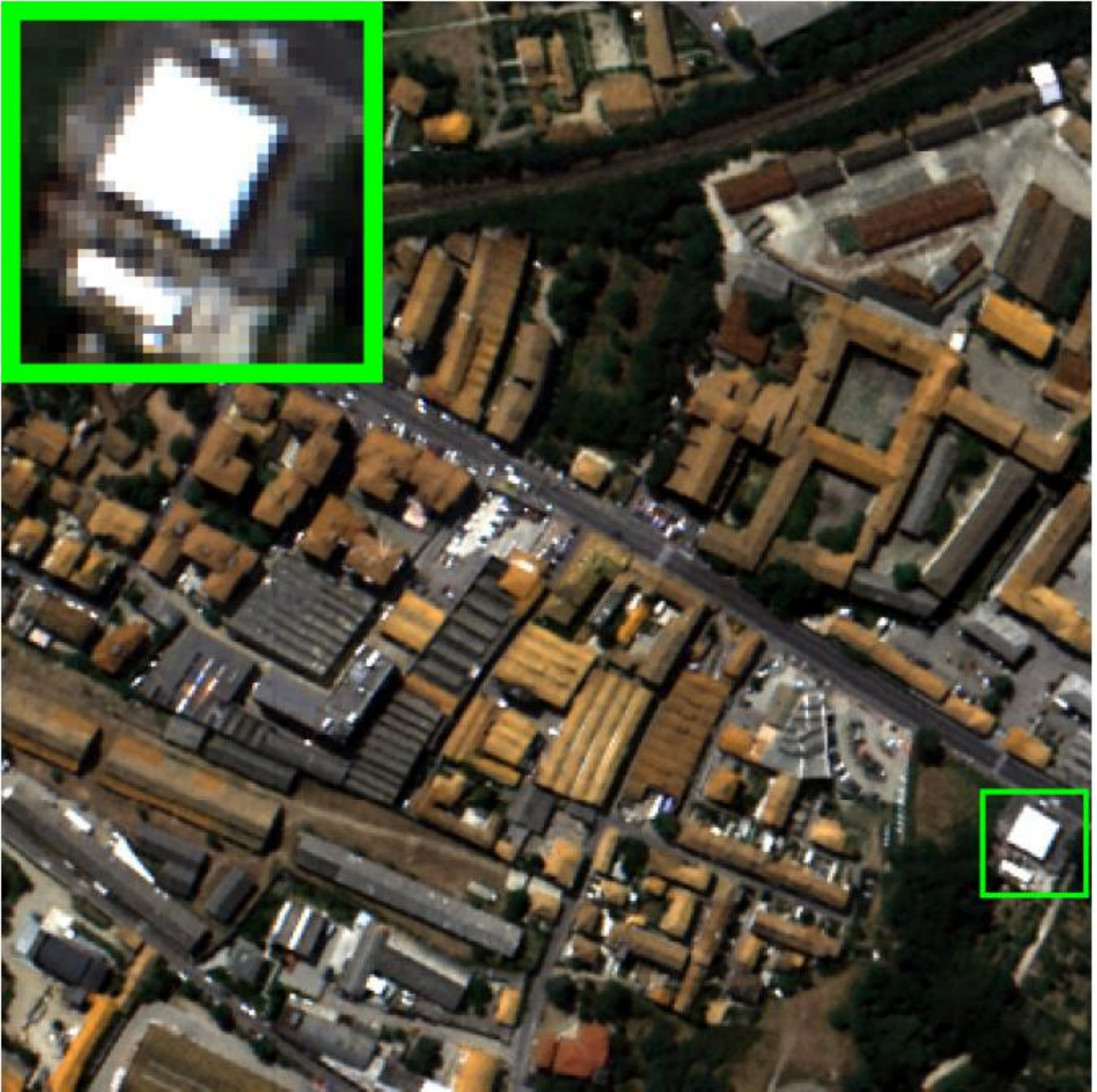}}
				{\includegraphics[width=1\linewidth]{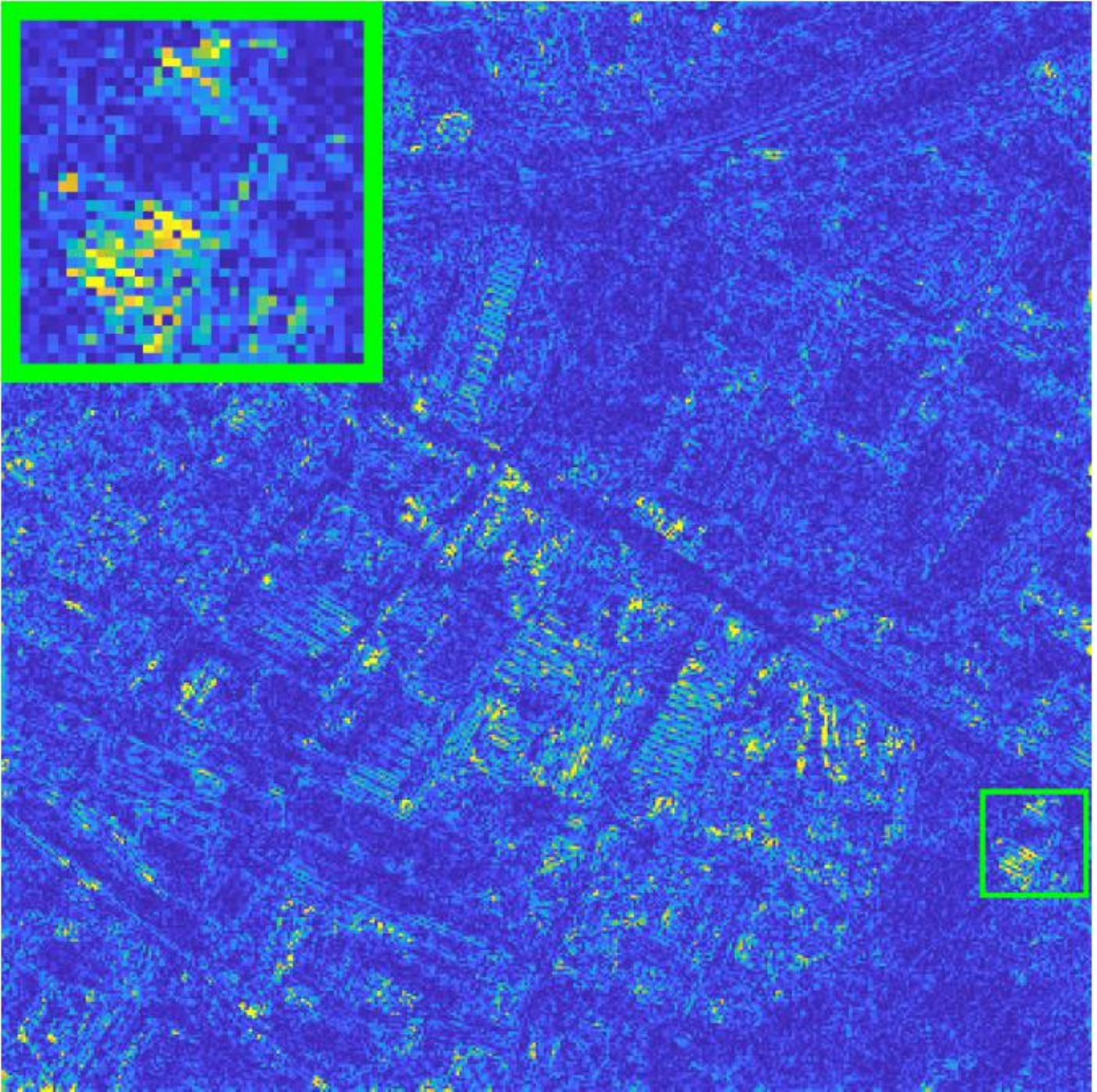}}
				\vspace{2pt}
				\scriptsize{FusionNet}
				\centering
				
			\end{minipage}
			\begin{minipage}[t]{0.086\linewidth}
				{\includegraphics[width=1\linewidth]{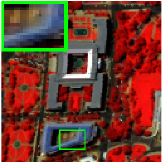}}
				\vspace{2pt}
				{\includegraphics[width=1\linewidth]{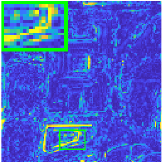}}
				{\includegraphics[width=1\linewidth]{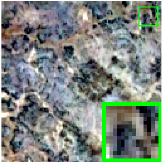}}
				\vspace{2pt}
				{\includegraphics[width=1\linewidth]{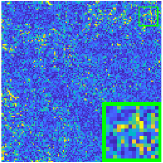}}
				{\includegraphics[width=1\linewidth]{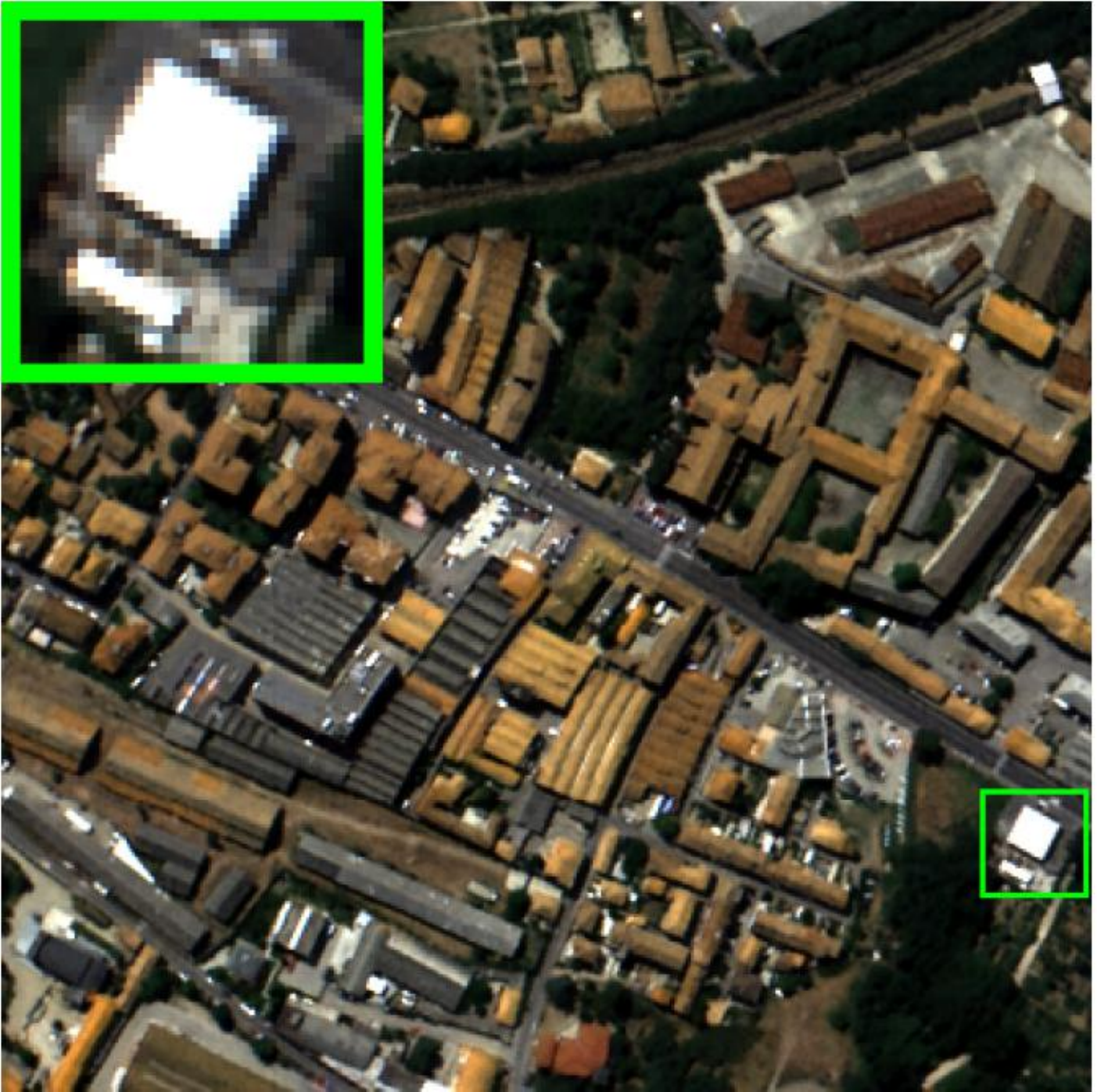}}
				{\includegraphics[width=1\linewidth]{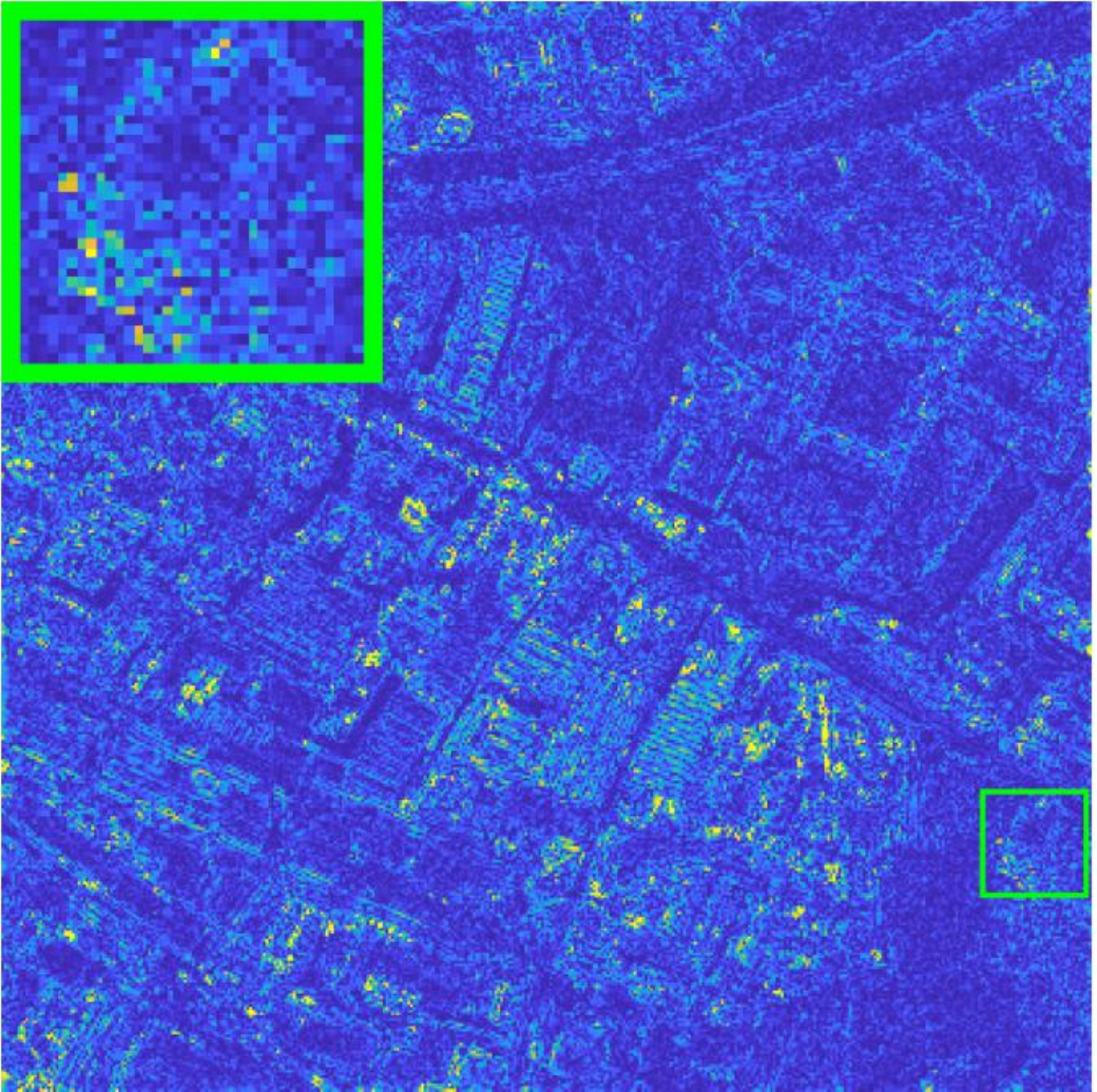}}
				\vspace{2pt}
				\scriptsize{Hyper-DSNet}
				\centering
				
			\end{minipage}
			\begin{minipage}[t]{0.086\linewidth}
				{\includegraphics[width=1\linewidth]{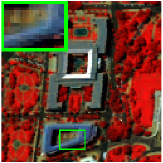}}
				\vspace{2pt}
				{\includegraphics[width=1\linewidth]{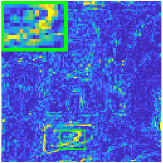}}
				{\includegraphics[width=1\linewidth]{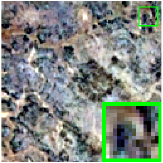}}
				\vspace{2pt}
				{\includegraphics[width=1\linewidth]{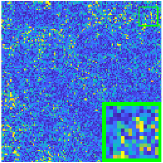}}
				{\includegraphics[width=1\linewidth]{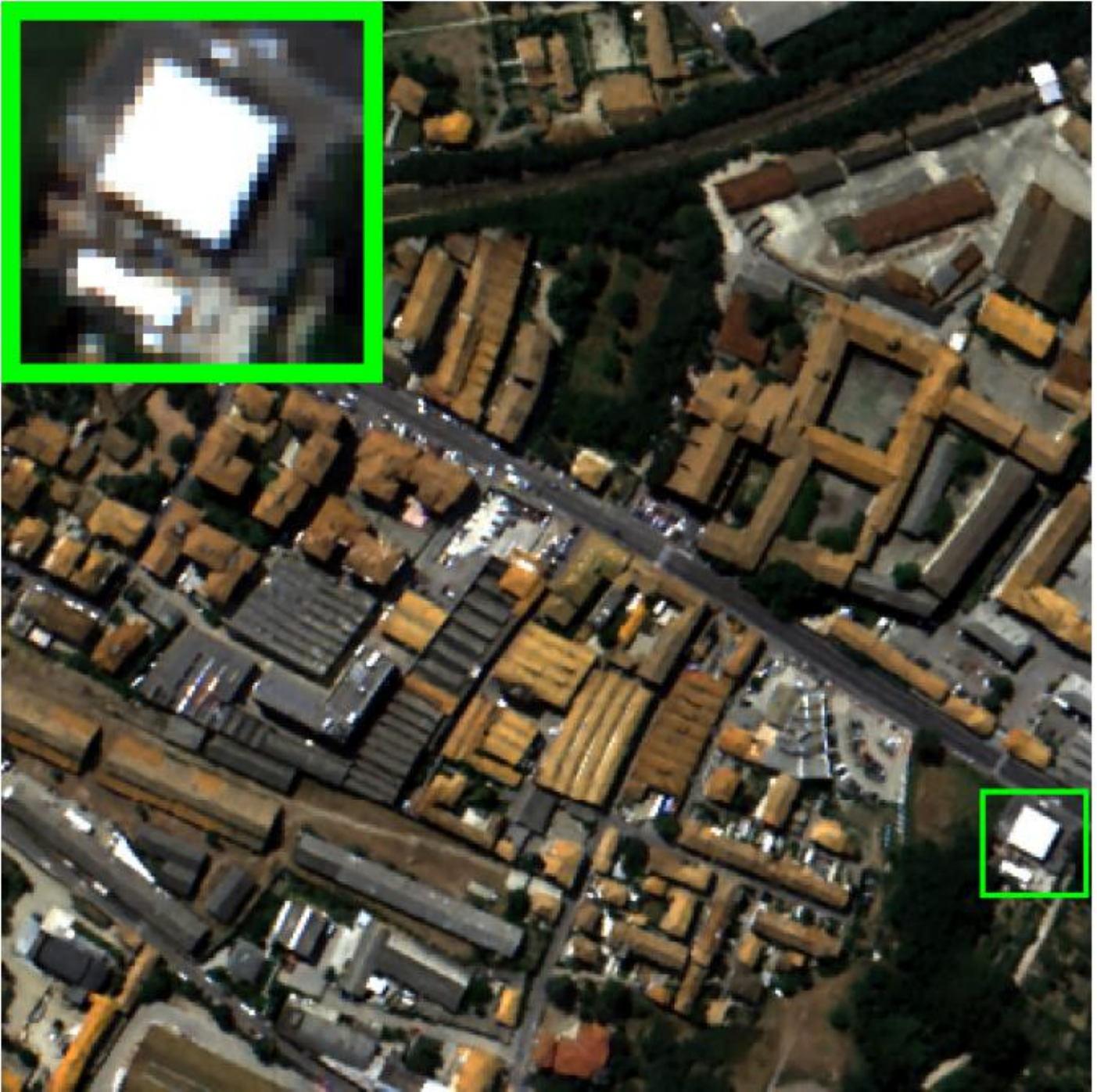}}
				{\includegraphics[width=1\linewidth]{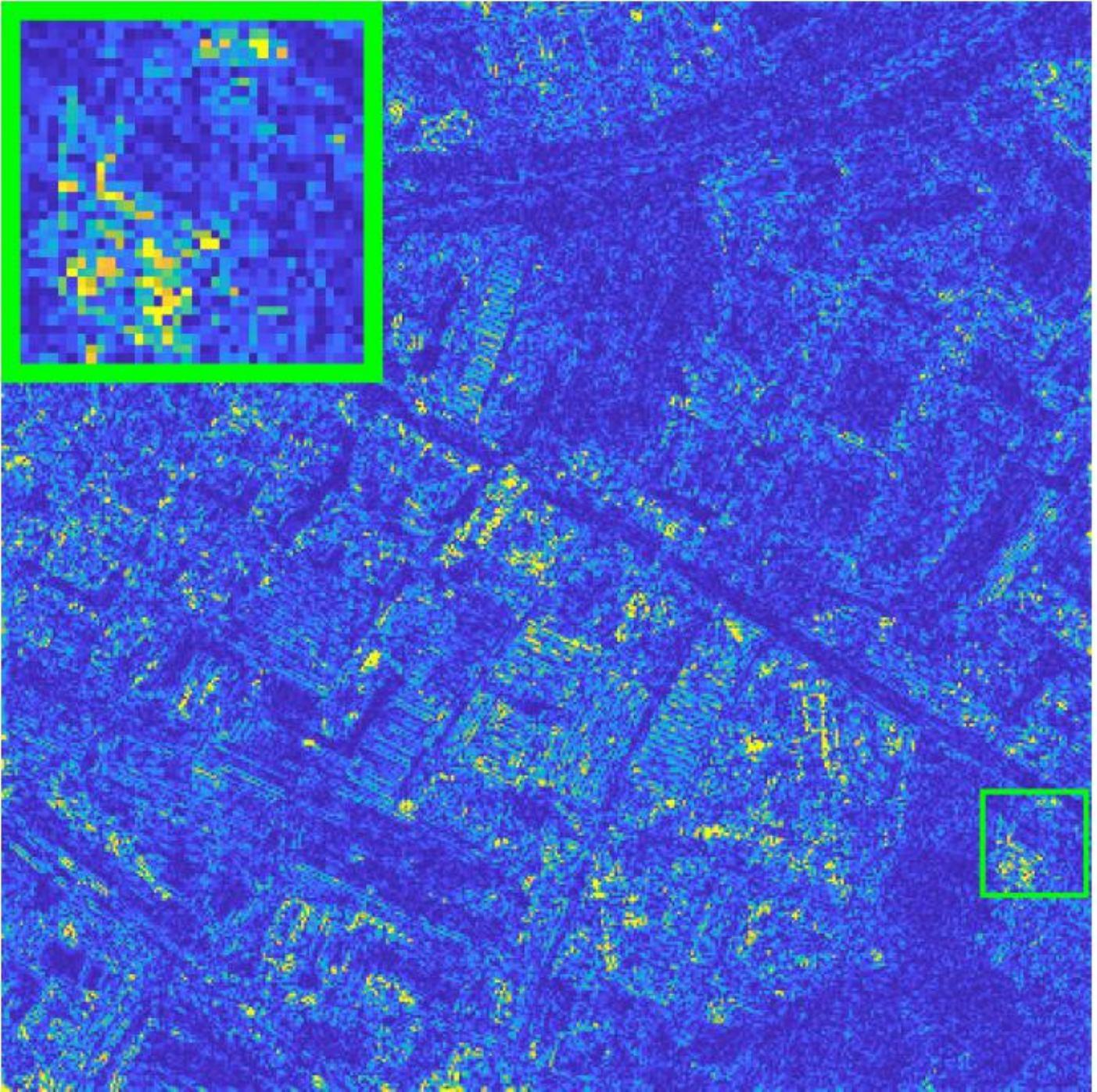}}
				\vspace{2pt}
				\scriptsize{FPFNet}
				\centering
				
			\end{minipage}
		    \begin{minipage}[t]{0.086\linewidth}
		    	{\includegraphics[width=1\linewidth]{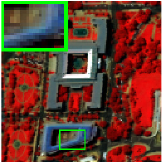}}
		    	\vspace{2pt}
		    	{\includegraphics[width=1\linewidth]{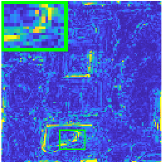}}
		    	{\includegraphics[width=1\linewidth]{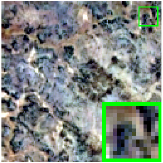}}
		    	\vspace{2pt}
		    	{\includegraphics[width=1\linewidth]{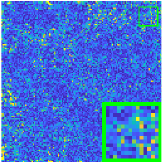}}
		    	{\includegraphics[width=1\linewidth]{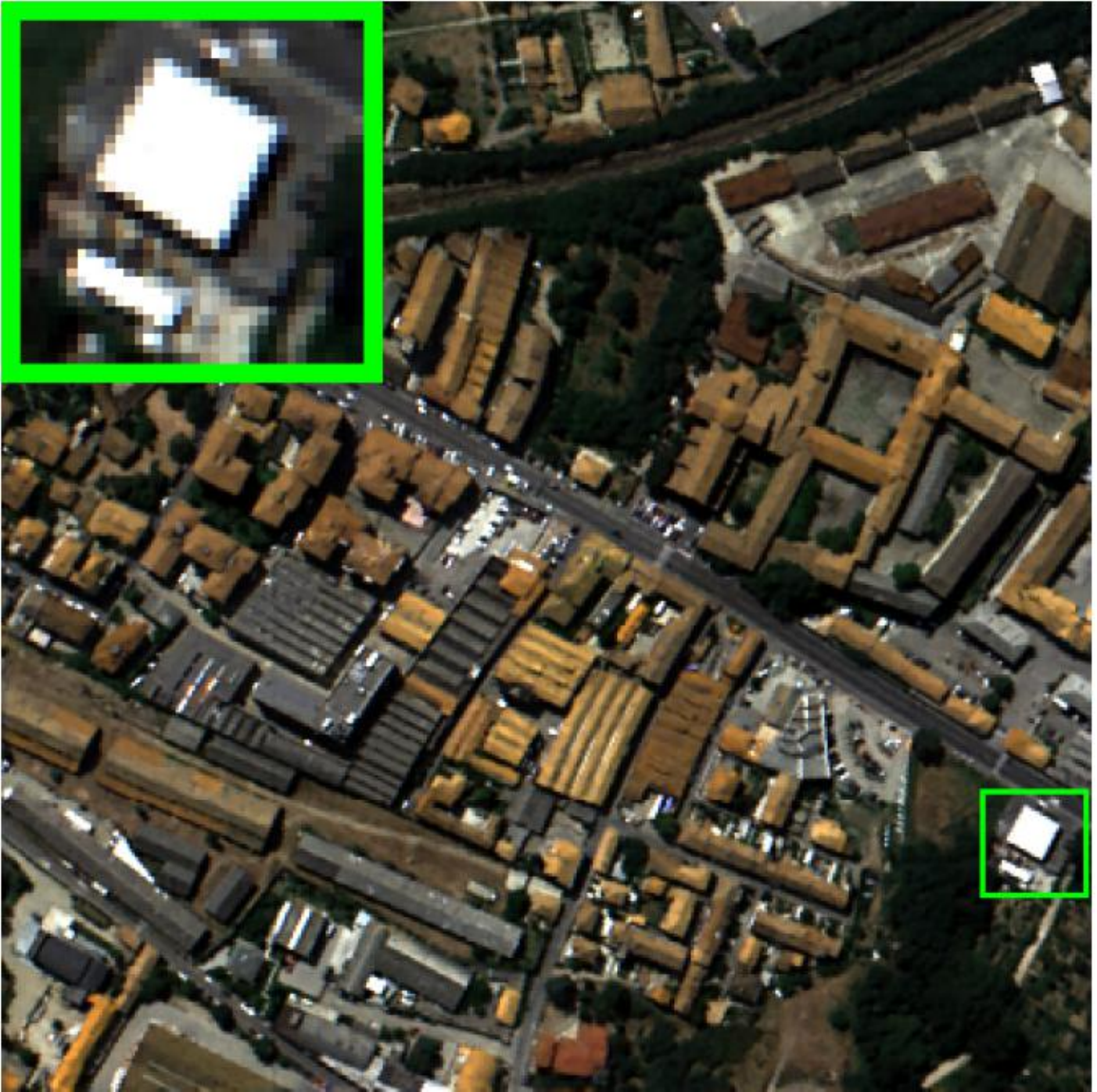}}
		    	{\includegraphics[width=1\linewidth]{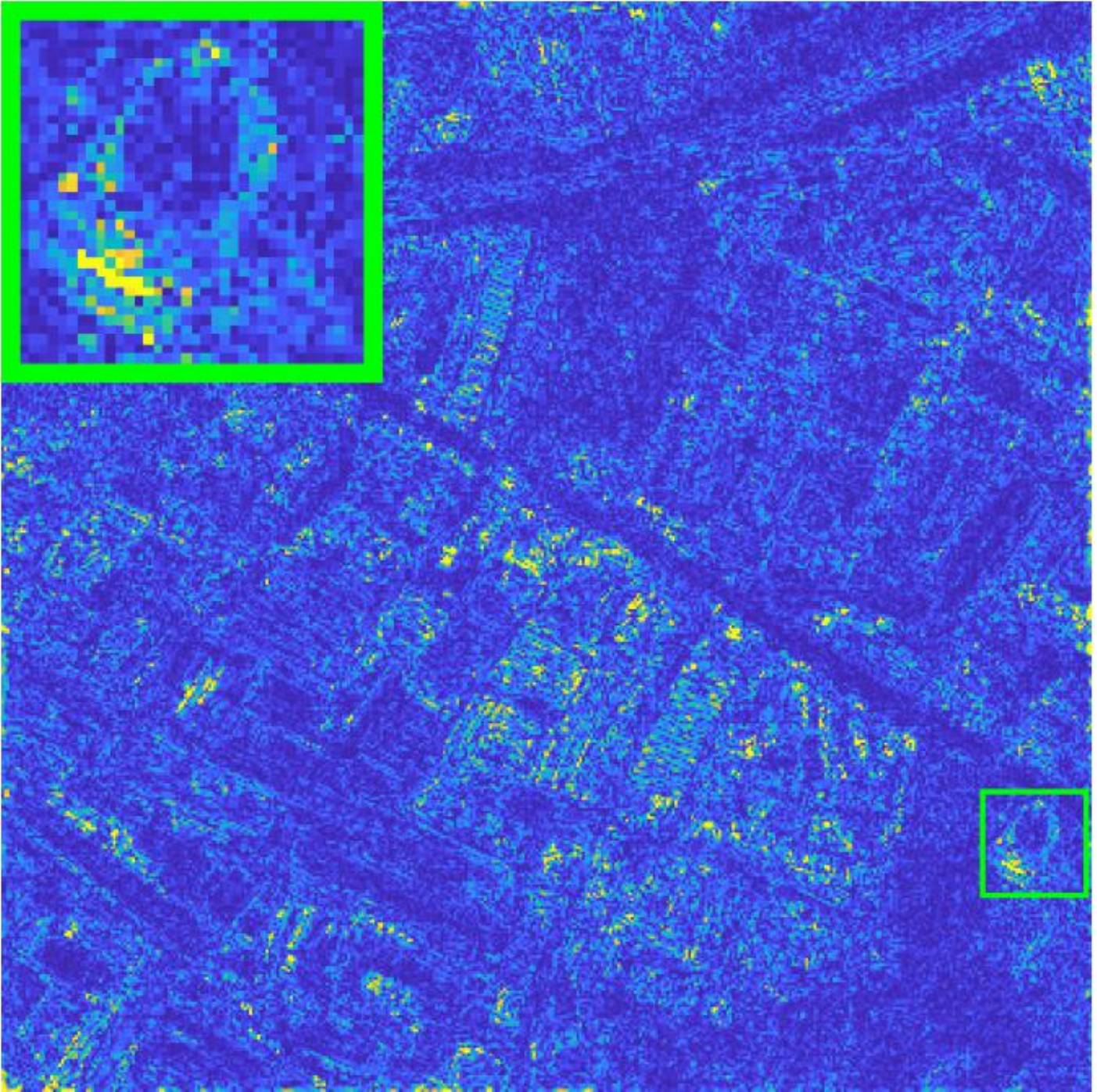}}
		    	\vspace{2pt}
		    	\scriptsize{Pan-Mamba}
		    	\centering
		    	
		    \end{minipage}
	    	\begin{minipage}[t]{0.086\linewidth}
	    		{\includegraphics[width=1\linewidth]{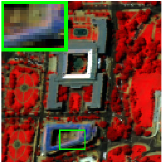}}
	    		\vspace{2pt}
	    		{\includegraphics[width=1\linewidth]{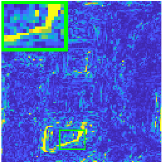}}
	    		{\includegraphics[width=1\linewidth]{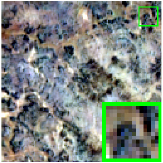}}
	    		\vspace{2pt}
	    		{\includegraphics[width=1\linewidth]{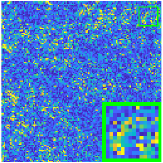}}
	    		{\includegraphics[width=1\linewidth]{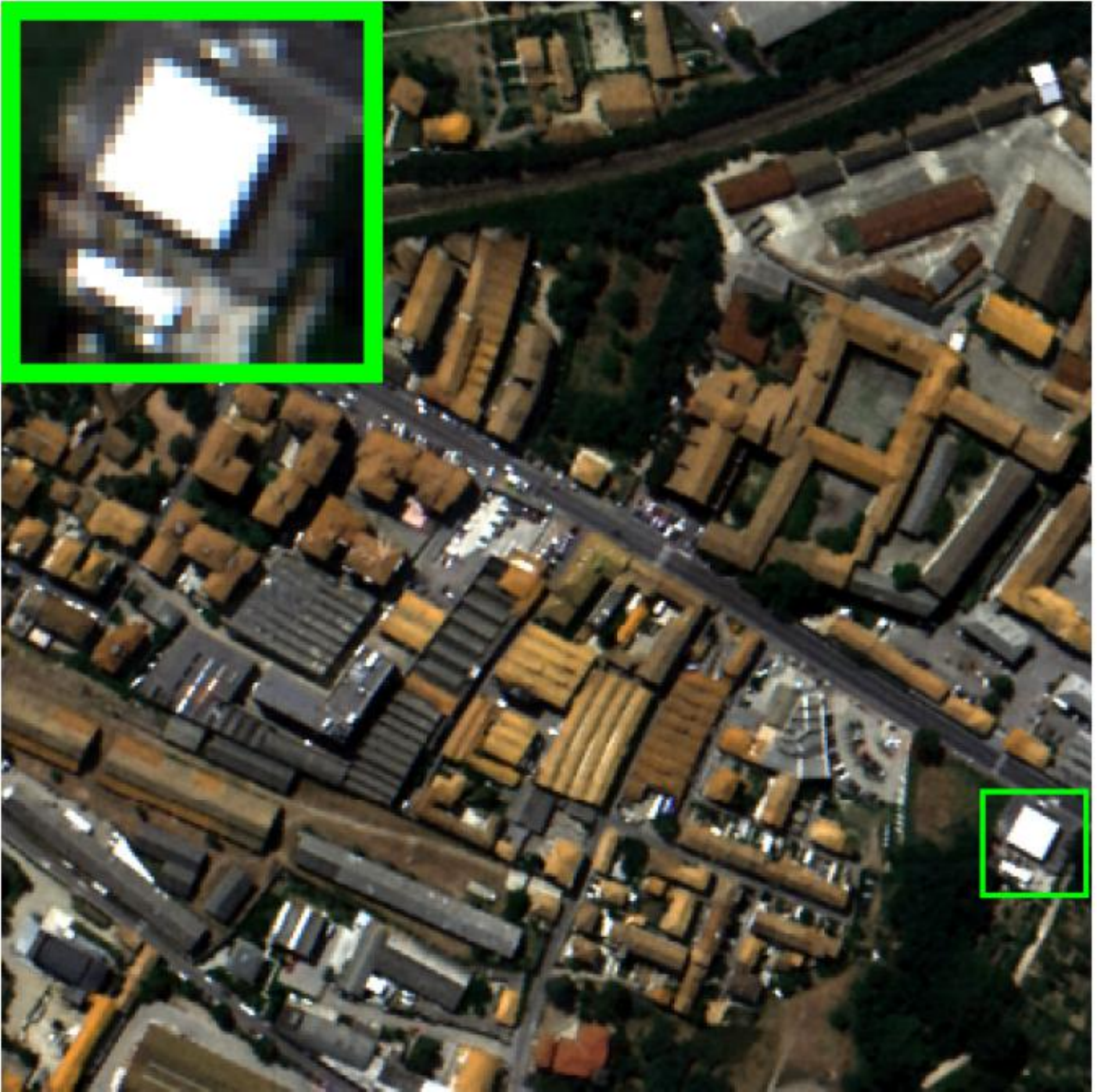}}
	    		{\includegraphics[width=1\linewidth]{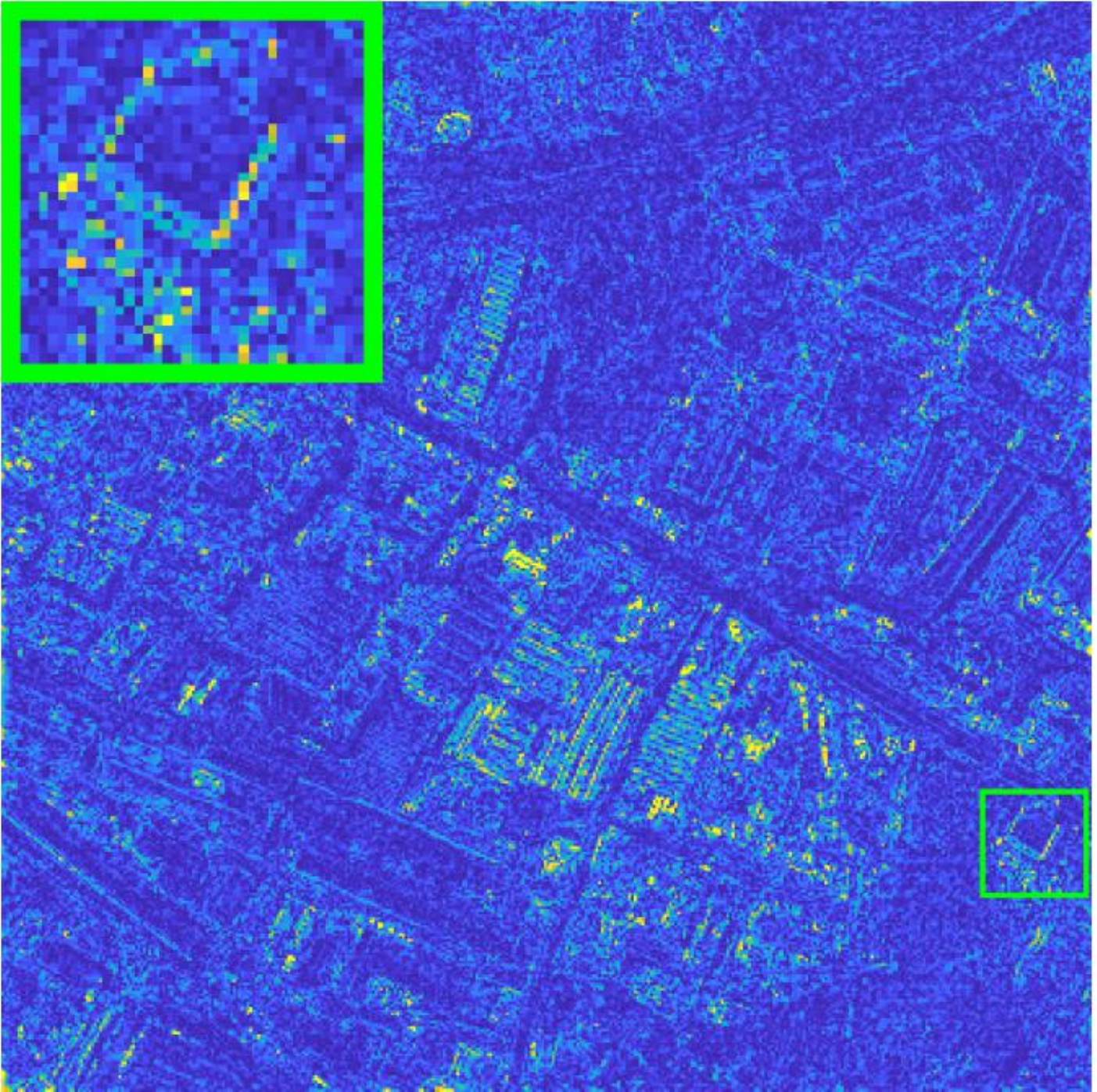}}
	    		\vspace{2pt}
	    		\scriptsize{ADWM}
	    		\centering
	    		
	    	\end{minipage}
    		\begin{minipage}[t]{0.086\linewidth}
    			{\includegraphics[width=1\linewidth]{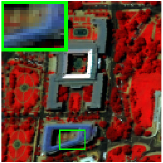}}
    			\vspace{2pt}
    			{\includegraphics[width=1\linewidth]{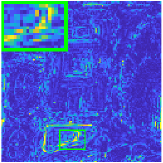}}
    			{\includegraphics[width=1\linewidth]{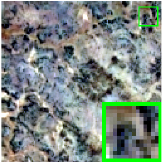}}
    			\vspace{2pt}
    			{\includegraphics[width=1\linewidth]{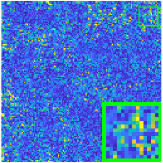}}
    			{\includegraphics[width=1\linewidth]{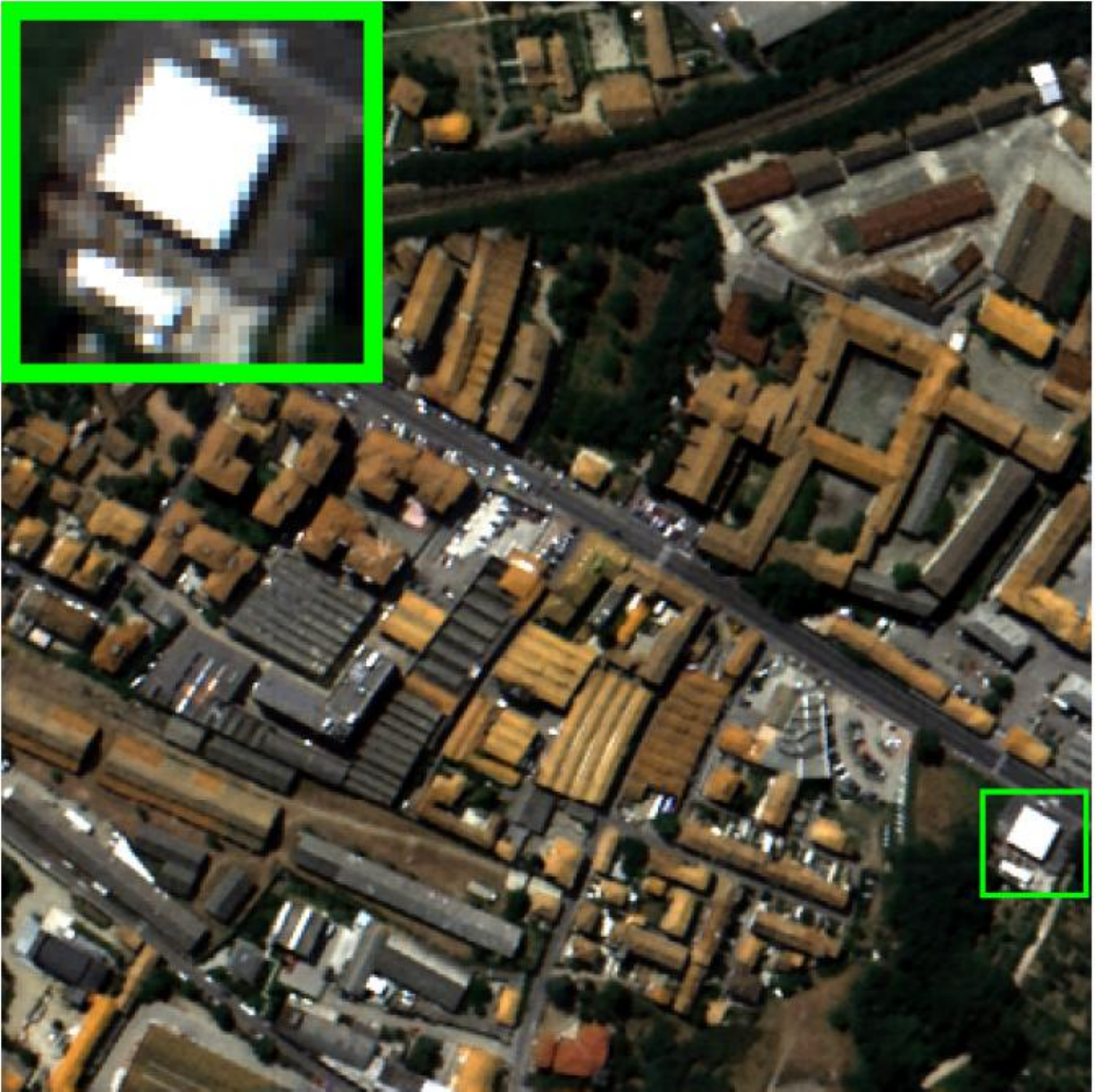}}
    			{\includegraphics[width=1\linewidth]{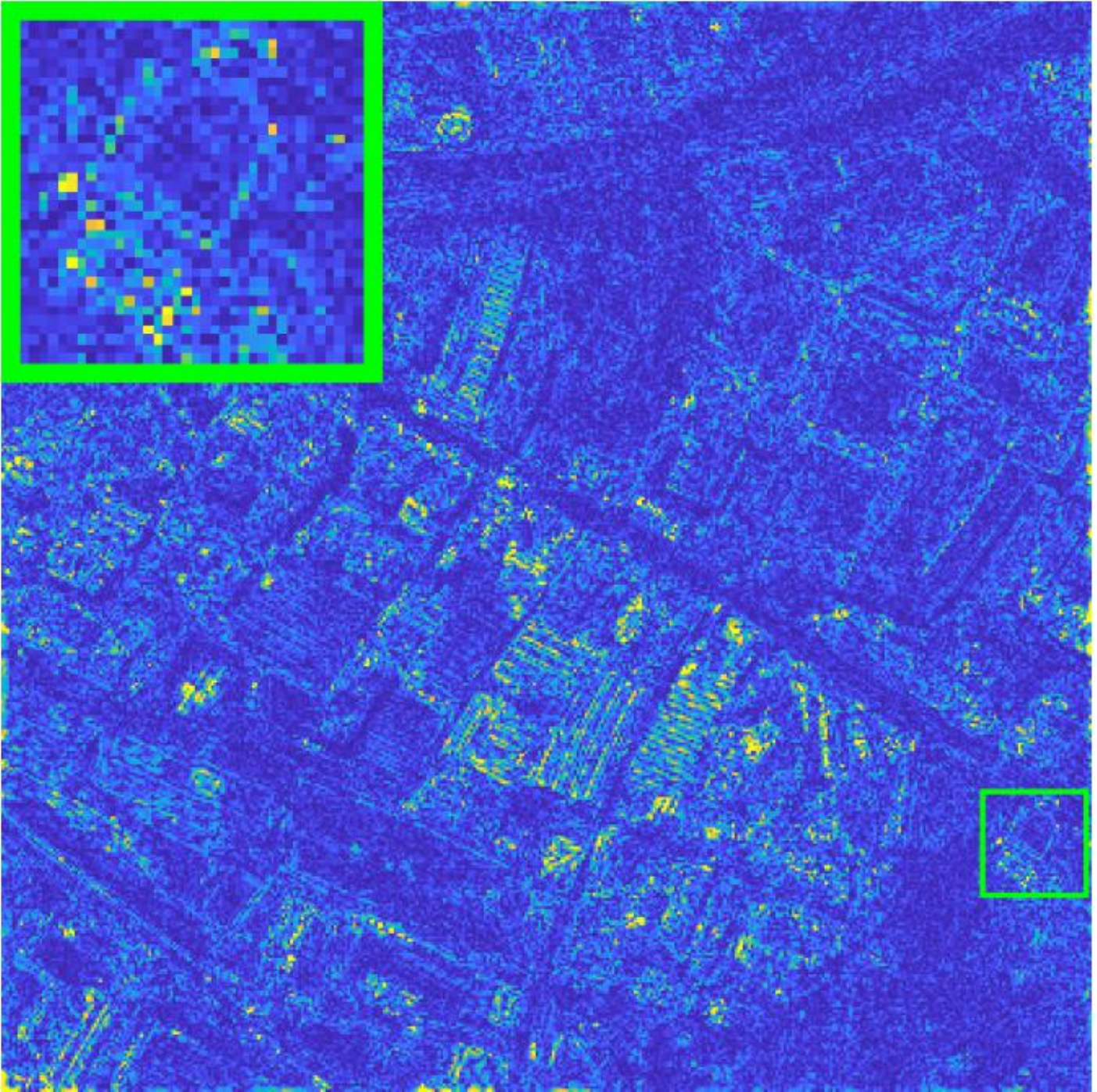}}
    			\vspace{2pt}
    			\scriptsize{DFCFN}
    			\centering
    			
    		\end{minipage}
			\begin{minipage}[t]{0.086\linewidth}
				{\includegraphics[width=1\linewidth]{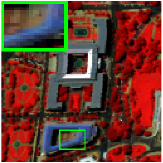}}
				\vspace{2pt}
				{\includegraphics[width=1\linewidth]{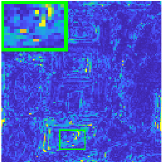}}
				{\includegraphics[width=1\linewidth]{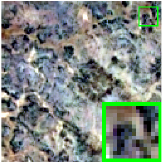}}
				\vspace{2pt}
				{\includegraphics[width=1\linewidth]{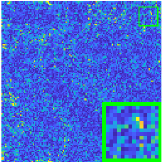}}
				{\includegraphics[width=1\linewidth]{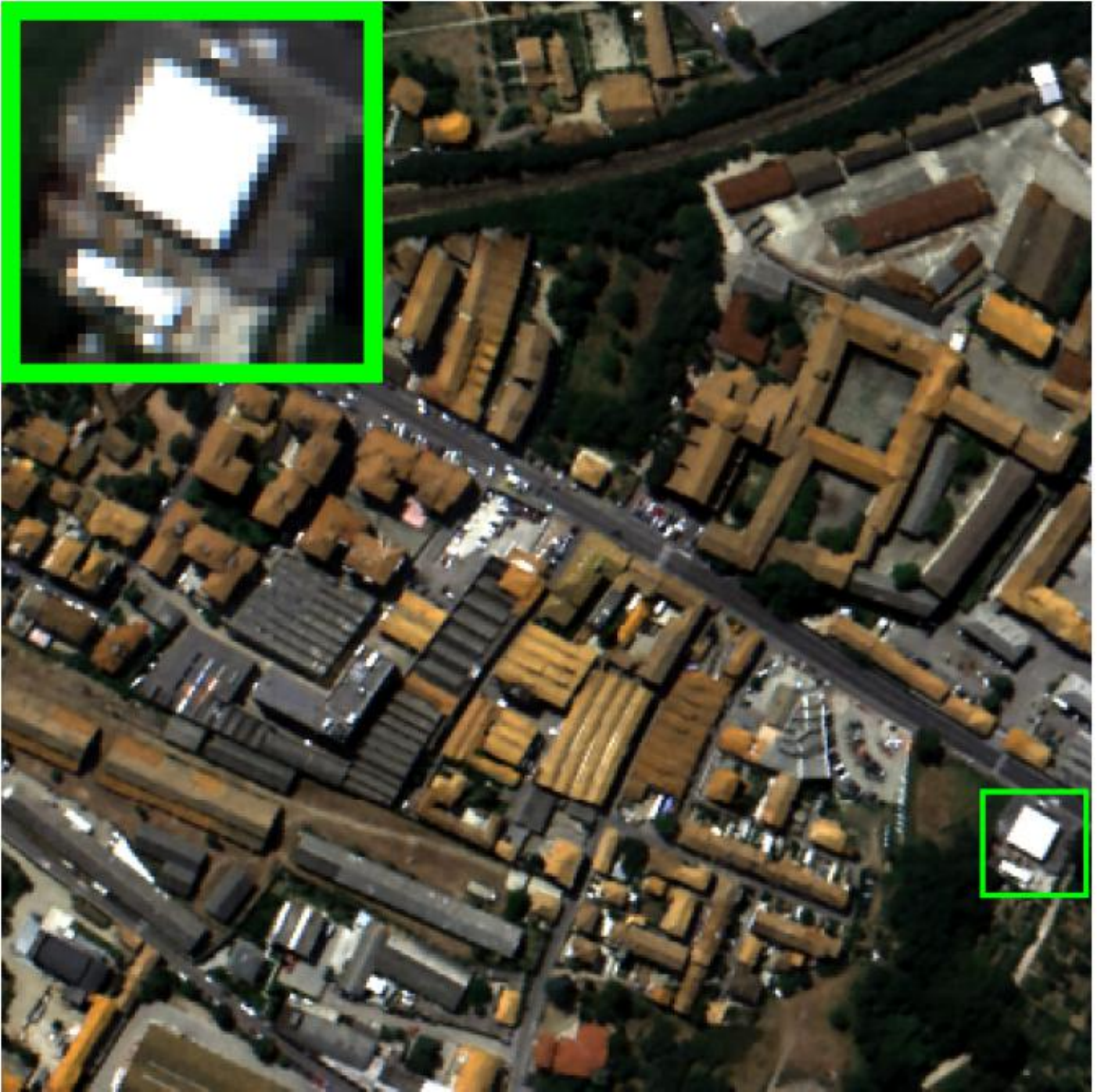}}
				{\includegraphics[width=1\linewidth]{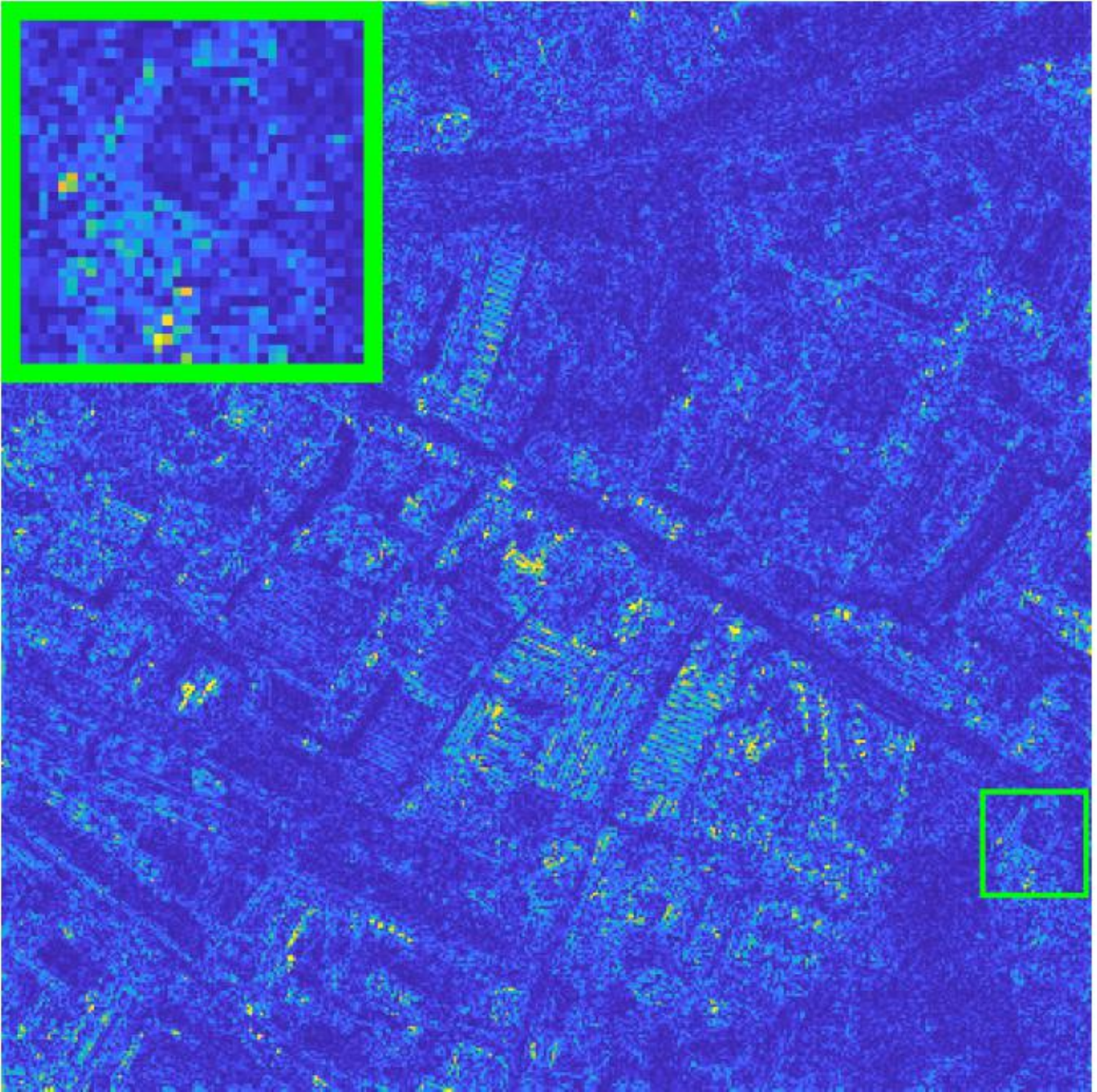}}
				\vspace{2pt}
				\scriptsize{Ada3D}
				\centering
				
			\end{minipage}
			\begin{minipage}[t]{0.086\linewidth}
				{\includegraphics[width=1\linewidth]{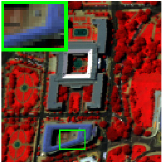}}
				\vspace{2pt}
				{\includegraphics[width=1\linewidth]{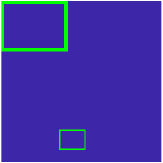}}
				{\includegraphics[width=1\linewidth]{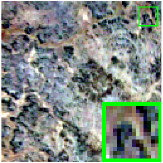}}
				\vspace{2pt}
				{\includegraphics[width=1\linewidth]{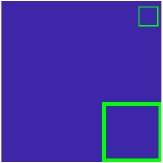}}
				{\includegraphics[width=1\linewidth]{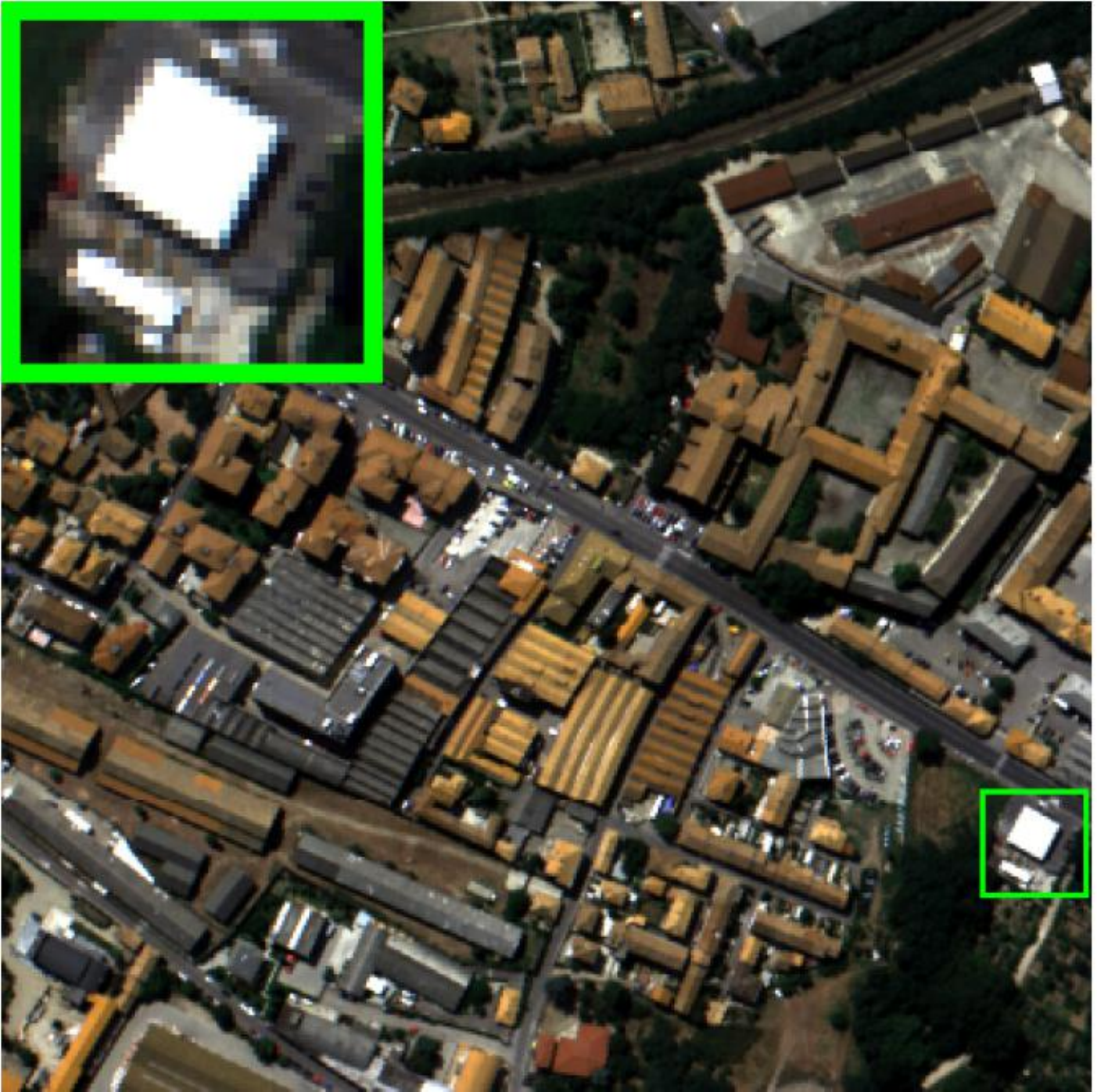}}
				{\includegraphics[width=1\linewidth]{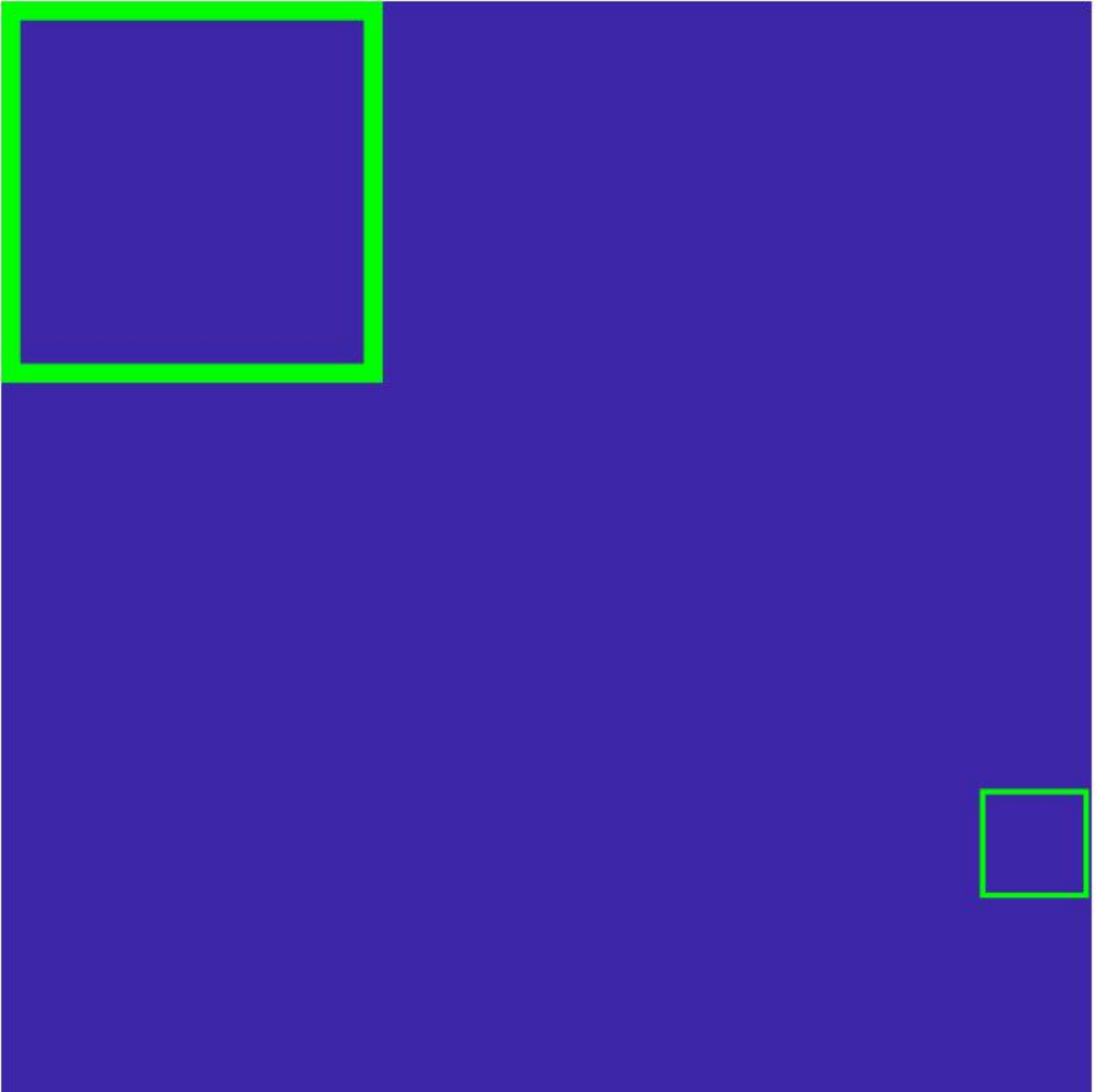}}
				\vspace{2pt}
				\scriptsize{GT}
				\centering
				
			\end{minipage}
		\end{minipage}
		\begin{minipage}[t]{0.98\linewidth}
			{\includegraphics[width=1\linewidth]{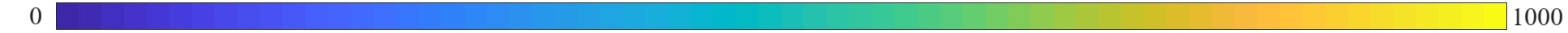}}
			\centering
		\end{minipage}
	\end{center}
    \vspace{-11pt}
	\caption{Qualitative evaluation results for hyper-spectral pansharpening on the WDC, Botswana, and Pavia datasets. Row 1: Pseudo-color images of spectral bands 20, 50, and 80 for a testing sample from the WDC dataset. Row 2: Absolute error maps (AEMs) of spectral band 25 for the testing sample in Row 1. Row 3: Pseudo-color images of spectral bands 30, 50, and 70 for a testing sample from the Botswana dataset. Row 4: AEMs of spectral band 34 for the testing sample in row 3. Row 5: Pseudo-color images of spectral bands 20, 40, and 60 for a testing sample from the Pavia dataset. Row 6: AEMs of spectral band 68 for the testing sample in row 5. The values on both sides of the color bar represent the degree of errors. A darker AEM signifies a better result. \label{hspp}}
\end{figure*}

\begin{figure*}[h]
	\begin{center}
		\begin{minipage}[t]{0.94\linewidth}
			\begin{minipage}[t]{0.33\linewidth}
				{\includegraphics[width=1\linewidth]{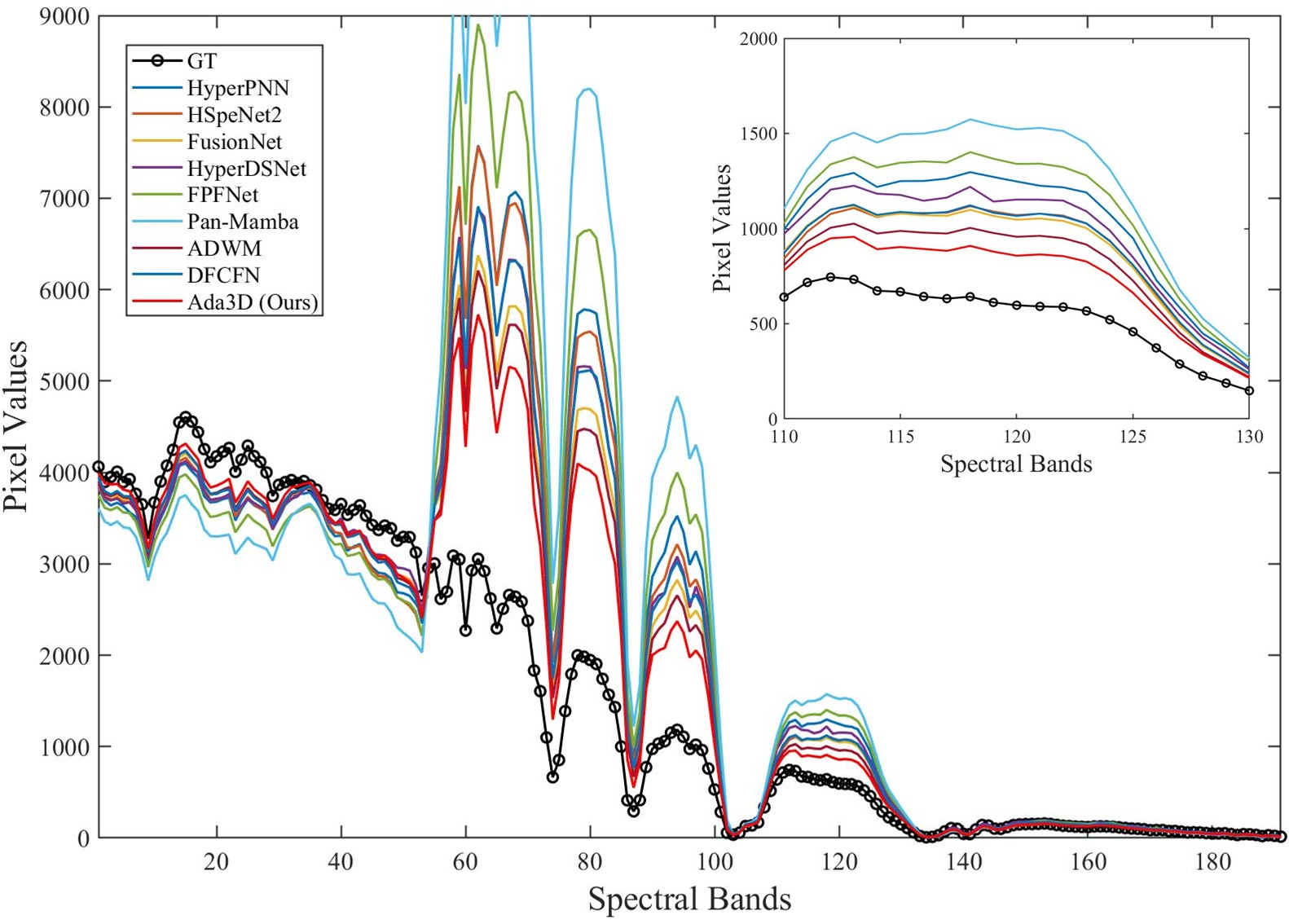}}
				\small{Spectral Vectors at $(1, 94)$}
				\centering
				
			\end{minipage}
			\begin{minipage}[t]{0.33\linewidth}
				{\includegraphics[width=1\linewidth]{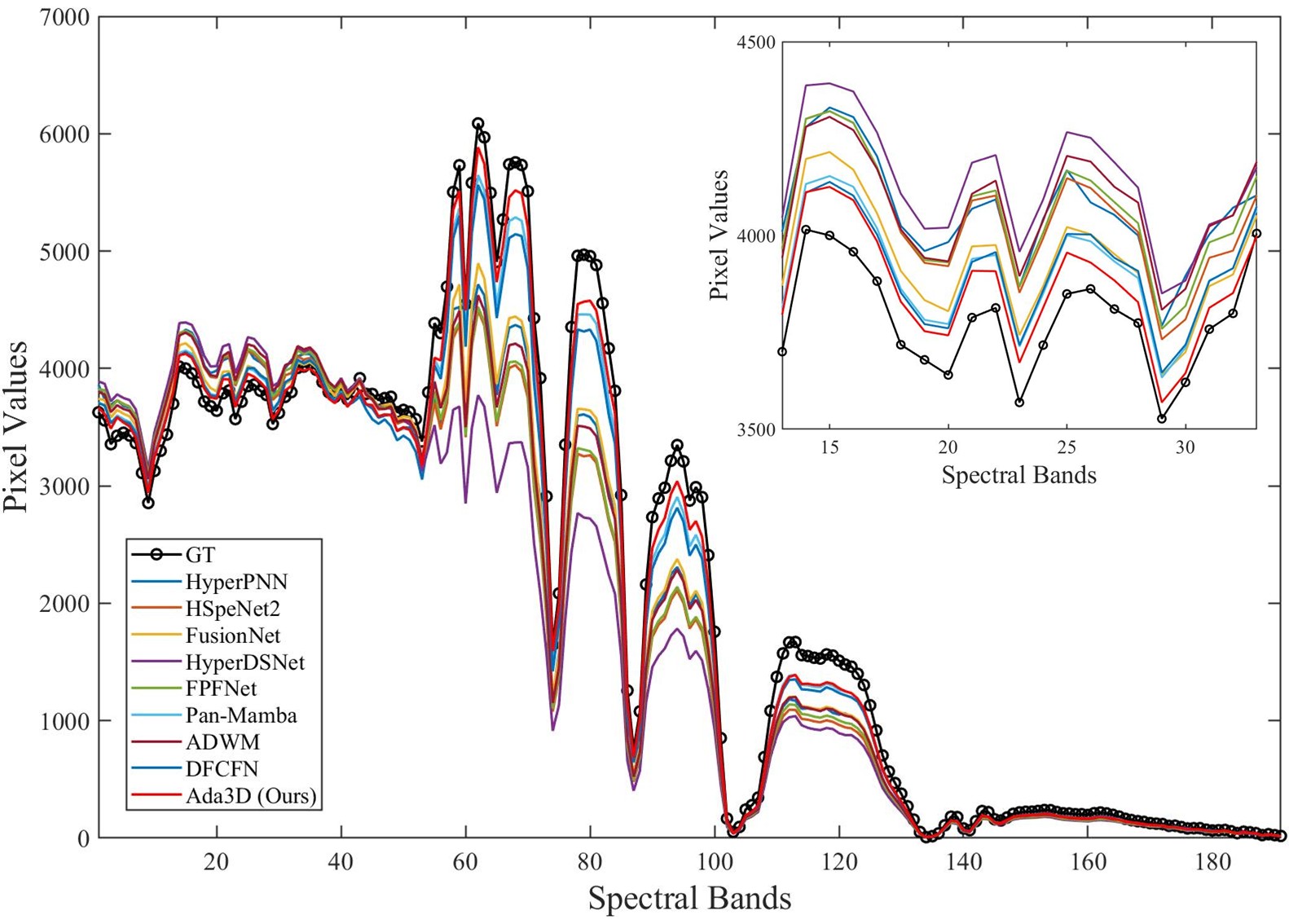}}
				\small{Spectral Vectors at $(15, 6)$}
				\centering
				
			\end{minipage}
			\begin{minipage}[t]{0.33\linewidth}
				{\includegraphics[width=1\linewidth]{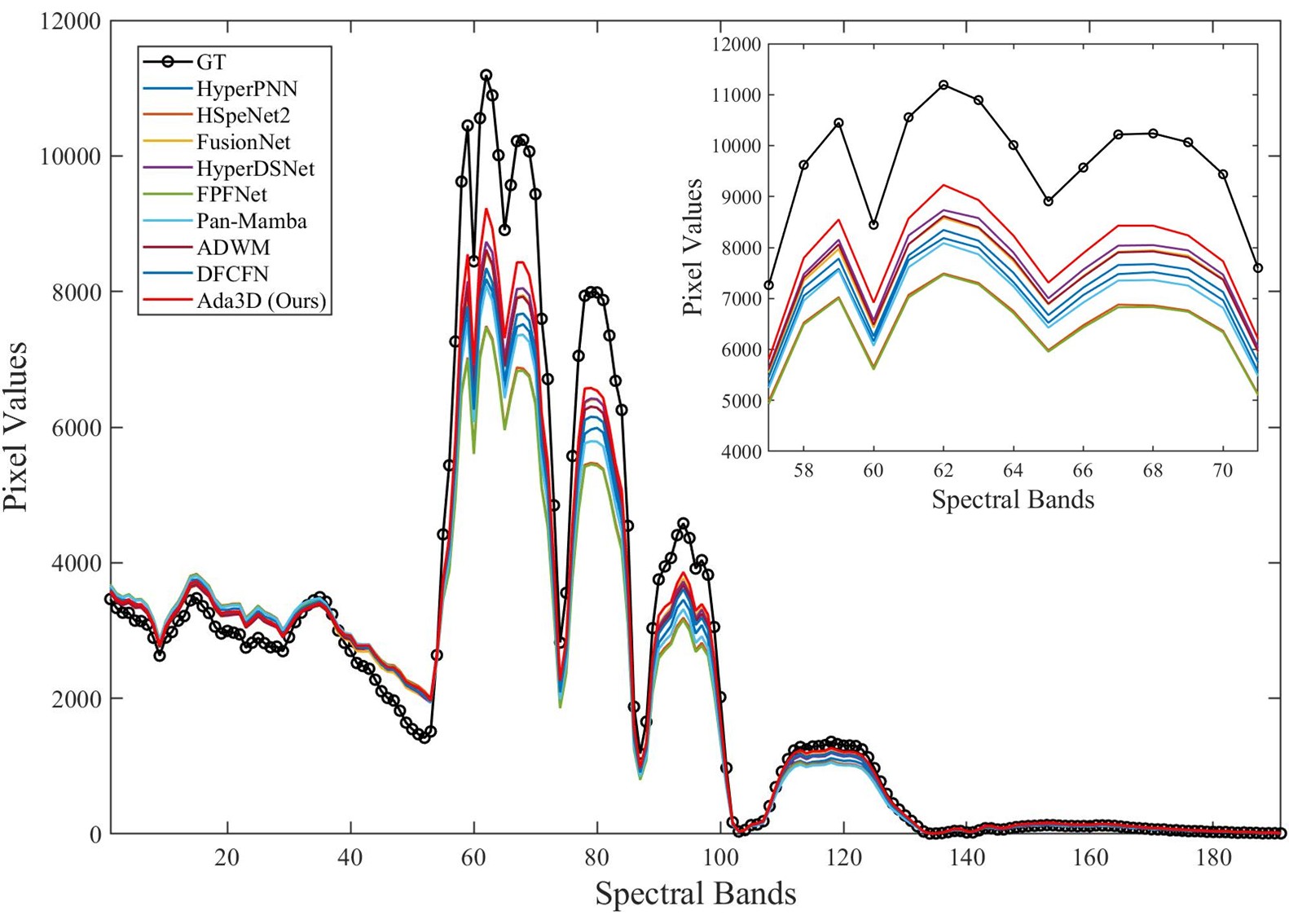}}
				\small{Spectral Vectors at $(33, 94)$}
				\centering
				
			\end{minipage}
		\end{minipage}
	\end{center}
    \vspace{-8pt}
	\caption{Comparison of spectral vectors at three randomly selected spatial locations from a testing sample in the WDC dataset.\label{wdccc}}
\end{figure*}

\subsection{Hyper-Spectral Pansharpening}
\subsubsection{Datasets}
We conduct experiments using three widely recognized hyper-spectral pansharpening datasets \cite{7284770}: Washington D.C. (WDC), Botswana, and Pavia. The WDC dataset consists of images captured by the HYper-spectral Digital Imagery Collection Experiment (HY-DICE) sensor, which collects data across 210 spectral bands, covering wavelengths from 0.4 to 2.4 $\upmu$m, with a spatial resolution of 1 meter. The Botswana dataset comprises images acquired by the Hyperion sensor aboard the Earth Observing-1 (EO-1) satellite, which is operated by the National Aeronautics and Space Administration (NASA). This sensor records data over 242 spectral bands, spanning wavelengths from 0.4 to 2.5 $\upmu$m, with a spatial resolution of 30 meters. The Pavia dataset contains images captured by the Reflective Optics System Imaging Spectrometer (ROSIS) sensor, which gathers data across 115 spectral bands, covering wavelengths from 0.4 to 0.9 $\upmu$m, with a spatial resolution of 1.3 meters. The hyper-spectral pansharpening datasets utilized in this study are sourced from the HyperPanCollection repository\footnote{\url{https://github.com/liangjiandeng/HyperPanCollection}}. These datasets are generated following Wald's protocol \cite{1997Fusion}, with a detailed explanation of the procedure provided in \cite{9870551}. Specifically, the WDC dataset consists of 1024 training samples, with 90\% designated for direct training purposes and 10\% for validation purposes. Additionally, it includes four testing samples. Each training sample is an image triplet in the PAN/LRHS/GT format, with sizes of $64\times64$, $16\times16\times191$, and $64\times64\times191$, respectively. The testing samples are also in the PAN/LRHS/GT format, but with larger dimensions of $128\times128$, $32\times32\times191$, and $128\times128\times191$, respectively.
The Botswana dataset contains 967 training samples, with 83\% allocated for direct training purposes and 17\% for validation purposes. Additionally, there are four separate testing samples. Each training sample comprises a PAN/LRHS/GT image triplet with dimensions of $64\times64$, $16\times16\times145$, and $64\times64\times145$, respectively. The testing samples consist of PAN/LRHS/GT image triplets sized $128\times128$, $32\times32\times145$, and $128\times128\times145$. 
In the Pavia dataset, there are 1680 training samples, divided into 90\% for direct training purposes and 10\% for validation purposes. Additionally, this dataset includes four testing samples. Each training sample contains a PAN/LRHS/GT image triplet of sizes $64\times64$, $16\times16\times102$, and $64\times64\times102$, respectively. The testing samples comprise PAN/LRHS/GT image triplets with sizes of $400\times400$, $100\times100\times102$, and $400\times400\times102$.

\subsubsection{Benchmarks}
Ada3D is compared against several representative techniques, including three traditional approaches: GLP \cite{aiazzi2006mtf}, CNMF \cite{6049465}, and Hysure \cite{7000523}; as well as eight DL-based methods: HyperPNN \cite{8731649}, the HSpeNet series \cite{9200718}, FusionNet \cite{2020Detail}, HyperDSNet \cite{9870551}, FPFNet \cite{10298274}, Pan-Mamba \cite{he2024pan}, ADWM \cite{Huang_2025_CVPR}, and DFCFN \cite{10835747}. For a fair comparison, all DL-based methods are trained under identical conditions, utilizing the same Nvidia 3090 GPU and PyTorch framework.

\begin{table*}[t]	
	\centering\renewcommand\arraystretch{1.}\setlength{\tabcolsep}{5.3pt}
	\belowrulesep=0pt\aboverulesep=0pt
	\caption{Quantitative evaluation results on 20 reduced-resolution and 20 full-resolution samples from the WV3 dataset, which belongs to the pansharpening task. The symbol $\ast$ represents DL-based methods that utilize adaptive 2D convolutions.}
	\label{rr}	
	\begin{tabular}{l|c|cccc|ccc}
		\toprule
		\multirow{2}{*}{\textbf{Methods}} & 
		\multirow{2}{*}{\textbf{Params}} & 
		\multicolumn{4}{c|}{\textbf{Reduced-Resolution}} & \multicolumn{3}{c}{\textbf{Full-Resolution (Real Data)}}\\
		\cmidrule(lr){3-6}\cmidrule(lr){7-9}
		&\multicolumn{1}{c|}{} 
		&\multicolumn{1}{c}{PSNR($\pm$std)} 
		&\multicolumn{1}{c}{Q2n($\pm$std)} 
		&\multicolumn{1}{c}{SAM($\pm$std)} 
		&\multicolumn{1}{c|}{ERGAS($\pm$std)} 
		&\multicolumn{1}{c}{${{\rm{D}}_{\rm{\lambda}}}$($\pm$std)} 
		&\multicolumn{1}{c}{${{\rm{D}}_{\rm{s}}}$($\pm$std)} 
		&\multicolumn{1}{c}{QNR($\pm$std)} \\
		\midrule
		\textbf{TV} \cite{palsson2013new} & $-$ & 32.381$\pm$2.328 & 0.795$\pm$0.120 & 5.692$\pm$1.808 & 4.855$\pm$1.434
		& 0.0234$\pm$0.0061  & 0.0393$\pm$0.0227 & 0.9383$\pm$0.0269\\ 
		\textbf{GLP-HPM} \cite{6616569} & $-$ & 33.095$\pm$2.800 & 0.835$\pm$0.092 & 5.333$\pm$1.761 & 4.616$\pm$1.503
		& 0.0206$\pm$0.0082  & 0.0630$\pm$0.0284 & 0.9180$\pm$0.0346\\ 
		\textbf{BDSD-PC} \cite{2019Robust} & $-$ & 32.970$\pm$2.784 & 0.829$\pm$0.097 & 5.428$\pm$1.822 & 4.697$\pm$1.617
		& 0.0625$\pm$0.0235  & 0.0730$\pm$0.0356 & 0.8698$\pm$0.0531\\ 
		\midrule
		\textbf{PNN} \cite{2016Pansharpening} & 0.10M & 37.313$\pm$2.646 & 0.893$\pm$0.092 & 3.677$\pm$0.762 & 2.681$\pm$0.647
		& 0.0213$\pm$0.0080  & 0.0428$\pm$0.0147 & 0.9369$\pm$0.0212\\ 
		\textbf{MSDCNN} \cite{8127731} & 0.23M & 37.068$\pm$2.686 & 0.890$\pm$0.090 & 3.777$\pm$0.803 & 2.760$\pm$0.689
		& 0.0230$\pm$0.0091 & 0.0467$\pm$0.0199 & 0.9316$\pm$0.0271\\
		\textbf{BDPN} \cite{zhang2019pan} & 1.49M & 36.191$\pm$2.702 & 0.871$\pm$0.100 & 4.201$\pm$0.857 & 3.046$\pm$0.732
		& 0.0364$\pm$0.0142  & 0.0459$\pm$0.0192 & 0.9196$\pm$0.0308\\ 
		\textbf{FusionNet} \cite{2020Detail} & 0.08M & 38.047$\pm$2.589 & 0.904$\pm$0.090 & 3.324$\pm$0.698 & 2.465$\pm$0.644
		& 0.0239$\pm$0.0090  & 0.0364$\pm$0.0137 & 0.9406$\pm$0.0197\\   
		\textbf{MUCNN} \cite{10.1145/3474085.3475600} & 2.32M & 38.262$\pm$2.703 & 0.911$\pm$0.089 & 3.206$\pm$0.681 & 2.400$\pm$0.617 
		& 0.0258$\pm$0.0111  & {0.0327}$\pm$0.0140 & {0.9424}$\pm$0.0205\\   
		\textbf{LAGNet}\textsuperscript{$\ast$} \cite{jin2022aaai}& 0.15M & 38.592$\pm$2.778 & 0.910$\pm$0.091 & 3.103$\pm$0.558 & {2.292}$\pm$0.607
		& 0.0368$\pm$0.0148  & 0.0418$\pm$0.0152& 0.9230$\pm$0.0247\\  
		\textbf{PMACNet} \cite{9764690} & 0.94M & {38.595}$\pm$2.882 & {0.912}$\pm$0.092 & {3.073}$\pm$0.623 & 2.293$\pm$0.532
		& 0.0540$\pm$0.0232  & 0.0336$\pm$0.0115 & 0.9143$\pm$0.0281\\  	
		\textbf{U2Net} \cite{10.1145/3581783.3612084}& 0.66M & {39.117}$\pm$3.009 & \underline{0.920}$\pm$0.085 & \underline{2.888}$\pm$0.581 & \underline{2.149}$\pm$0.525
		& \underline{0.0178}$\pm$0.0072  & {0.0313}$\pm$0.0075 & {0.9514}$\pm$0.0115\\   
		\textbf{Pan-Mamba} \cite{he2024pan} & 0.48M & {39.012}$\pm$2.986 & \underline{0.920}$\pm$0.085 & {2.914}$\pm$0.592 & 2.184$\pm$0.521
		& 0.0183$\pm$0.0071  & 0.0307$\pm$0.0108 & \underline{0.9516}$\pm$0.0146\\ 
		\textbf{CANNet}\textsuperscript{$\ast$} \cite{Duan_2024_CVPR} & 0.78M & {39.003}$\pm$2.900 & {0.919}$\pm$0.084 & {2.941}$\pm$0.590 & 2.175$\pm$0.530
		& 0.0196$\pm$0.0083  & {0.0301}$\pm$0.0074 & 0.9510$\pm$0.0126\\ 
	\textbf{ADWM} \cite{Huang_2025_CVPR} & 0.40M & \underline{39.137}$\pm$2.894 & {0.919}$\pm$0.086 & {2.927}$\pm$0.589 & 2.154$\pm$0.539 & 0.0261$\pm$0.0105 & \underline{0.0286}$\pm$0.0135 & 0.9462$\pm$0.0191 \\
		\textbf{Ada3D (Ours)} & 1.68M & \textbf{39.327}$\pm$2.871 & \textbf{0.922}$\pm$0.085 & \textbf{2.853}$\pm$0.569 & \textbf{2.099}$\pm$0.512 
		& \textbf{0.0168}$\pm$0.0079  & \textbf{0.0282}$\pm$0.0061 & \textbf{0.9554}$\pm$0.0111\\      
		\midrule
		\textbf{Ideal Values} 
		& $-$
		&\multicolumn{1}{c}{\textbf{+$\infty$}}
		&\multicolumn{1}{c}{\textbf{\textbf{1}}}
		&\multicolumn{1}{c}{\textbf{\textbf{0}}}
		&\multicolumn{1}{c|}{\textbf{\textbf{0}}}
		&\multicolumn{1}{c}{\textbf{\textbf{0}}}
		&\multicolumn{1}{c}{\textbf{\textbf{0}}}
		&\multicolumn{1}{c}{\textbf{\textbf{1}}}
		\\ 
		\bottomrule
	\end{tabular}
\end{table*}

\subsubsection{Quality Indices}
To align with the research standards for hyper-spectral pansharpening, we select five widely recognized quality indices for evaluation, namely PSNR \cite{1284395}, cross-correlation (CC), SSIM \cite{1284395}, SAM \cite{yuhas1992discrimination}, and ERGAS \cite{1997Fusion}. The ideal values for these indices are +$\infty$, 1, 1, 0, and 0.

\subsubsection{Settings}
This paper initially assumes an equal number of channels for spatial and spectral feature maps to simplify the description. However, in practice, we allocate different channel counts to enhance the flexibility of network design. Specifically, for the hyper-spectral pansharpening task, we configure the spatial and spectral feature maps with 32 and 8 channels, respectively. The values of $k$, $\alpha$, and $\beta$ are set to 3, 1, and 1, respectively. Additionally, we use Bicubic interpolation for up-sampling. Furthermore, we apply three ResBlocks for spatial feature extraction and employ six Ada3D blocks to effectively integrate spatial and spectral information. All learnable weights are initialized using the Kaiming uniform distribution \cite{He_2015_ICCV}. During the training of our networks on the WDC, Botswana, and Pavia datasets, we set the number of epochs to 1200, 2000, and 2000. Additionally, the batch size and initial learning rate are consistently set to 4 and $5\times 10^{-4}$ across all datasets. Furthermore, we utilize the Adam optimizer, with the learning rate halved every 250 epochs. 

\subsubsection{Results}
The quantitative evaluation results for the three distinct datasets are summarized in Table \ref{hsp}. Evidently, our method significantly outperforms other techniques across all quality indices. In addition to these results, the qualitative evaluation in Fig.~\ref{hspp} shows that Ada3D produces the darkest absolute error maps (AEMs). Moreover, Fig.~\ref{wdccc} presents spectral vectors from three different spatial locations in a WDC testing sample, demonstrating the minimal spectral distortions achieved by our method. These findings highlight Ada3D's superiority in the hyper-spectral pansharpening task.

\subsection{Pansharpening}
\subsubsection{Datasets}
For the pansharpening task \cite{9245579}, we perform experiments using the popular WorldView-3 (WV3) dataset, which contains imagery captured by the sensor aboard the WV3 satellite. This sensor collects data across eight spectral bands, spanning wavelengths from 0.4 to 1 ${\upmu}$m, with a spatial resolution of 1.2 meters. The dataset employed in this study is sourced from the PanCollection\footnote{\url{https://github.com/liangjiandeng/PanCollection}}, and the data generation process strictly follows Wald's protocol \cite{1997Fusion}. A comprehensive description of this process can be found in \cite{9844267}. Specifically, the WV3 dataset consists of 10000 training samples, with 90\% designated for direct training purposes and 10\% for validation purposes. It also includes 20 reduced-resolution and 20 full-resolution testing samples. Each training sample comprises an image triplet in the PAN/LRMS/GT format, with sizes of $64\times64$, $16\times16\times8$, and $64\times64\times8$, respectively. The reduced-resolution testing samples consist of PAN/LRMS/GT image triplets sized $256\times256$, $64\times64\times8$, and $256\times256\times8$, respectively. In addition, the full-resolution testing samples contain image pairs in the PAN/LRMS format, with dimensions of $512\times512$ and $128\times128\times8$, respectively.

\begin{figure*}[t]
	\begin{center}
		\begin{minipage}[t]{0.98\linewidth}
			\begin{minipage}[t]{0.12\linewidth}
				{\includegraphics[width=1\linewidth]{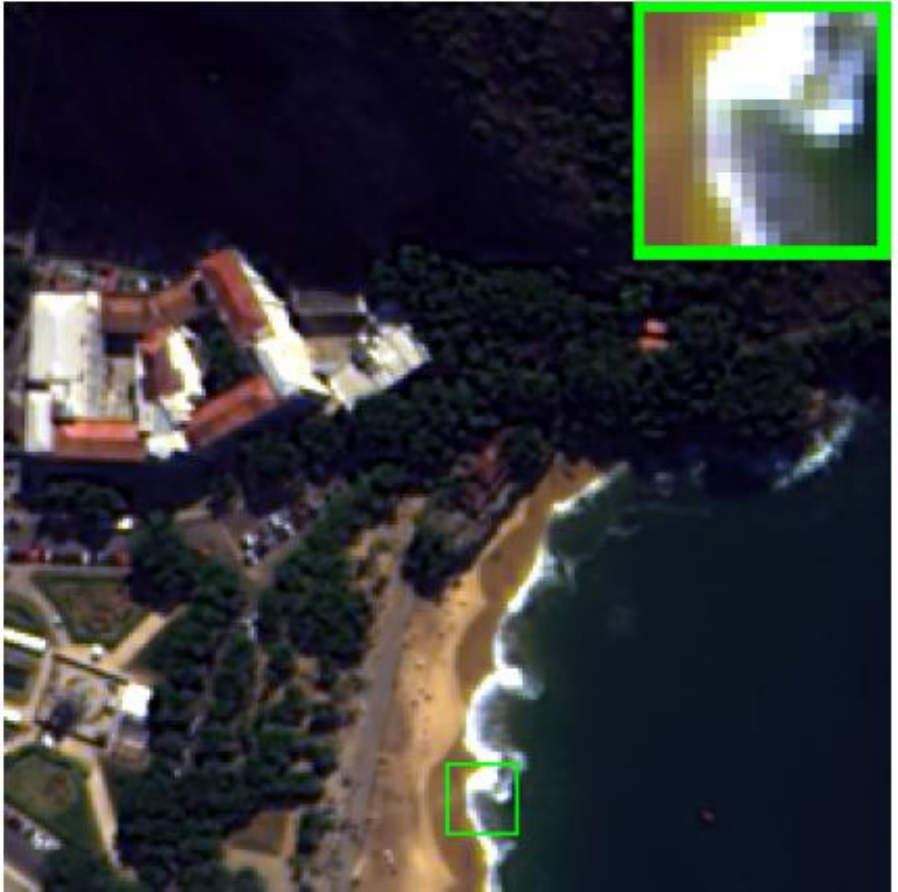}}
				{\includegraphics[width=1\linewidth]{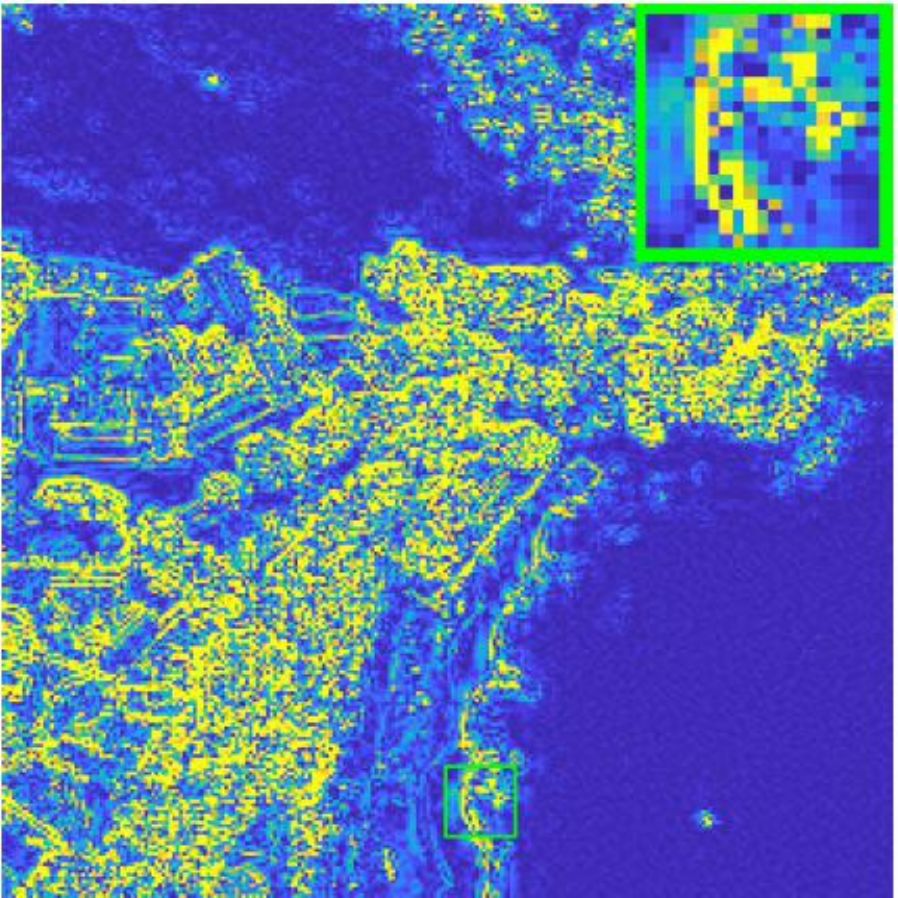}}
				\vspace{4pt}
				{TV}
				{\includegraphics[width=1\linewidth]{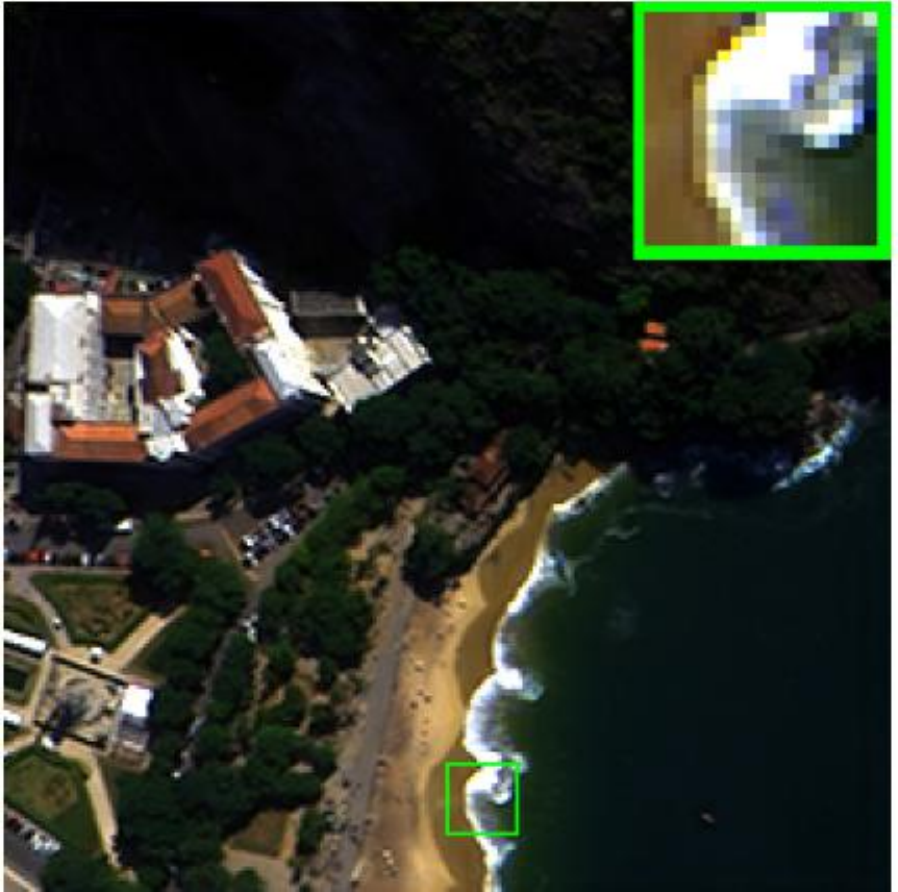}}
				{\includegraphics[width=1\linewidth]{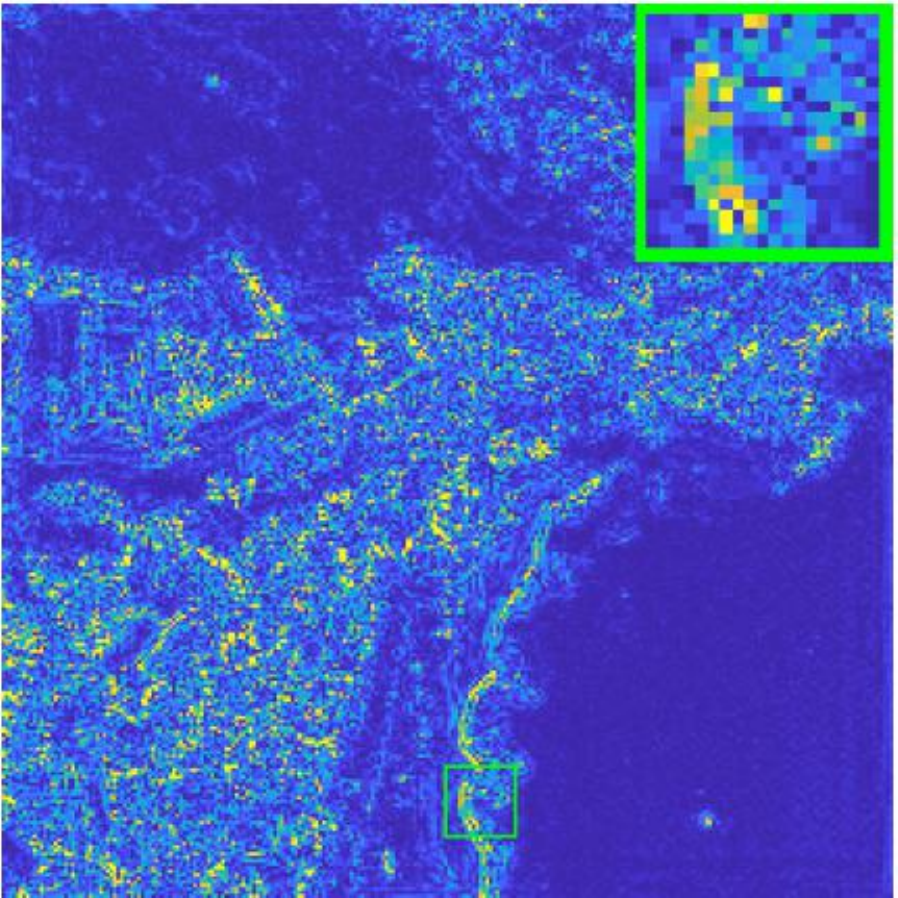}}
				\vspace{4pt}
				{LAGNet}
				\centering
				
			\end{minipage}
			\begin{minipage}[t]{0.12\linewidth}
				{\includegraphics[width=1\linewidth]{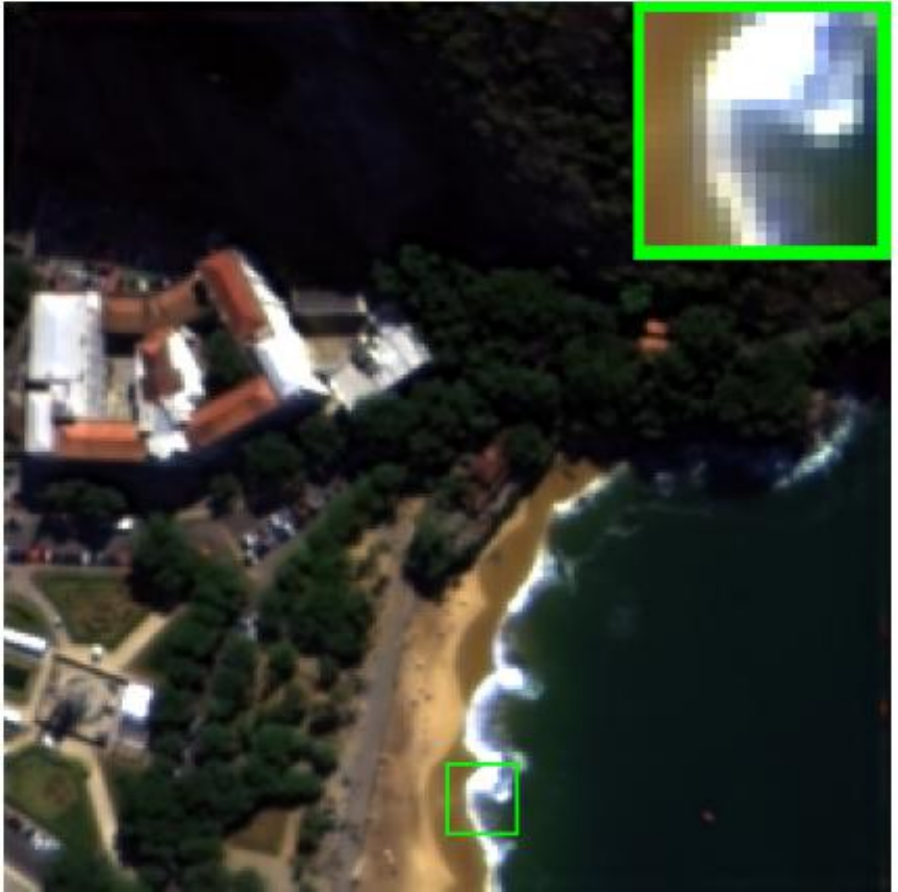}}
				{\includegraphics[width=1\linewidth]{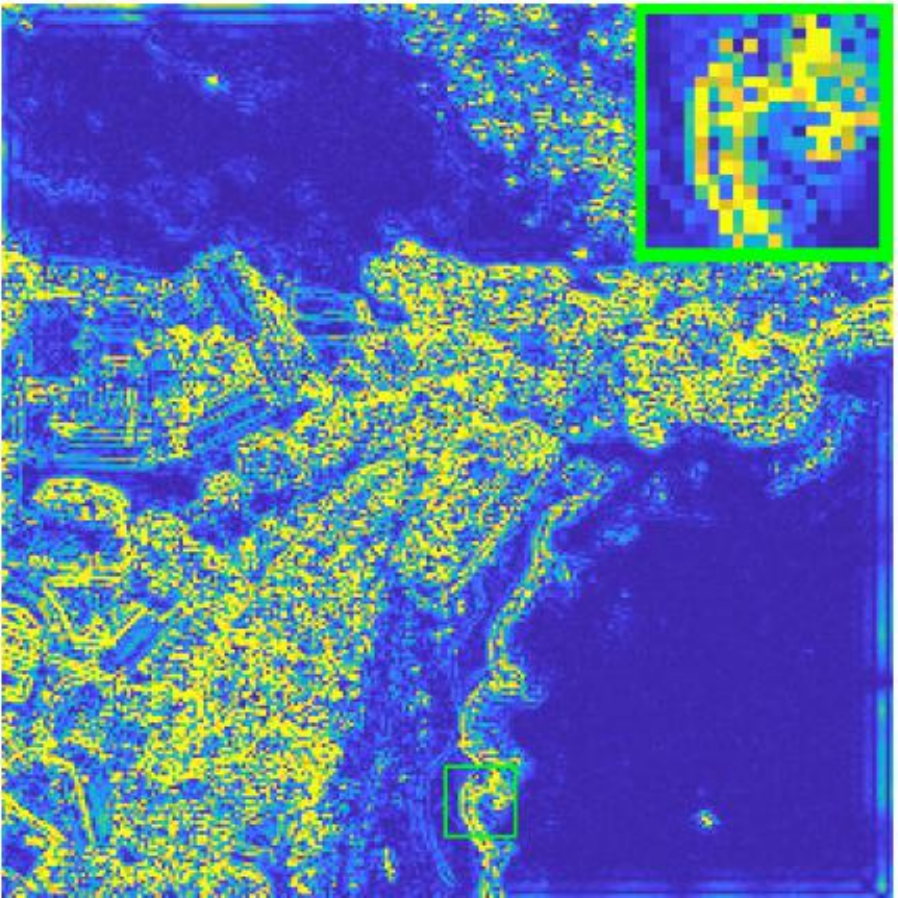}}
				\vspace{4pt}
				{GLP-HPM}
				{\includegraphics[width=1\linewidth]{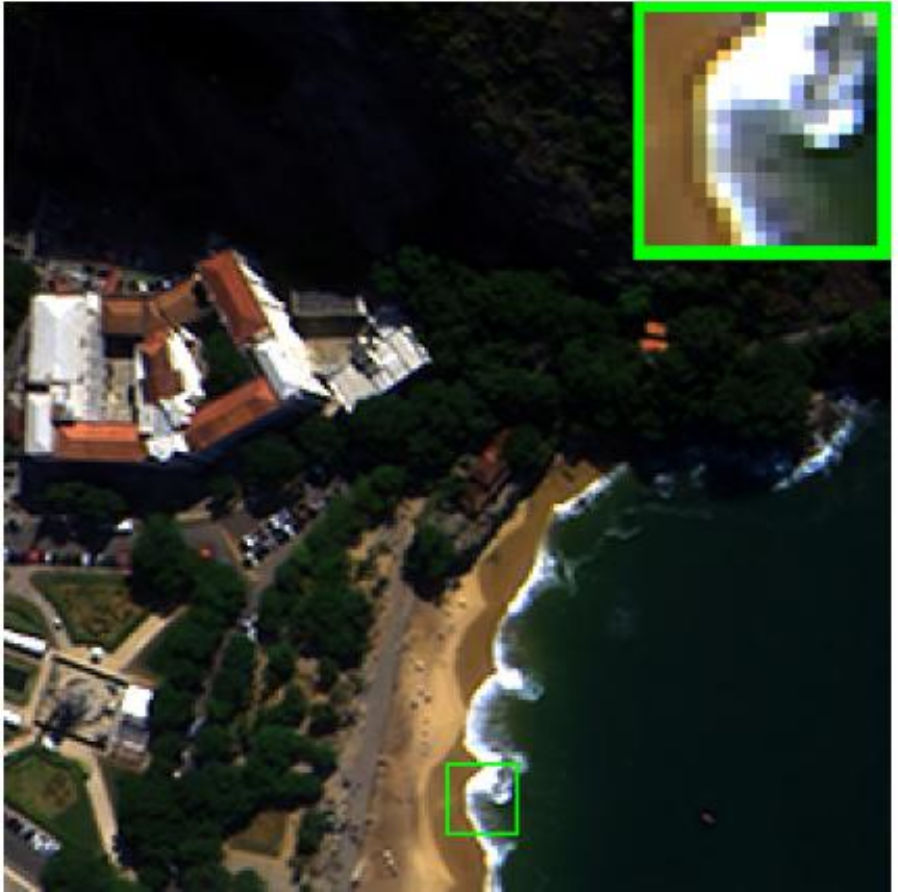}}
				{\includegraphics[width=1\linewidth]{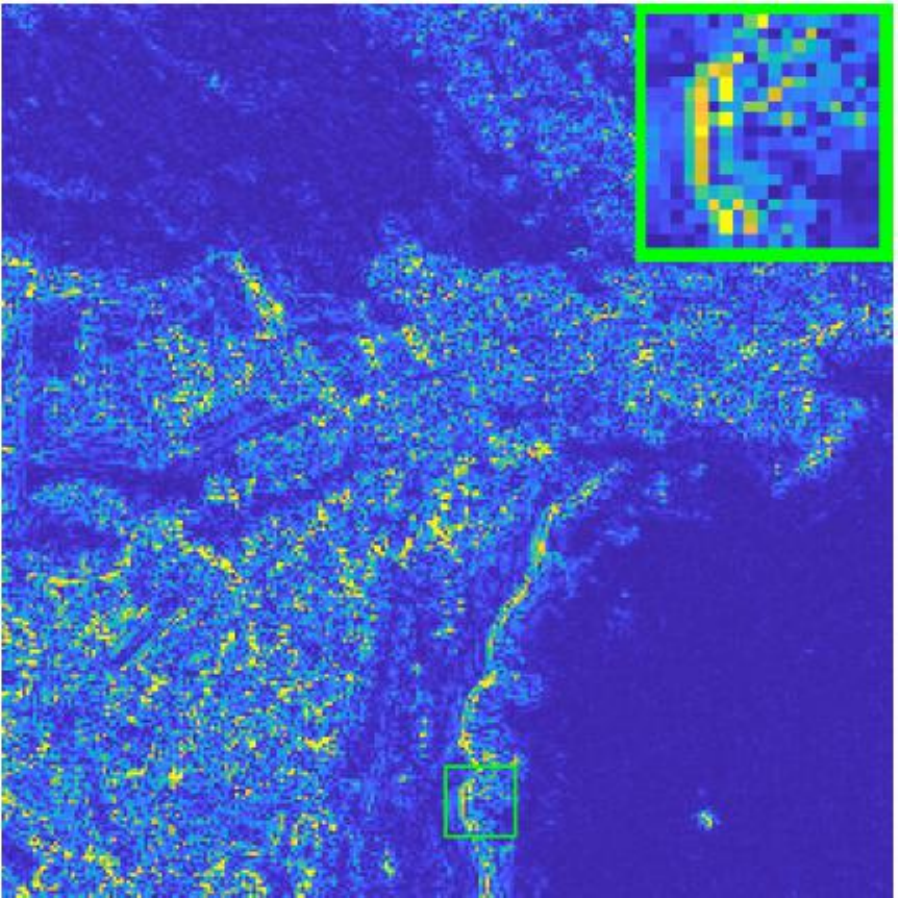}}
				\vspace{4pt}
				{PMACNet}
				\centering
				
			\end{minipage}
			\begin{minipage}[t]{0.12\linewidth}
				{\includegraphics[width=1\linewidth]{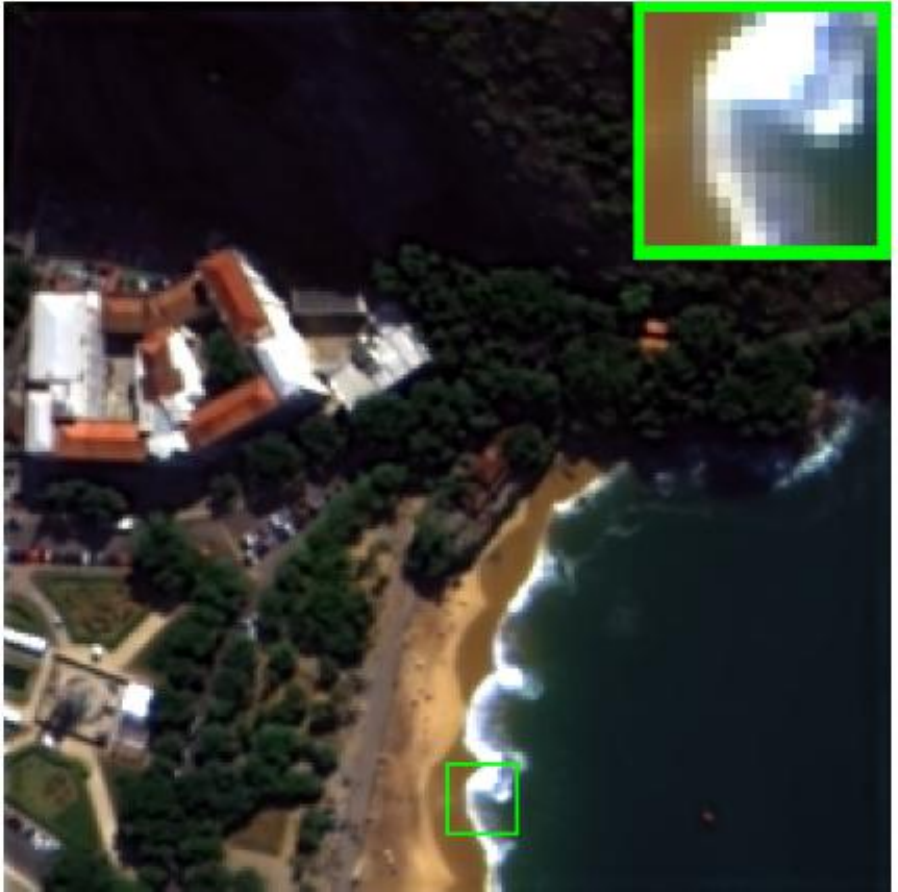}}
				{\includegraphics[width=1\linewidth]{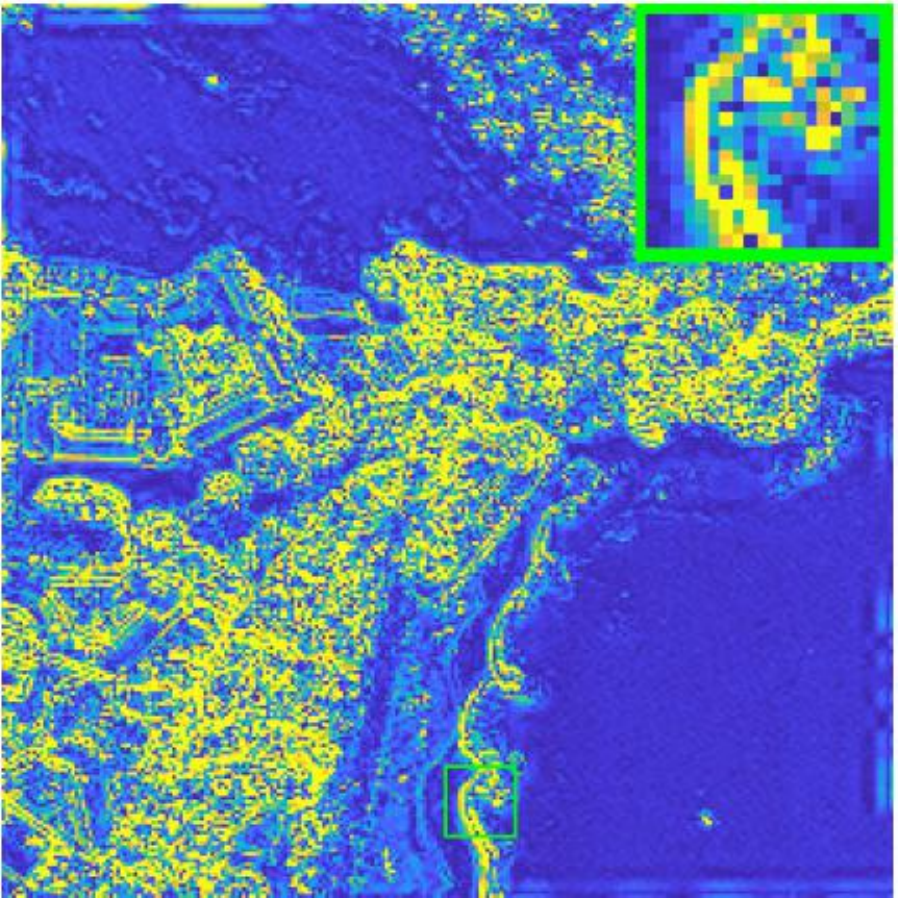}}
				\vspace{4pt}
				{BDSD-PC}
				{\includegraphics[width=1\linewidth]{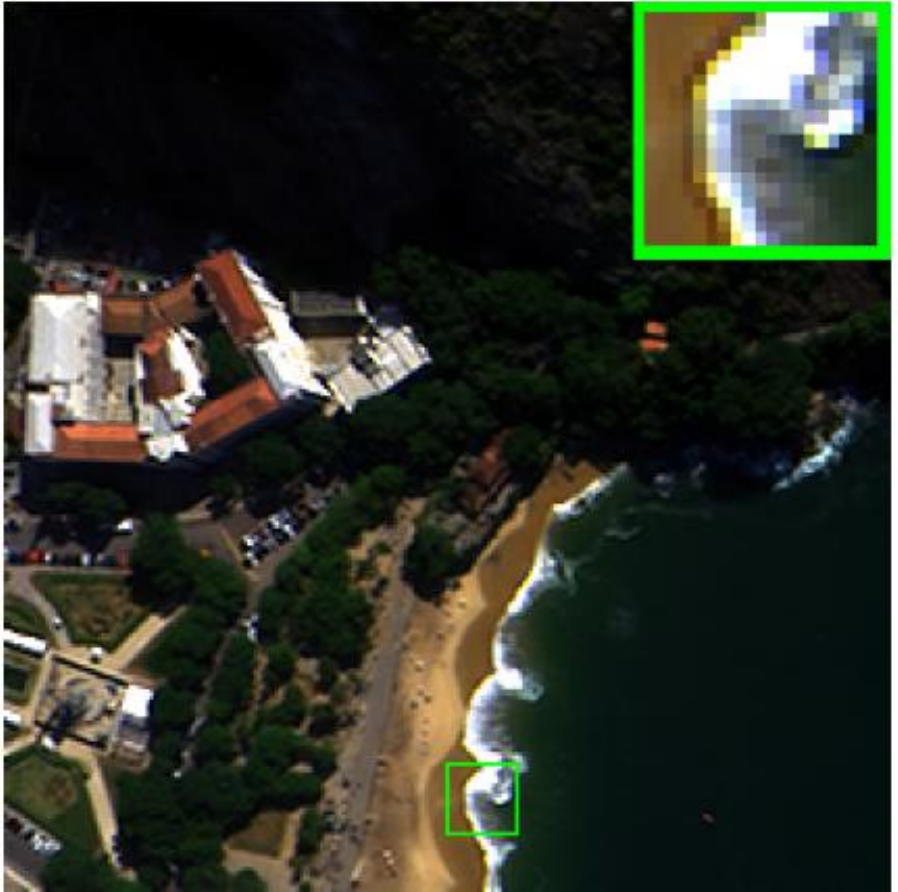}}
				{\includegraphics[width=1\linewidth]{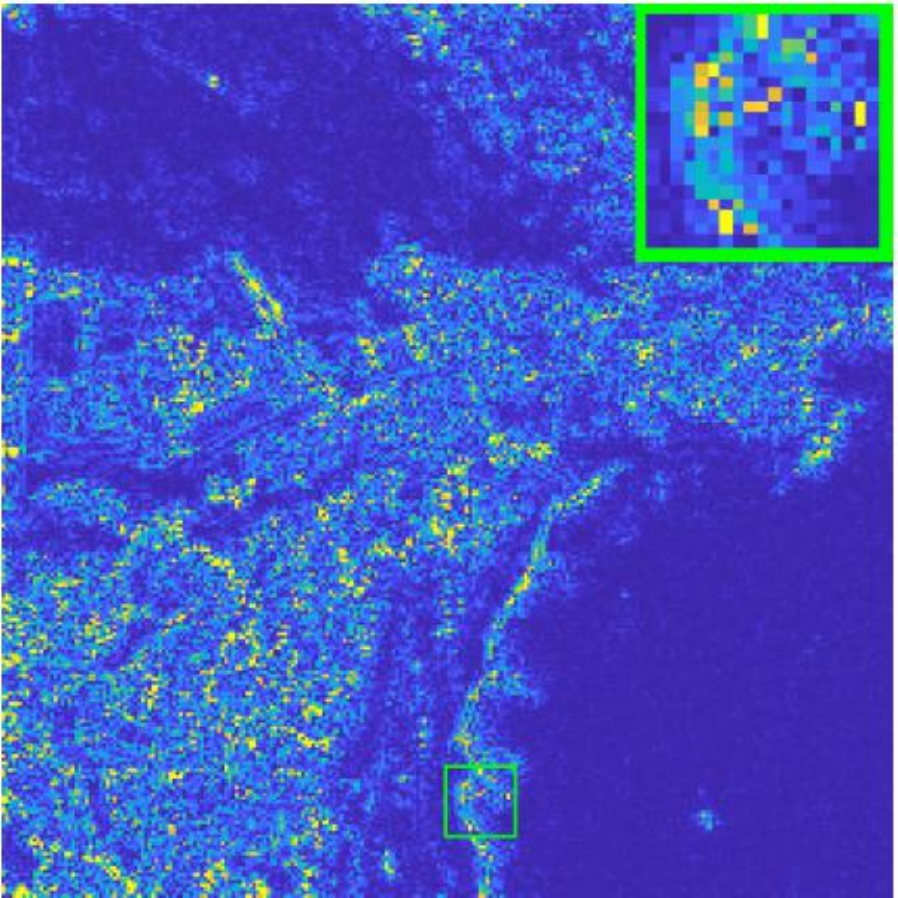}}
				\vspace{4pt}
				{U2Net}
				\centering
				
			\end{minipage}
			\begin{minipage}[t]{0.12\linewidth}
				{\includegraphics[width=1\linewidth]{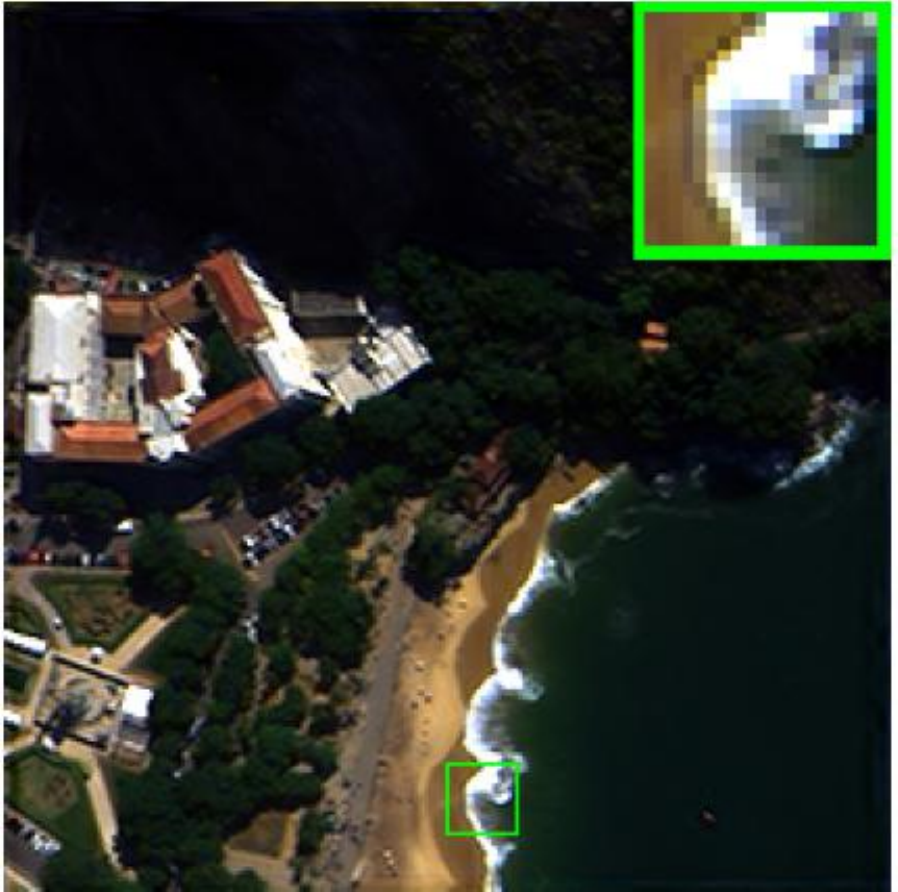}}
				{\includegraphics[width=1\linewidth]{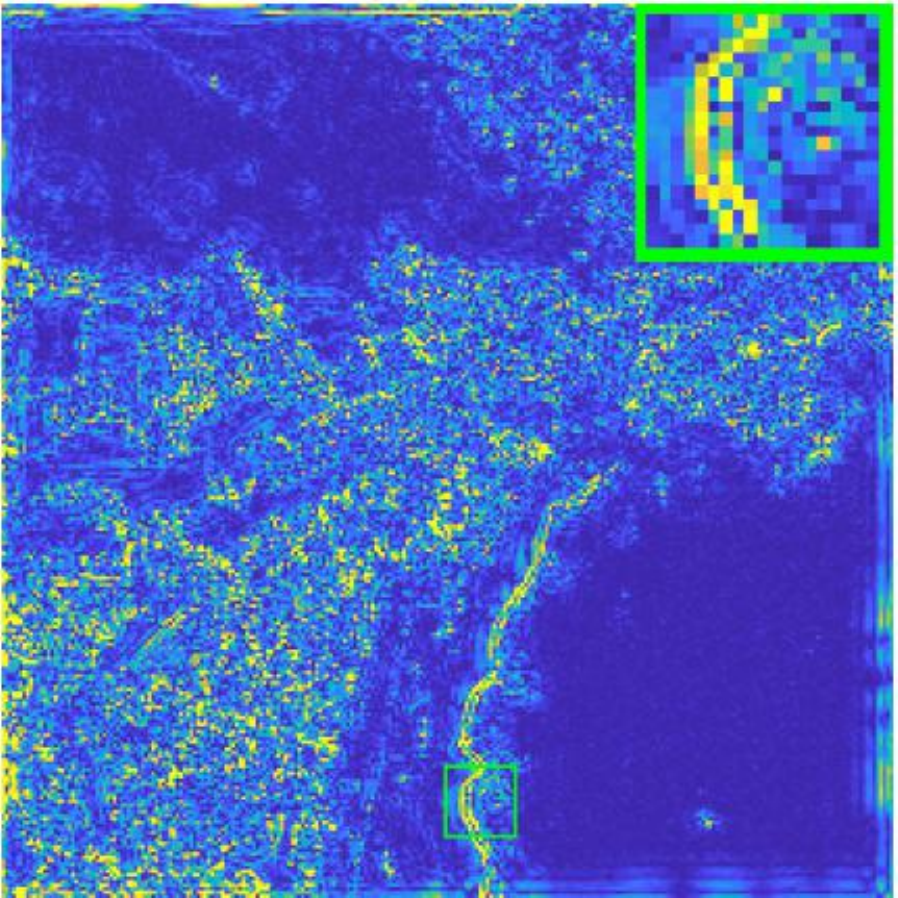}}
				\vspace{4pt}
				{PNN}
				{\includegraphics[width=1\linewidth]{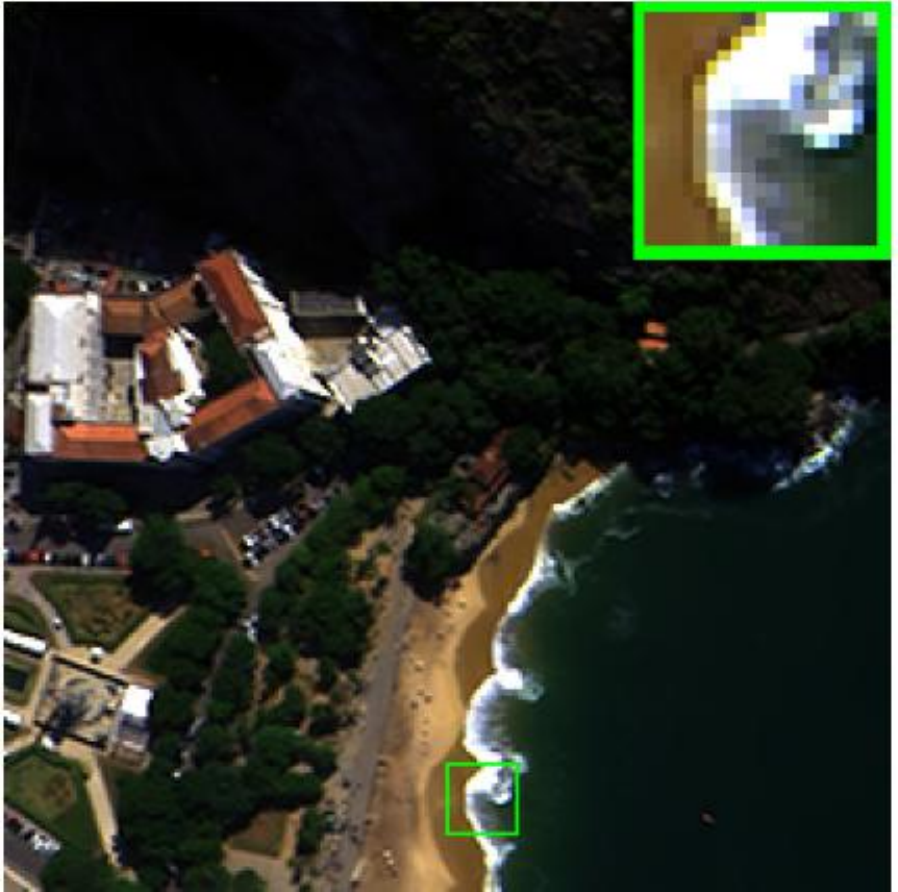}}
				{\includegraphics[width=1\linewidth]{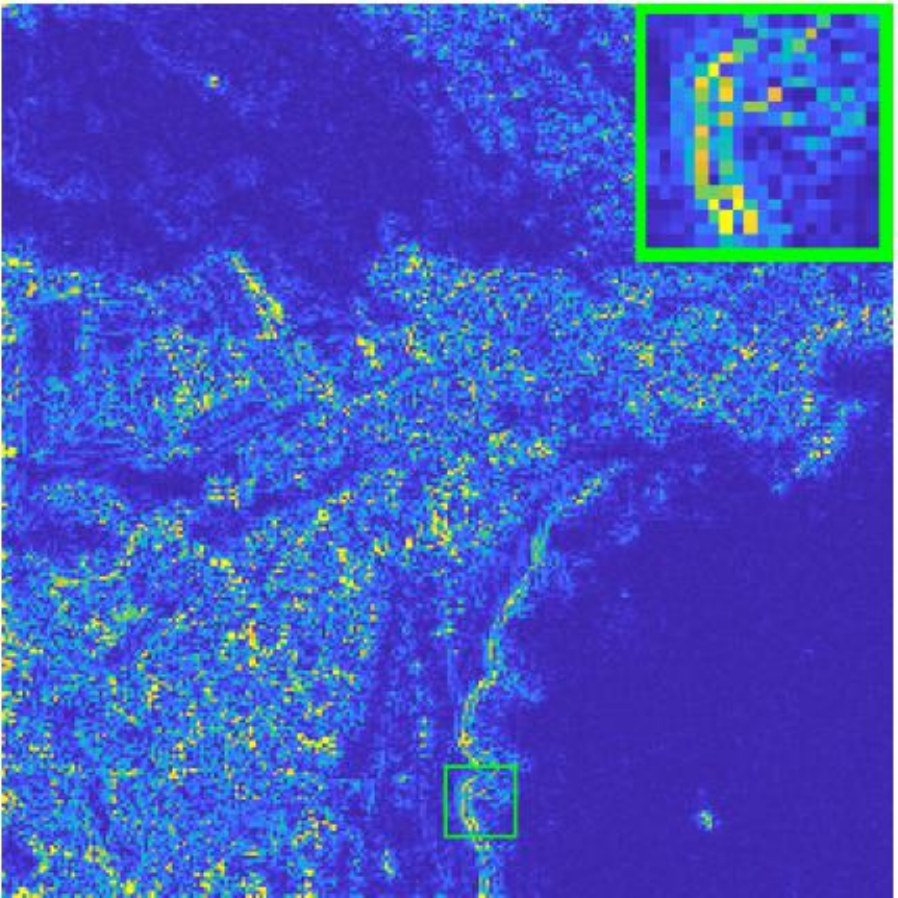}}
				\vspace{4pt}
				{Pan-Mamba}
				\centering
				
			\end{minipage}
			\begin{minipage}[t]{0.12\linewidth}
				{\includegraphics[width=1\linewidth]{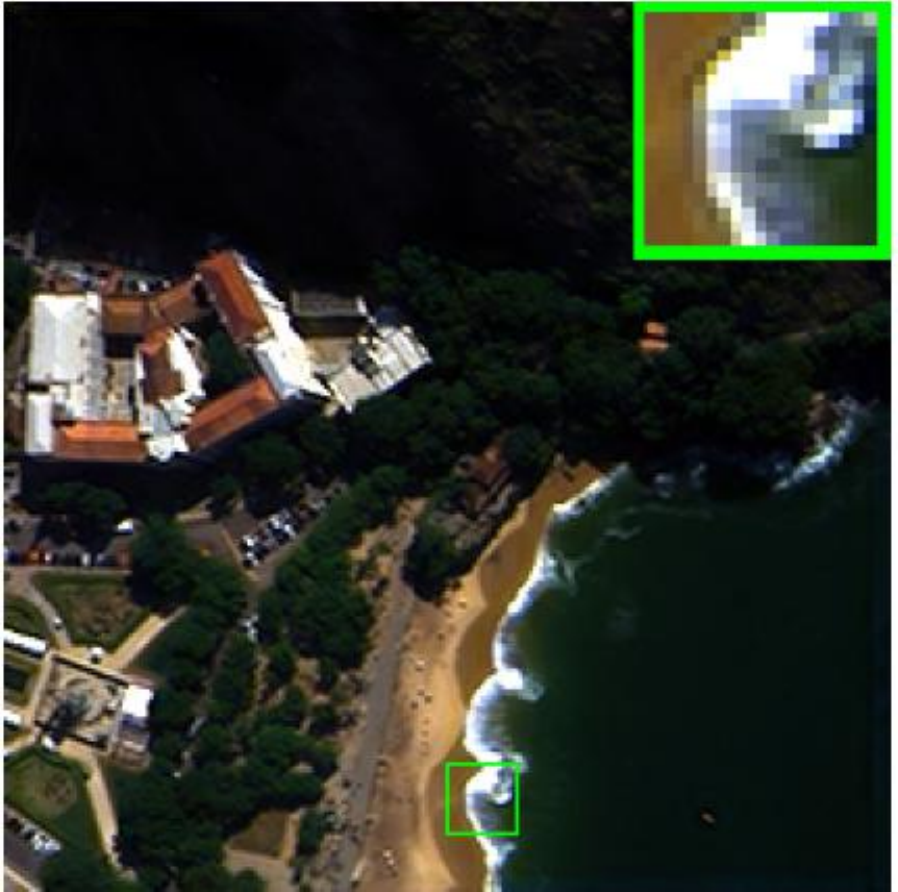}}
				{\includegraphics[width=1\linewidth]{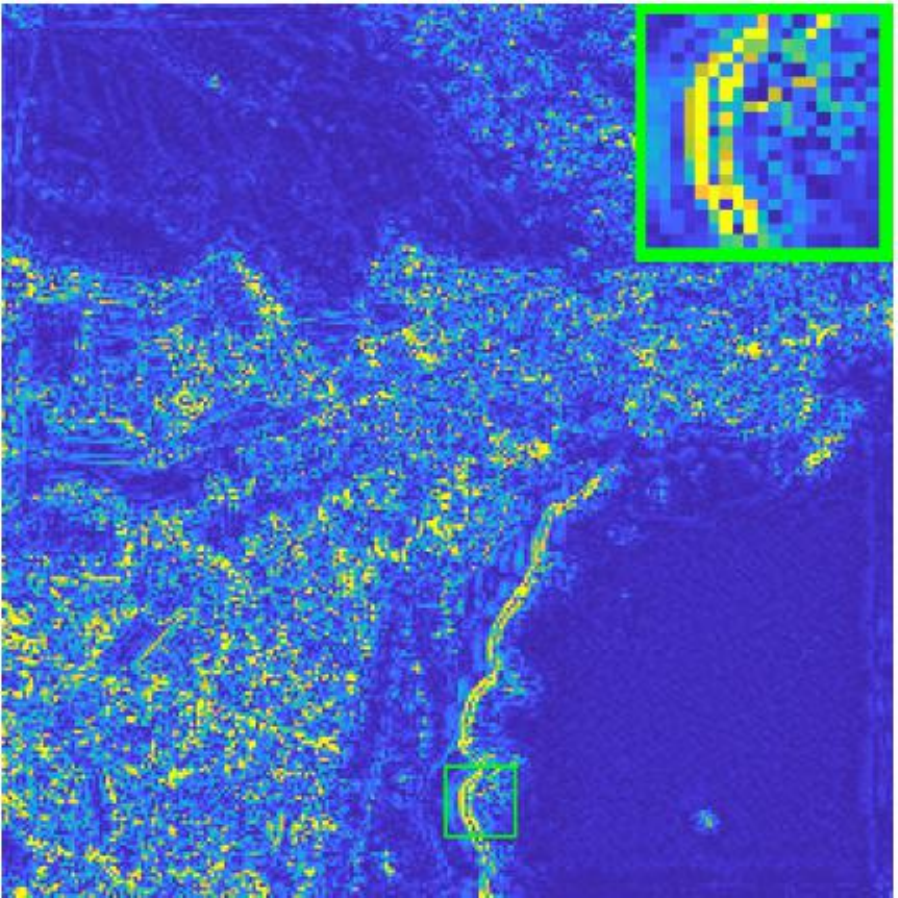}}
				\vspace{4pt}
				{MSDCNN}
				{\includegraphics[width=1\linewidth]{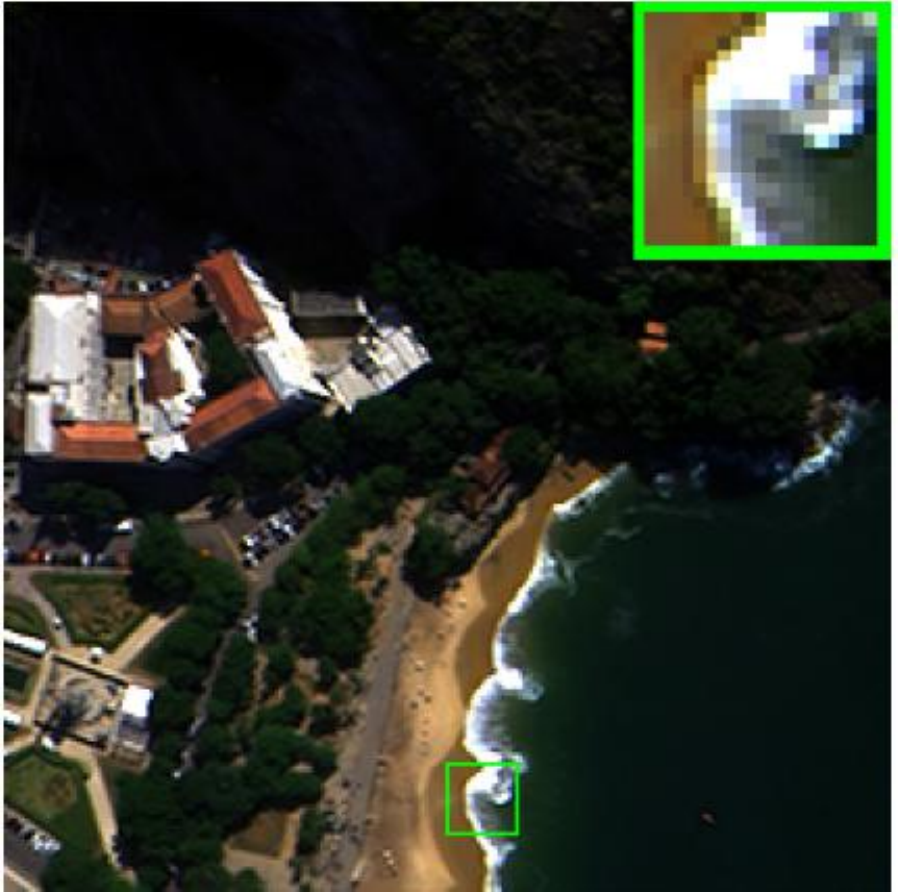}}
				{\includegraphics[width=1\linewidth]{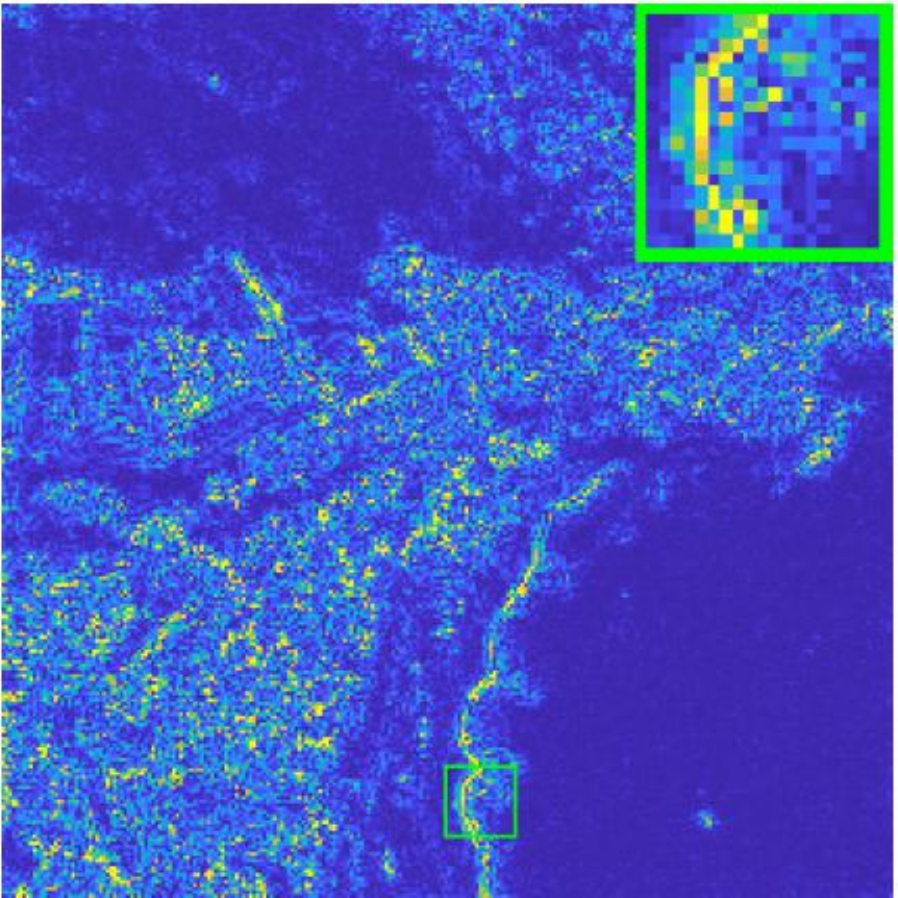}}
				\vspace{4pt}
				{CANNet}
				\centering
				
			\end{minipage}
			\begin{minipage}[t]{0.12\linewidth}
				{\includegraphics[width=1\linewidth]{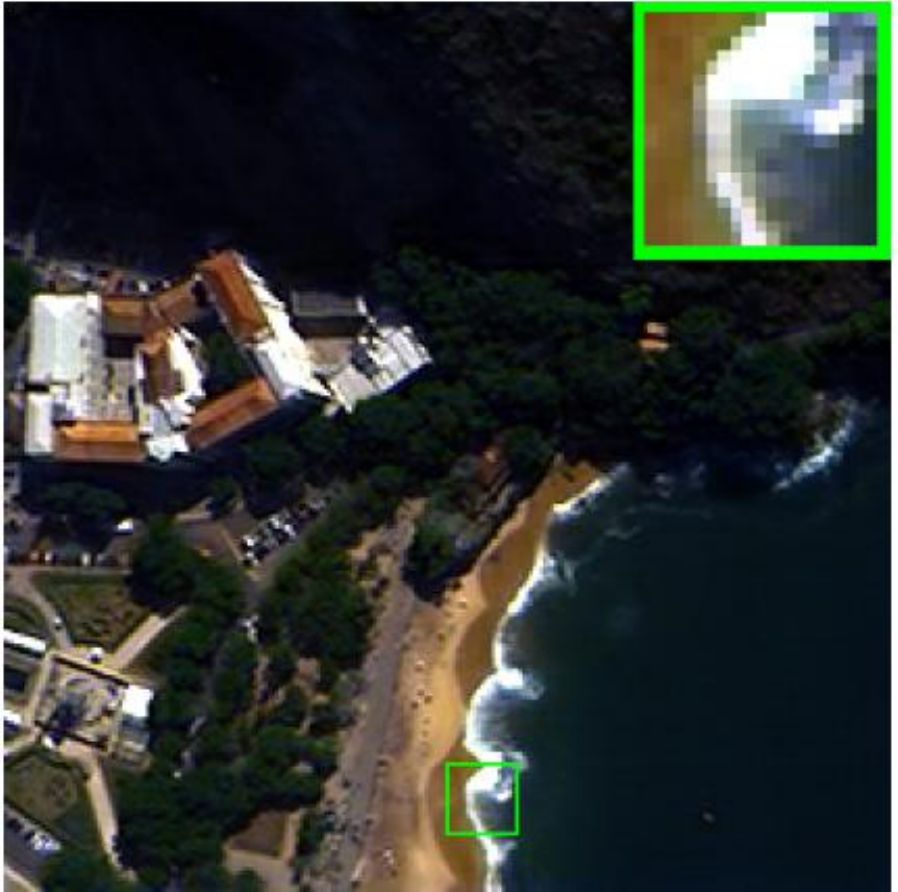}}
				{\includegraphics[width=1\linewidth]{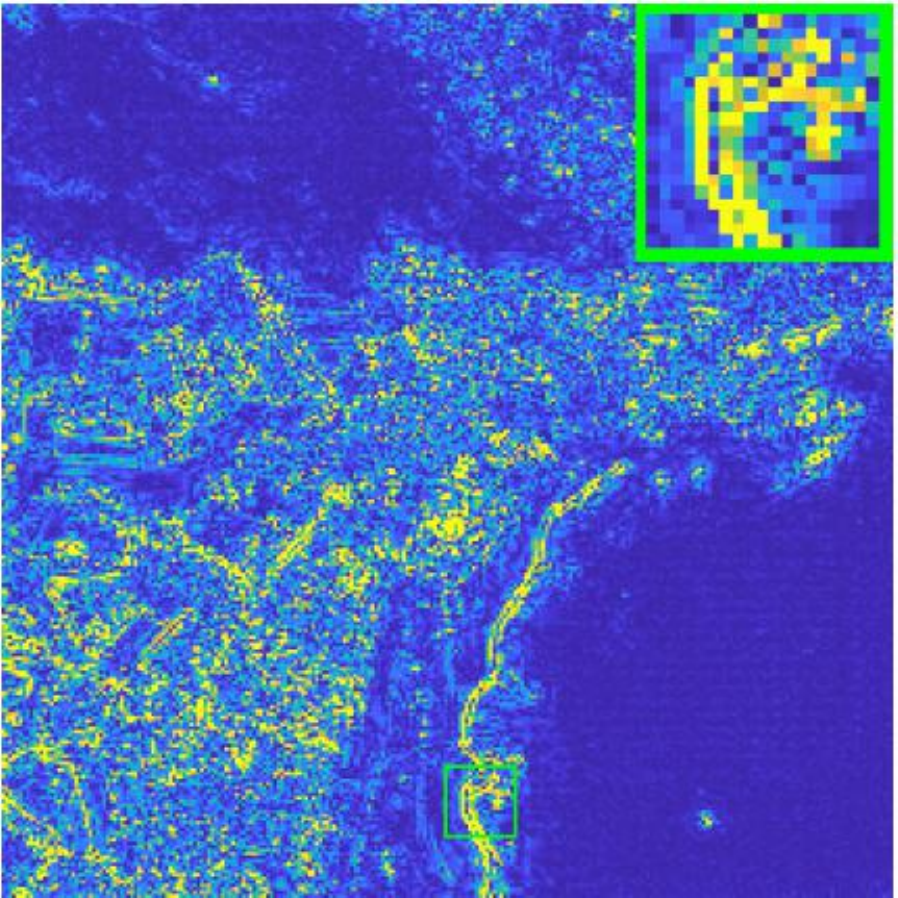}}
				\vspace{4pt}
				{BDPN}
				{\includegraphics[width=1\linewidth]{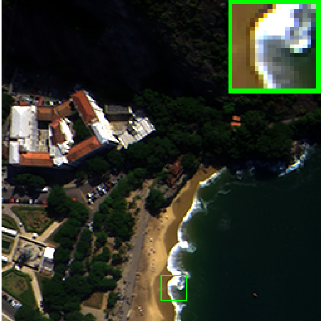}}
				{\includegraphics[width=1\linewidth]{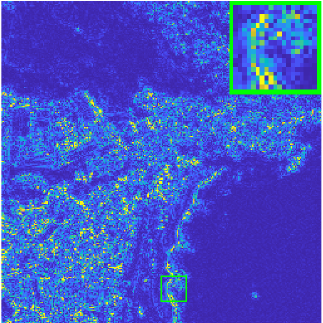}}
				\vspace{4pt}
				{ADWM}
				\centering
				
			\end{minipage}
			\begin{minipage}[t]{0.12\linewidth}
				{\includegraphics[width=1\linewidth]{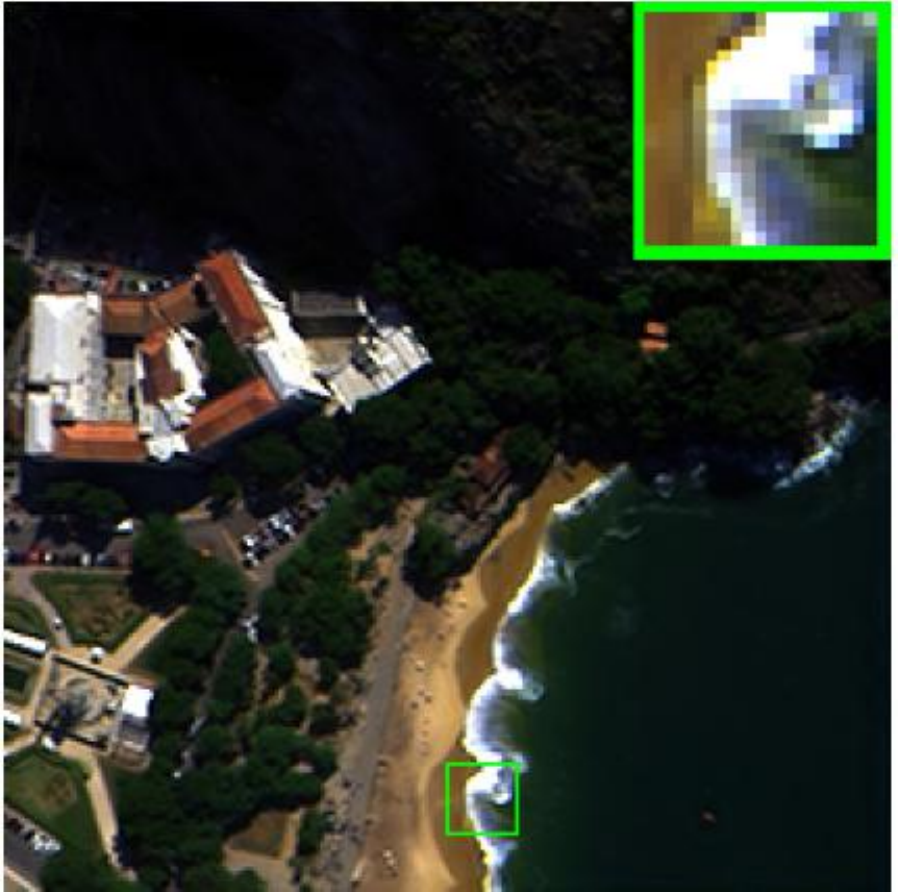}}
				{\includegraphics[width=1\linewidth]{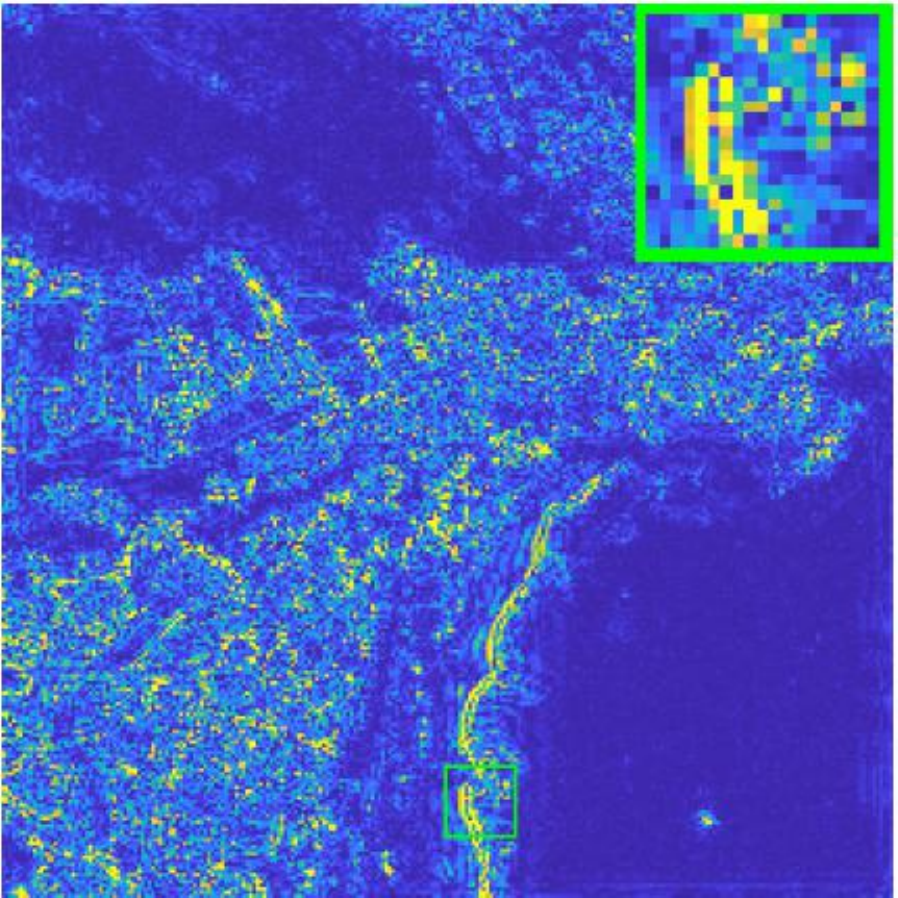}}
				\vspace{4pt}
				{FusionNet}
				{\includegraphics[width=1\linewidth]{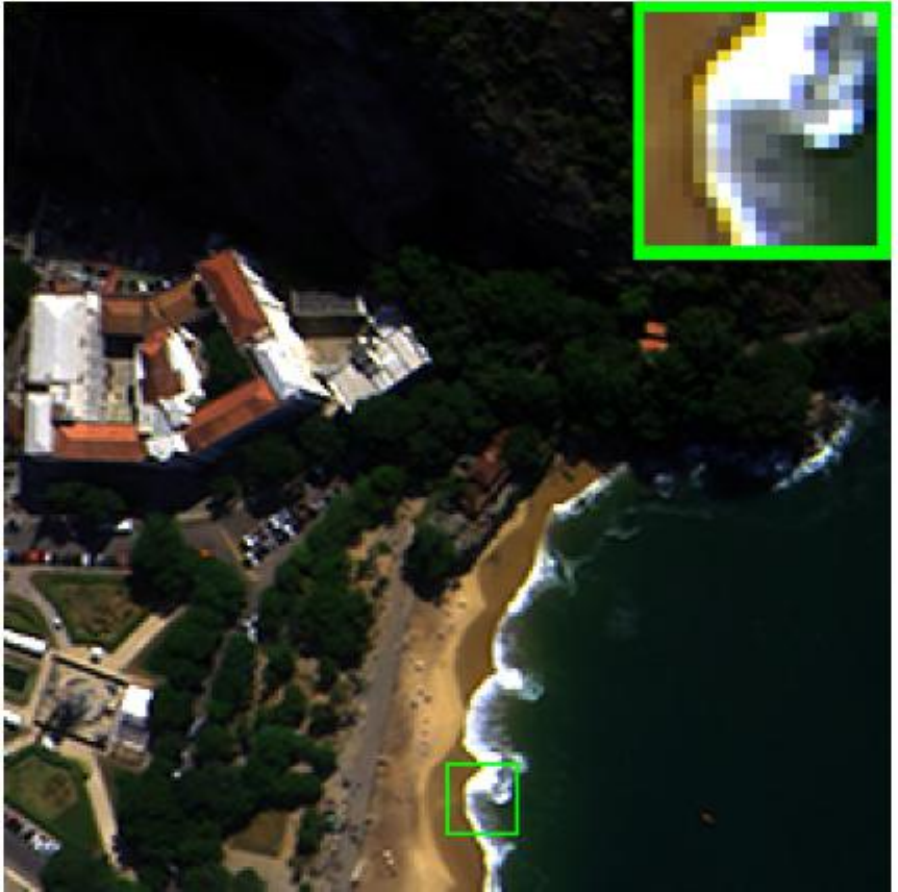}}
				{\includegraphics[width=1\linewidth]{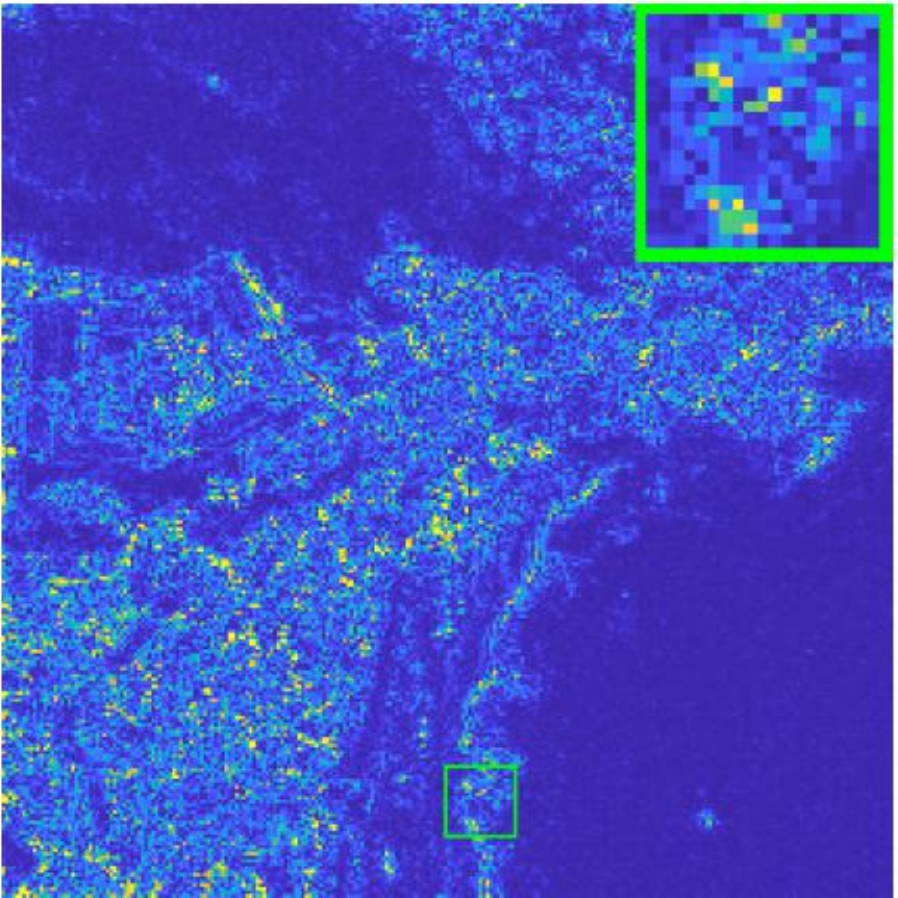}}
				\vspace{4pt}
				{Ada3D}
				\centering
				
			\end{minipage}
			\begin{minipage}[t]{0.12\linewidth}
				{\includegraphics[width=1\linewidth]{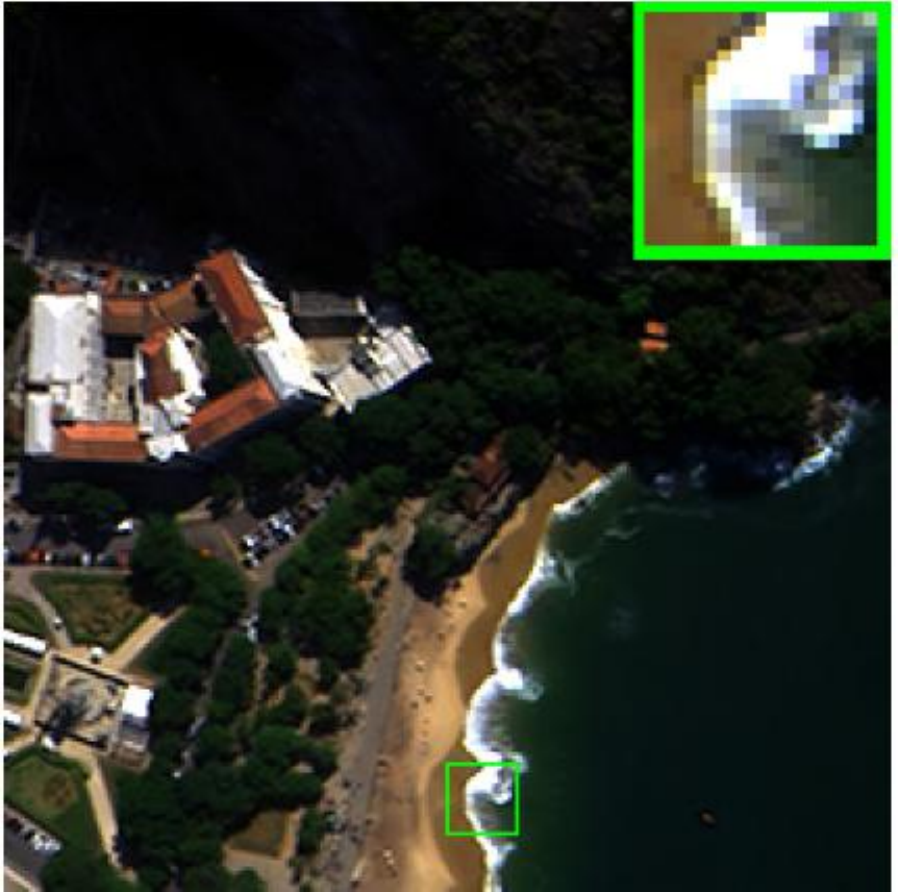}}
				{\includegraphics[width=1\linewidth]{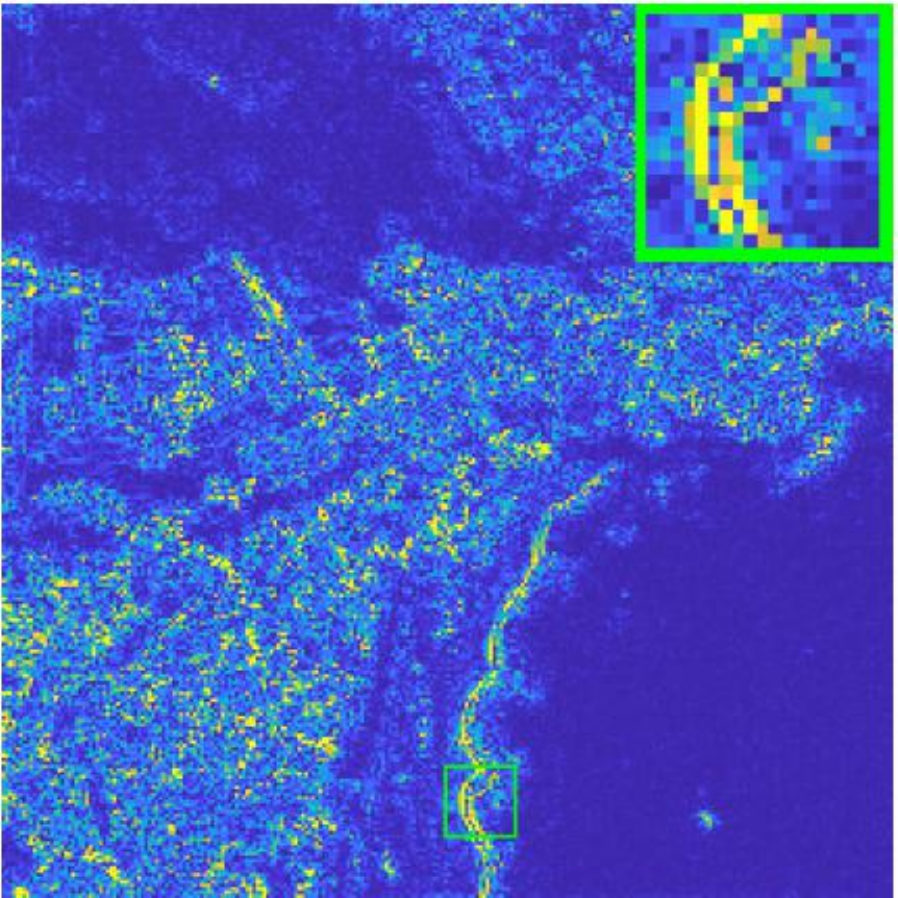}}
				\vspace{4pt}
				{MUCNN}
				{\includegraphics[width=1\linewidth]{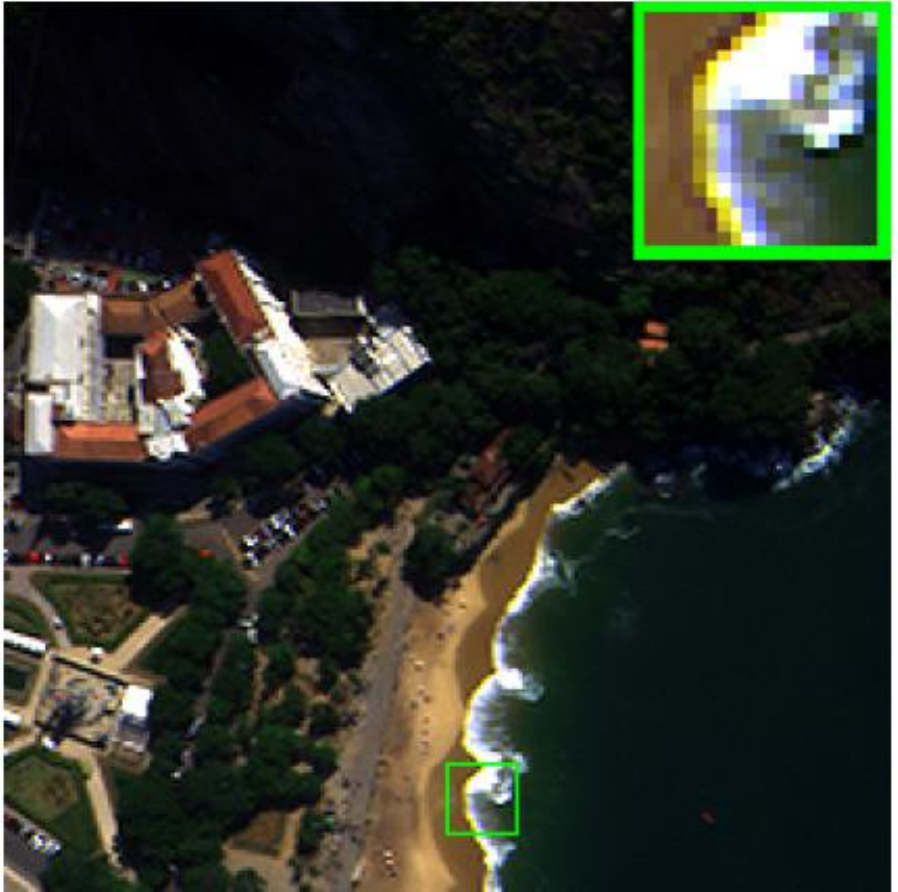}}
				{\includegraphics[width=1\linewidth]{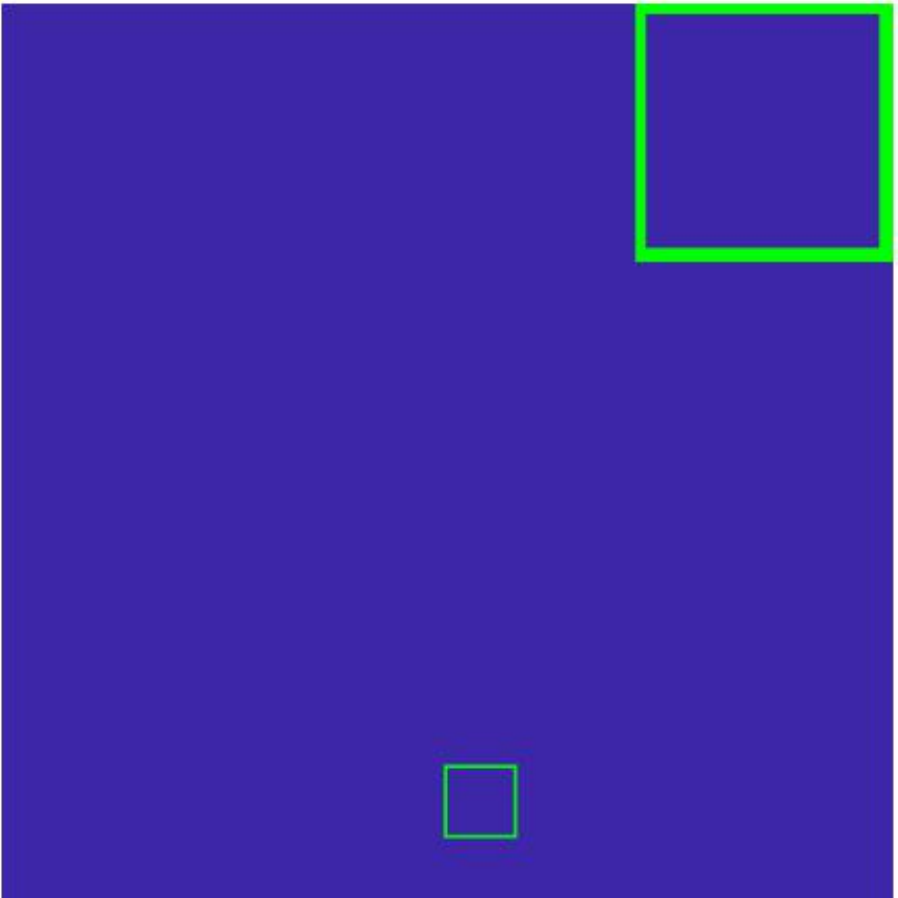}}
				\vspace{4pt}
				{GT}
				\centering
				
			\end{minipage}
		\end{minipage}
		\begin{minipage}[t]{0.98\linewidth}
			{\includegraphics[width=1\linewidth]{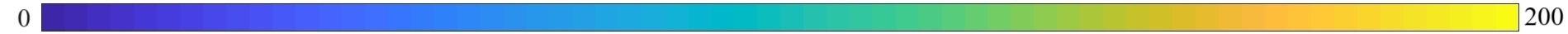}}
			\centering
		\end{minipage}
	\end{center}
	\vspace{-7pt}
	\caption{Qualitative evaluation results on a reduced-resolution example from the WV3 dataset, which belongs to the pansharpening task. Rows 1 and 3: Pseudo-color images representing spectral bands 1, 3, and 5. Rows 2 and 4: The corresponding AEMs for spectral band 7.\label{rr_v}}
	\vspace{-11pt}
\end{figure*}

\subsubsection{Benchmarks}
We compare Ada3D with several representative pansharpening techniques, including three traditional approaches: TV \cite{palsson2013new}, GLP-HPM \cite{6616569}, and BDSD-PC \cite{2019Robust}; and eleven DL-based methods: PNN \cite{2016Pansharpening}, MSDCNN \cite{8127731}, BDPN \cite{zhang2019pan}, FusionNet \cite{2020Detail}, MUCNN \cite{10.1145/3474085.3475600}, LAGNet \cite{jin2022aaai}, PMACNet \cite{9764690}, U2Net \cite{10.1145/3581783.3612084}, Pan-Mamba \cite{he2024pan}, CANNet \cite{Duan_2024_CVPR}, and ADWM \cite{Huang_2025_CVPR}. For fairness, all DL-based methods are trained in the same Nvidia 3090 GPU and PyTorch environment.

\subsubsection{Quality Indices}
In line with the research standards for pansharpening, we utilize four quality indices to assess the results on reduced-resolution samples: PSNR, Q2n \cite{2009Hypercomplex}, SAM, and ERGAS. The ideal values for these indices are +$\infty$, 1, 0, and 0, respectively. For full-resolution samples, we employ ${\rm{D}}_{\rm{\lambda}}$, ${\rm{D}}_{\rm{s}}$, and QNR \cite{6998089} as evaluation metrics, with ideal values of 0, 0, and 1. Notably, QNR, which integrates ${\rm{D}}_{\rm{\lambda}}$ and ${\rm{D}}_{\rm{s}}$, offers a comprehensive measure of overall fusion quality.

\subsubsection{Settings} 
For the pansharpening task, we configure both the spatial and spectral feature maps with 48 channels. The values of $k$, $\alpha$, and $\beta$ are set to 3, 0.25, and 0.25, respectively. Additionally, we select PixelShuffle for up-sampling. Furthermore, we utilize four ResBlocks to extract spatial features and stack eight Ada3D blocks for information fusion. All learnable weights are initialized using the Kaiming uniform distribution. During training, we configure the number of epochs, batch size, initial learning rate, and ${\lambda}_{ergas}$ as 500, 8, $1\times 10^{-4}$, and $1\times 10^{-4}$. In addition, we employ the Adam optimizer and reduce the learning rate by half at the $300$-th epoch.

\subsubsection{Results}
The quantitative evaluation results for the WV3 dataset are summarized in Table \ref{rr}. It is evident that the proposed Ada3D outperforms all other techniques on both reduced-resolution and full-resolution testing samples. With indicator values nearing their limits, our method demonstrates substantial improvements over other approaches. Additionally, the qualitative evaluation outcomes, as shown in Fig. \ref{rr_v}, illustrate that Ada3D produces the darkest AEMs. These findings demonstrate that Ada3D excels in the pansharpening task.

\begin{table}[t]
	\centering\renewcommand\arraystretch{1.}\setlength{\tabcolsep}{3.2pt}
	\belowrulesep=0pt\aboverulesep=0pt
	\caption{Ablation study of our network design using testings samples from the WDC dataset for hyper-spectral pansharpening.}\label{abl1}
	\begin{tabular}{l|c|c|ccccc}
		\toprule
		\textbf{Methods} & Params & FLOPs & PSNR & CC & SSIM & SAM & ERGAS \\  
		\midrule  
		\textbf{w/o ResBlocks} & 0.52M & 2.41G & \underline{31.842} & \underline{0.970} & \underline{0.892} & 3.543 & 3.466 \\
		\textbf{Ada3D-tiny} & 0.18M & 0.82G & 31.382 & \textbf{0.971} & 0.891 & 3.529 & 3.434 \\
		\textbf{Ada3D-small} & 0.37M & 1.66G & 31.641 & \underline{0.970} & 0.891 & \underline{3.465} & \underline{3.427} \\
		\textbf{Ada3D-large} & 1.26M & 5.44G & 31.823 & 0.969 & 0.889 & 3.592 & 3.562 \\
		\textbf{The Proposed} & 0.58M & 2.64G & \textbf{31.998} & \textbf{0.971} & \textbf{0.893} & \textbf{3.461} & \textbf{3.414} \\
		\bottomrule
	\end{tabular}
\end{table}

\subsection{Ablation Studies}
\subsubsection{Network Design}
We evaluate the effectiveness of our network design using testing samples from the WDC dataset. To assess the impact of ResBlocks on spatial feature extraction, we conduct experiments without them (w/o ResBlocks). Additionally, we develop three variants of Ada3D that share the same architectural design but vary in model size (Ada3D-tiny, Ada3D-small, and Ada3D-large). This helps verify the correctness of our network configuration. As shown in Table~\ref{abl1}, the proposed method achieves the best results across all quality indices, demonstrating the validity of our design.

\begin{table}[t]
	\centering\renewcommand\arraystretch{1.}\setlength{\tabcolsep}{2.8pt}
	\belowrulesep=0pt\aboverulesep=0pt
	\caption{Ablation study of the Ada3D block on the WDC dataset. All compared methods have an identical number of parameters.}\label{abl2}
	\begin{tabular}{l|c|ccccc}
		\toprule
		\textbf{Methods} & FLOPs & PSNR & CC & SSIM & SAM & ERGAS \\  
		\midrule  
		\textbf{Ada3D-rev} & 0.69G & 30.069 & 0.961 & 0.874 & 4.128 & 3.971 \\
		\textbf{w/o Spatial Kernels} & 2.58G & 30.852 & 0.963 & 0.876 & 3.912 & 3.878 \\
		\textbf{w/o Spectral Kernels} & 2.73G & 30.691 & 0.961 & 0.870 & 4.135 & 3.976 \\
		\textbf{w/o Spatial Biases} & 2.63G & \underline{31.907} & \textbf{0.971} & \underline{0.891} & \underline{3.518} & \underline{3.448} \\
		\textbf{w/o Spectral Biases} & 2.60G & 31.829 & \underline{0.970} & 0.889 & 3.682 & 3.555 \\
	\textbf{w/o Channel Biases} & 2.63G & 31.846 & 0.969 & \underline{0.891} & 3.611 & 3.539 \\
		\textbf{w/o Adaptive Biases} & 2.59G & {31.742} & {0.969} & {0.890} & {3.617} & {3.585} \\
		\textbf{w/o Group Convolution} & 5.40G & 31.623 & \underline{0.970} & 0.889 & 3.556 & 3.527 \\
		\textbf{The Proposed} & 2.64G & \textbf{31.998} & \textbf{0.971} & \textbf{0.893} & \textbf{3.461} & \textbf{3.414} \\
		\bottomrule
	\end{tabular}
\end{table}

\subsubsection{Ada3D Block}
To validate the effectiveness of the Ada3D block, we develop eight variants and evaluate them using testing samples from the WDC dataset. For fairness, all compared methods are designed with the same number of network parameters. First, we create spatial kernels from the spectral feature map and spectral kernels from the spatial feature map (Ada3D-rev). This variant helps verify the correctness of our kernel generation strategy. Next, we examine the efficacy of the spatial and spectral kernels. Since directly removing spatial kernels would eliminate spatial information, we develop a new pipeline that integrates spatial and spectral characteristics through a dot-product operation. Then, the spectral kernels are duplicated to match the size of the adaptive 3D kernels (w/o Spatial Kernels). Additionally, we directly remove the spectral kernels to evaluate their effectiveness (w/o Spectral Kernels). Furthermore, we investigate the effect of excluding various biases, including spatial biases (w/o Spatial Biases), spectral biases (w/o Spectral Biases), channel biases (w/o Channel Biases), and all adaptive biases (w/o Adaptive Biases). Lastly, we examine the role of group convolution by removing it from the design (w/o Group Convolution). The quantitative results, as presented in Table~\ref{abl2}, strongly affirm the validity of our structural design for the Ada3D block. 

\begin{table}[t]
	\centering\renewcommand\arraystretch{1.}\setlength{\tabcolsep}{5pt}
	\belowrulesep=0pt\aboverulesep=0pt
	\caption{Comparison with other convolutional paradigms using testing samples from the WDC dataset. All compared methods are configured with the same number of network parameters. The techniques above the dividing line represent 2D methods, while those below correspond to 3D approaches.}\label{abl5}
	\begin{tabular}{l|c|ccccc}
		\toprule
		\textbf{Methods} & FLOPs & PSNR & CC & SSIM & SAM & ERGAS \\  
		\midrule  
		\textbf{Conv2D} & 2.42G & 30.918 & 0.964 & 0.875 & 3.842 & 3.874 \\
		\textbf{Ada2D} & 1.89G & 30.890 &  0.966 & 0.886 & 3.630 & 3.786 \\
		\midrule
		\textbf{Conv3D} & 327G & 31.696 & 0.963 & 0.876 & \textbf{3.425} & 4.183 \\
		\textbf{DW3D} & 81.9G & 31.261 & 0.964 & 0.877 & 3.475 &  4.110 \\
		\textbf{GPAC3D} & 342G & \underline{31.874} & 0.969 & \underline{0.891} & 3.576 & \underline{3.524} \\
		\textbf{D2Conv3D} \cite{Schmidt_2022_WACV} & 658G & 31.302 & 0.969 & 0.888 &  3.530 & 3.536 \\
		\textbf{ST-A3DNet} \cite{10.1145/3510829} & 14.3G & 31.000 &  \underline{0.970} & 0.886 & 3.465 & 5.817 \\
		\textbf{The Proposed} & 2.64G & \textbf{31.998} & \textbf{0.971} & \textbf{0.893} & \underline{3.461} & \textbf{3.414} \\
		\bottomrule
	\end{tabular}
\end{table}

\begin{figure}[t]
	\begin{center}
		\begin{minipage}{1\linewidth}
			{\includegraphics[width=1\linewidth]{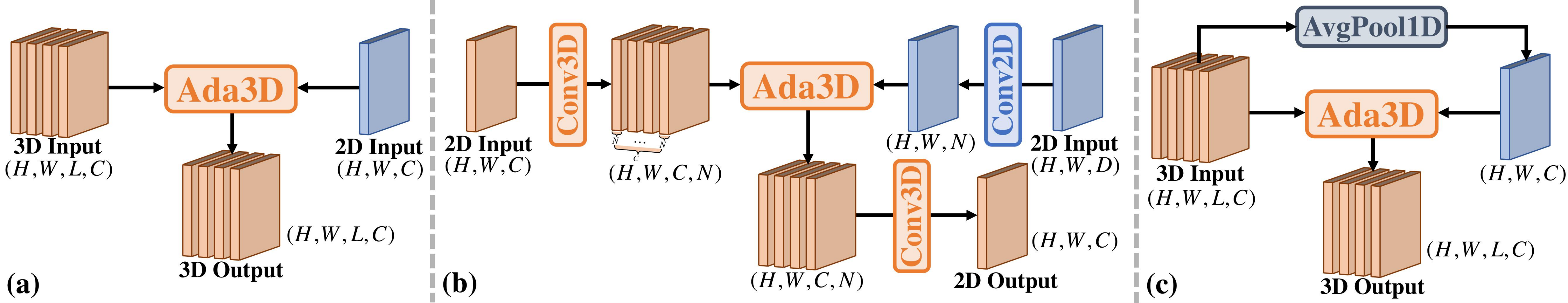}}
			\centering
		\end{minipage}
	\end{center}
	\vspace{-5pt}
	\caption{Graphical illustration of Ada3D's applications. \textbf{(a)} The original Ada3D block. \textbf{(b)} Ada3D used as an information integration module, merging two 2D inputs of sizes $H\times W\times C$ and $H\times W\times D$. Here, $D$ and $N$ represent the number of channels in the 2D and 3D feature maps, respectively. \textbf{(c)} Ada3D functioning as a single-input 3D feature extraction block. \label{ada3d_app}}
\end{figure}

\subsubsection{Comparison with Other Convolutional Paradigms}
We compare Ada3D with several convolutional paradigms, including standard 2D convolution (Conv2D), adaptive 2D convolution (Ada2D), standard 3D convolution (Conv3D), depth-wise 3D convolution (DW3D), GPAC3D, D2Conv3D \cite{Schmidt_2022_WACV}, and ST-A3DNet \cite{10.1145/3510829}. Ada2D, based on the design in \cite{ijcai2022p179}, serves as the 2D counterpart to Ada3D. Given the high computational cost associated with the PAC3D block, we adopt a more efficient group convolution version, which we call GPAC3D. 
In terms of network architecture, for 2D methods (Conv2D and Ada2D), we incorporate a spectral branch specifically designed for the 2D domain. For 3D approaches (all others), we replace the original Ada3D blocks in Fig.~\ref{pip} with the corresponding 3D convolutional paradigms. Except for Ada2D, we apply a concatenation operation before the convolution block to integrate spatial and spectral information. To ensure fairness, all methods are configured to have an equivalent number of parameters. 
The quantitative evaluation outcomes on testing samples of the WDC dataset, as presented in Table~\ref{abl5}, highlight two key findings: (\emph{i}) 3D approaches generally outperform 2D methods but require significantly more computational resources; (\emph{ii}) the proposed Ada3D achieves the best overall results, with FLOPs comparable to those of 2D methods. Consequently, our method combines the strengths of 2D and 3D convolutional paradigms.

\begin{table}[t]
	\centering\renewcommand\arraystretch{1.}\setlength{\tabcolsep}{5.8pt}
	\belowrulesep=0pt\aboverulesep=0pt
	\caption{Applications of Ada3D in various pansharpening frameworks, evaluated on reduced-resolution WV3 testing samples.}\label{abl3}
	\begin{tabular}{l|c|cccc}
		\toprule
		\textbf{Methods} & Params & PSNR & Q2n & SAM & ERGAS \\  
		\midrule  
		\textbf{PNN \cite{2016Pansharpening}} & 0.10M & 37.313 & 0.893 & 3.677 & 2.681 \\
		\textbf{PNN + Ada3D} & 0.12M & \textbf{37.914} & \textbf{0.902} & \textbf{3.331} & \textbf{2.479} \\
		\midrule
		\textbf{FusionNet \cite{2020Detail}} & 0.08M & 38.047 & 0.904 & 3.324 & 2.465 \\
		\textbf{FusionNet + Ada3D}  & 0.10M & \textbf{38.569} & \textbf{0.912} & \textbf{3.108} & \textbf{2.302} \\
		\midrule
		\textbf{U2Net \cite{10.1145/3581783.3612084}} & 0.66M & 39.117 & \textbf{0.920} & 2.888 & 2.149 \\
		\textbf{U2Net + Ada3D} & 0.67M & \textbf{39.183} & \textbf{0.920} & \textbf{2.885} & \textbf{2.129} \\
		\bottomrule
	\end{tabular}
\end{table}

\begin{figure}[t]
	\begin{center}
		\begin{minipage}{1\linewidth}
			{\includegraphics[width=1\linewidth]{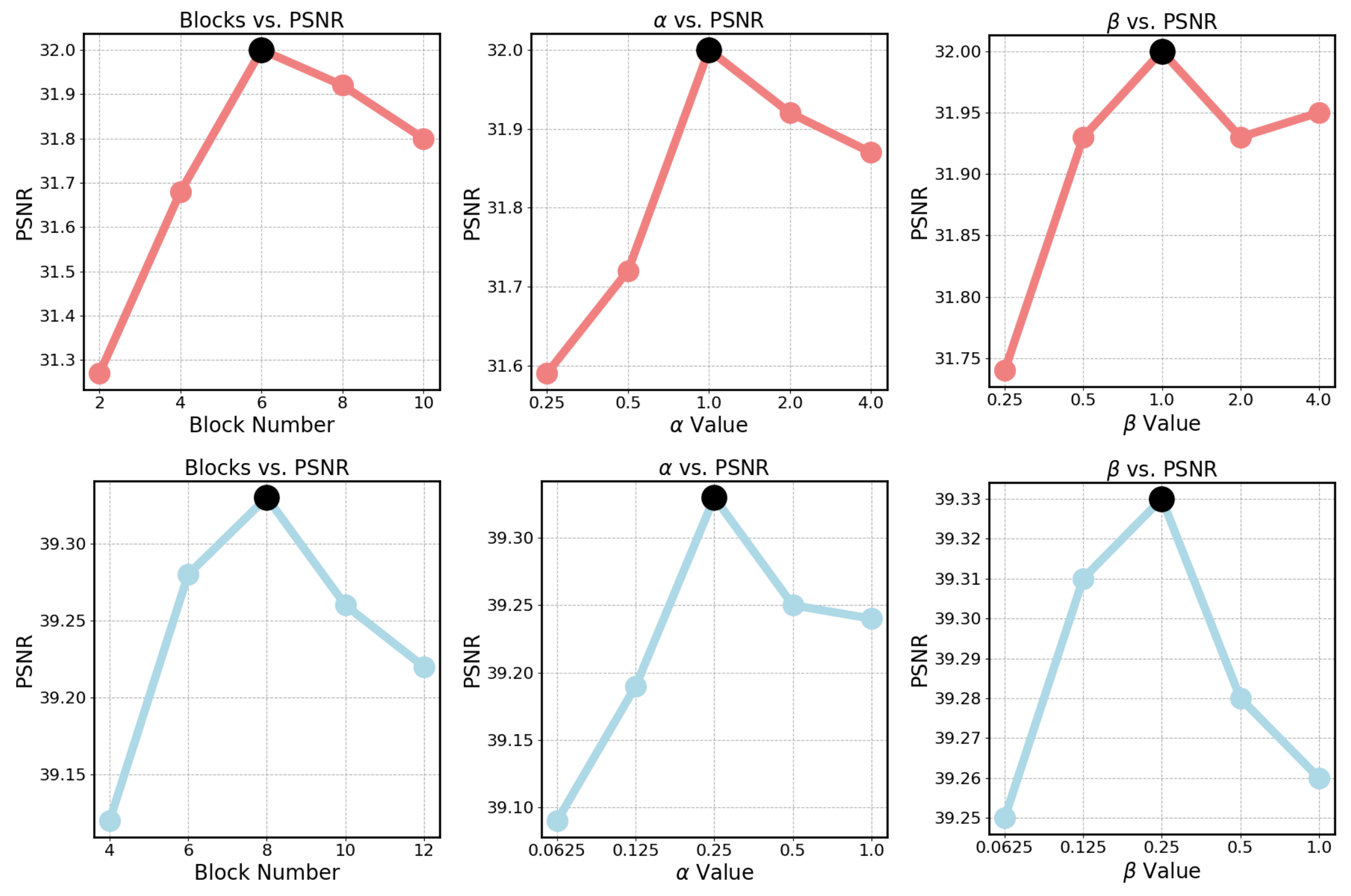}}
			\centering
		\end{minipage}
	\end{center}
	\vspace{-5pt}
	\caption{Ablation study on three hyper-parameters: block number, $\alpha$, and $\beta$. The first row corresponds to the hyper-spectral pansharpening task, evaluated on the testing samples of the WDC dataset, while the second row pertains to the pansharpening task, evaluated on the reduced-resolution testing samples of the WV3 dataset. The black dots represent our chosen settings. \label{hyper_param}}
\end{figure}

\subsubsection{The Applications of Ada3D}
The original Ada3D block is designed to fuse 2D and 3D inputs, which is relatively uncommon in practice. To expand its practical utility, we explore two additional applications of Ada3D, as illustrated in Fig.~\ref{ada3d_app}. First, the Ada3D block can serve as a plug-and-play module for integrating two 2D inputs, even with differing dimensions, by incorporating extra convolutional layers. Second, it can be repurposed for 3D feature extraction from a single input by applying average pooling across the third dimension of the 3D input. 
We examine the first additional application of Ada3D by embedding it into several representative pansharpening frameworks: PNN \cite{2016Pansharpening}, FusionNet \cite{2020Detail}, and U2Net \cite{10.1145/3581783.3612084}. Specifically, we replace the concatenation operation in PNN and FusionNet, as well as the S2Block in U2Net, with the enhanced Ada3D block. Table~\ref{abl3} presents the quantitative evaluation results using reduced-resolution samples from the WV3 dataset. Our method significantly improves performance, especially when baseline metrics are low. Even when the baseline performance is high, our approach exceeds the existing benchmarks. These findings indicate that the Ada3D block is an effective plug-and-play module for various architectures.

\subsubsection{Hyper-parameters}
We perform a comprehensive set of ablation studies to validate our hyper-parameter choices, focusing on three key parameters: the number of Ada3D blocks, the value of $\alpha$, and the value of $\beta$. As depicted in Fig.~\ref{hyper_param}, our selected settings consistently produce the best performance, underscoring the soundness of our design decisions.

\begin{table}[t]	
	\centering\renewcommand\arraystretch{1.}\setlength{\tabcolsep}{6.6pt}
	\belowrulesep=0pt\aboverulesep=0pt
	\caption{Quantitative evaluation results on testing samples from the CAVE dataset, which belongs to the HISR task. The symbol $\divideontimes$ represents methods utilizing 3D modeling. \label{cave}}	
	\begin{tabular}{l|c|cccc}
		\toprule
		\textbf{Methods} & {Params} & PSNR & SSIM & SAM & ERGAS \\ 
		\midrule
		\textbf{ResTFNet} \cite{2018Remote} & 2.39M & 45.584 & 0.994 & 2.764 & 2.313 \\ 
		\textbf{SSRNet} \cite{9186332} & 0.03M & 48.620 & 0.995& 2.542 & 1.636 \\ 
		\textbf{MoG-DCN}\textsuperscript{$\divideontimes$} \cite{9429905} & 6.84M & \underline{51.688} & \underline{0.997} & \underline{1.971} & \underline{1.104} \\
		\textbf{Fusformer} \cite{9841513} & 0.50M & 49.983 & 0.994 & 2.203 & 2.534 \\ 
		\textbf{3DT-Net} \cite{ma2023learning} & 3.16M & {51.471}& \underline{0.997} & {2.117} & {1.119} \\ 
		\textbf{U2Net} \cite{10.1145/3581783.3612084} & 2.65M & 50.433& \underline{0.997} & 2.187 & 1.277 \\
		\textbf{Ada3D (Ours)} &  2.30M & \textbf{52.018}& \textbf{0.998}& \textbf{1.875} & \textbf{1.035} \\ 
		\midrule
		\textbf{Ideal Values} &  $-$ & \textbf{+$\infty$} & \textbf{1} & \textbf{0} & \textbf{0} \\ 
		\bottomrule
	\end{tabular}
\end{table}

\begin{figure}[t]
	\begin{center}
			\begin{minipage}[t]{0.97\linewidth}
				    \begin{minipage}[t]{0.155\linewidth}
				    	{\includegraphics[width=1\linewidth]{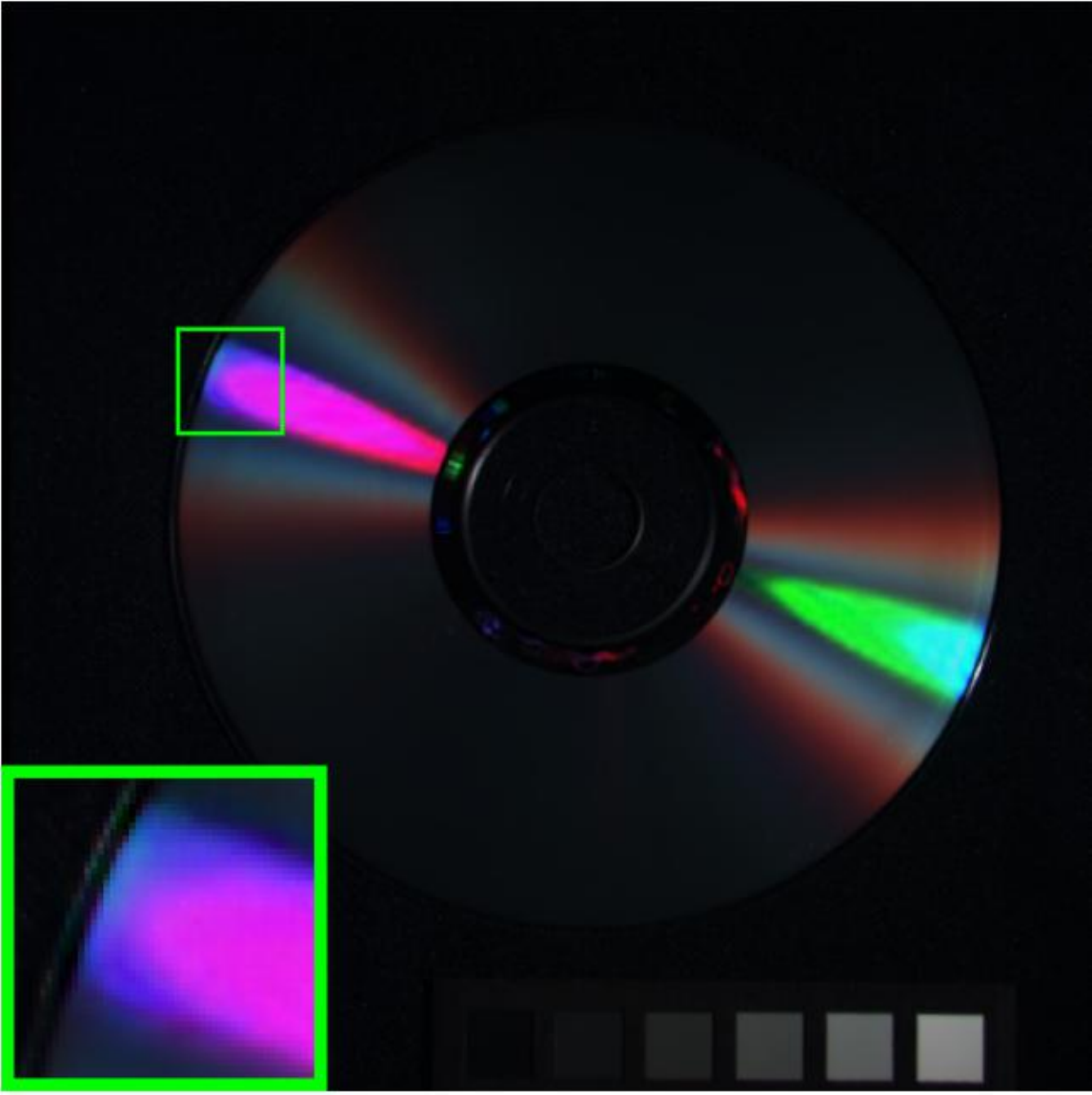}}
				    	\vspace{2pt}
				    	{\includegraphics[width=1\linewidth]{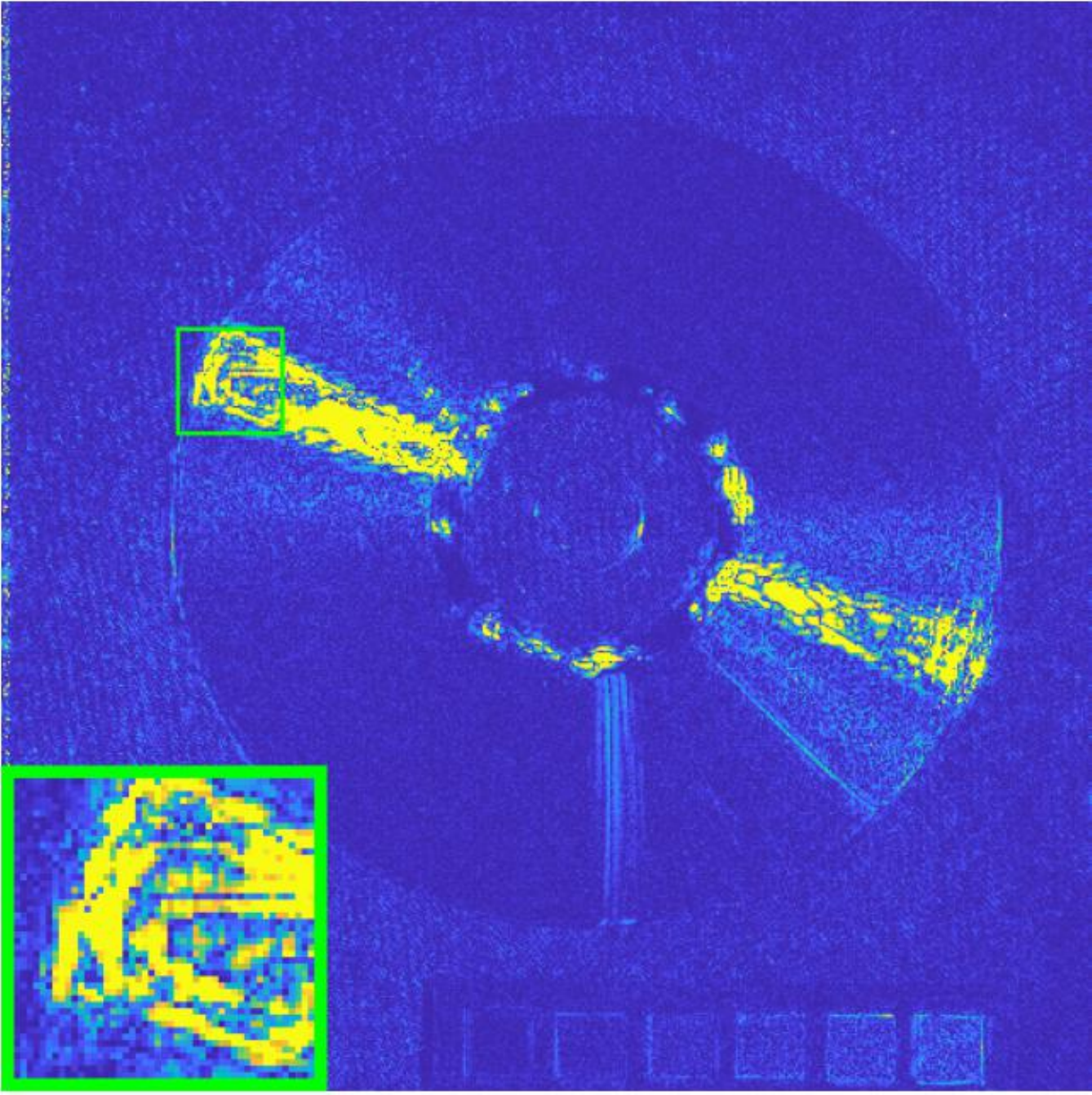}}
				    	{\includegraphics[width=1\linewidth]{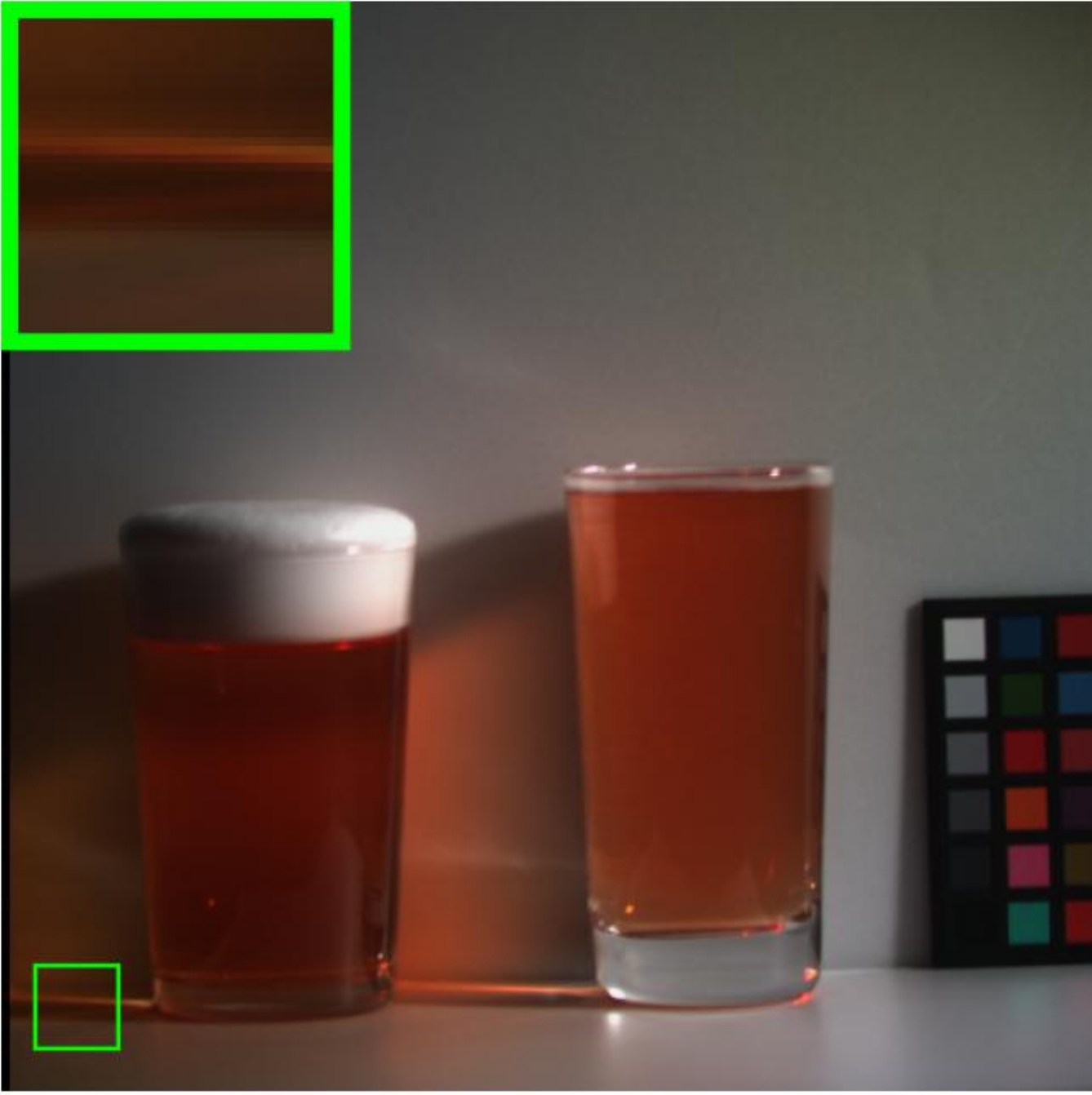}}
				    	{\includegraphics[width=1\linewidth]{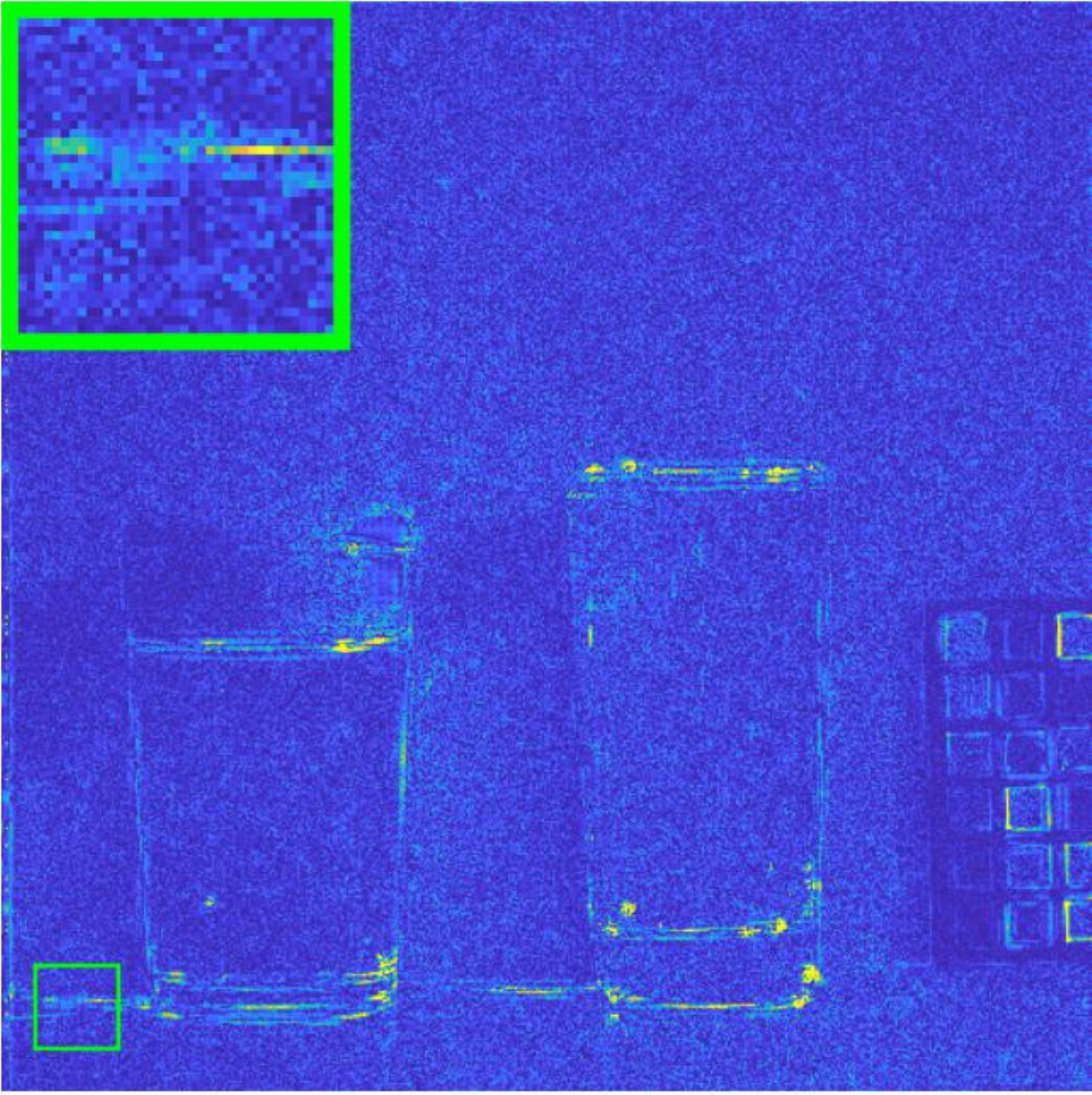}}
				    	\vspace{2pt}
				    	\scriptsize{MoG-DCN}
				    	\centering
				    	
				    \end{minipage}
					\begin{minipage}[t]{0.155\linewidth}
							{\includegraphics[width=1\linewidth]{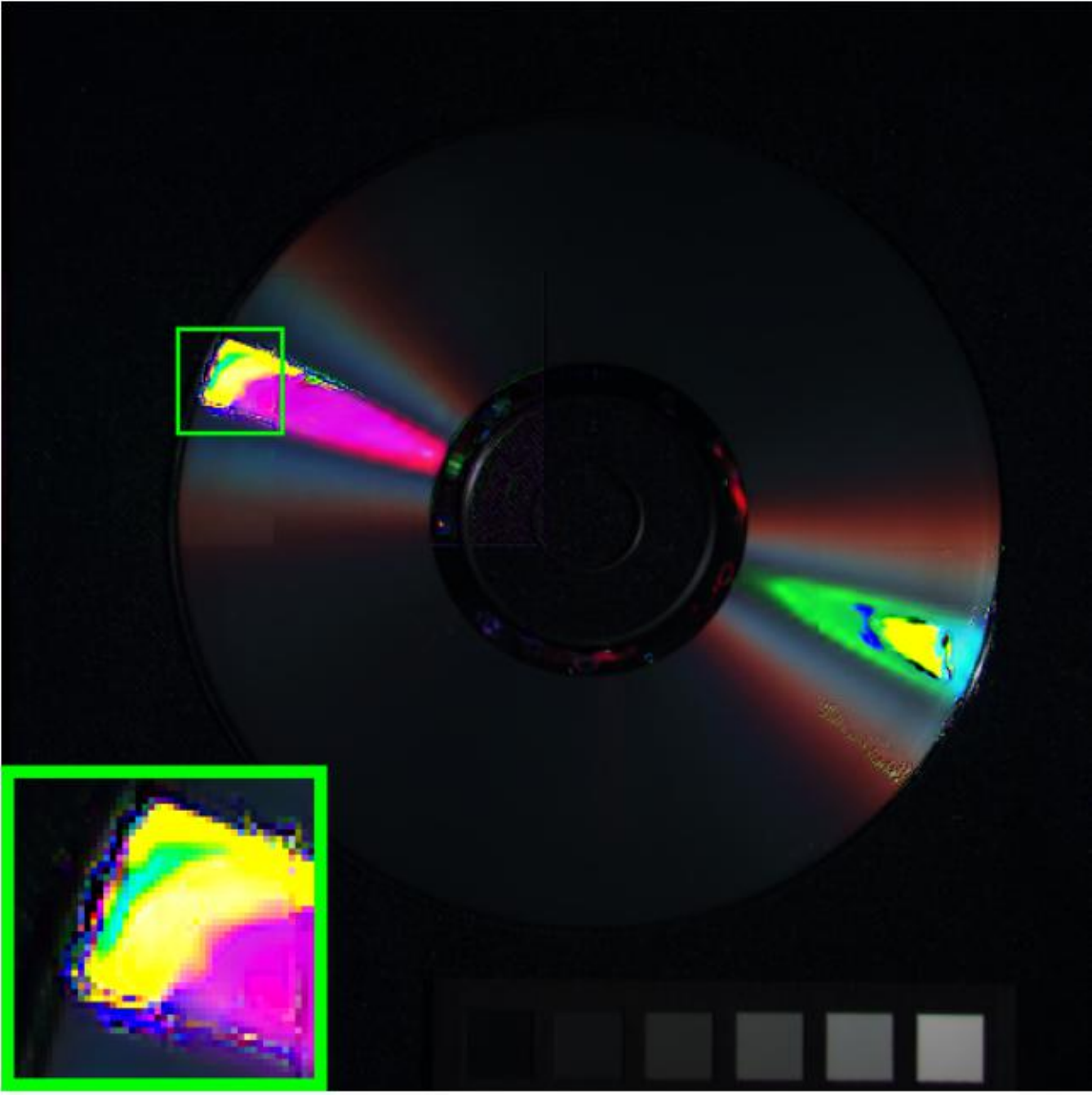}}
							\vspace{2pt}
							{\includegraphics[width=1\linewidth]{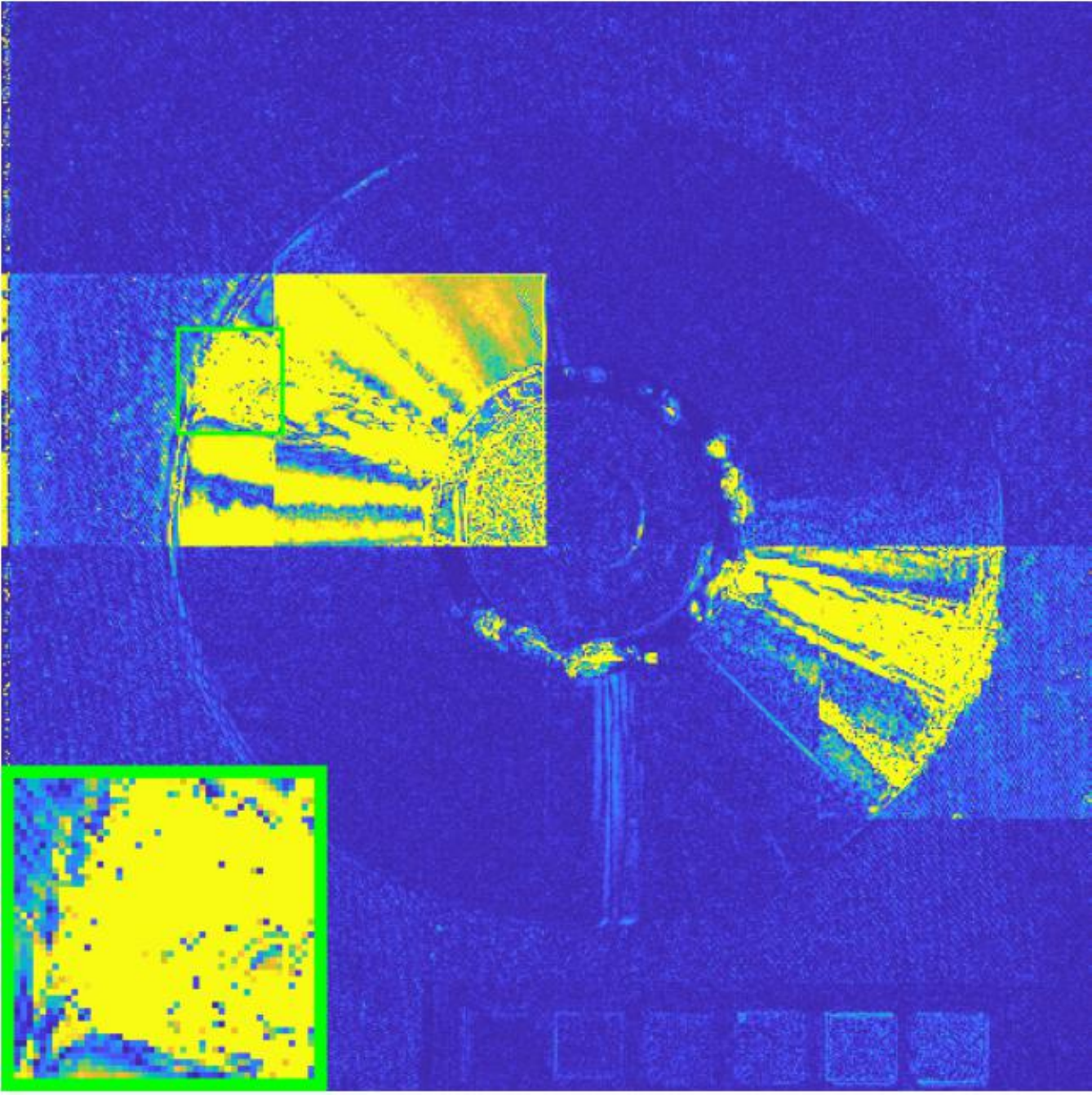}}
							{\includegraphics[width=1\linewidth]{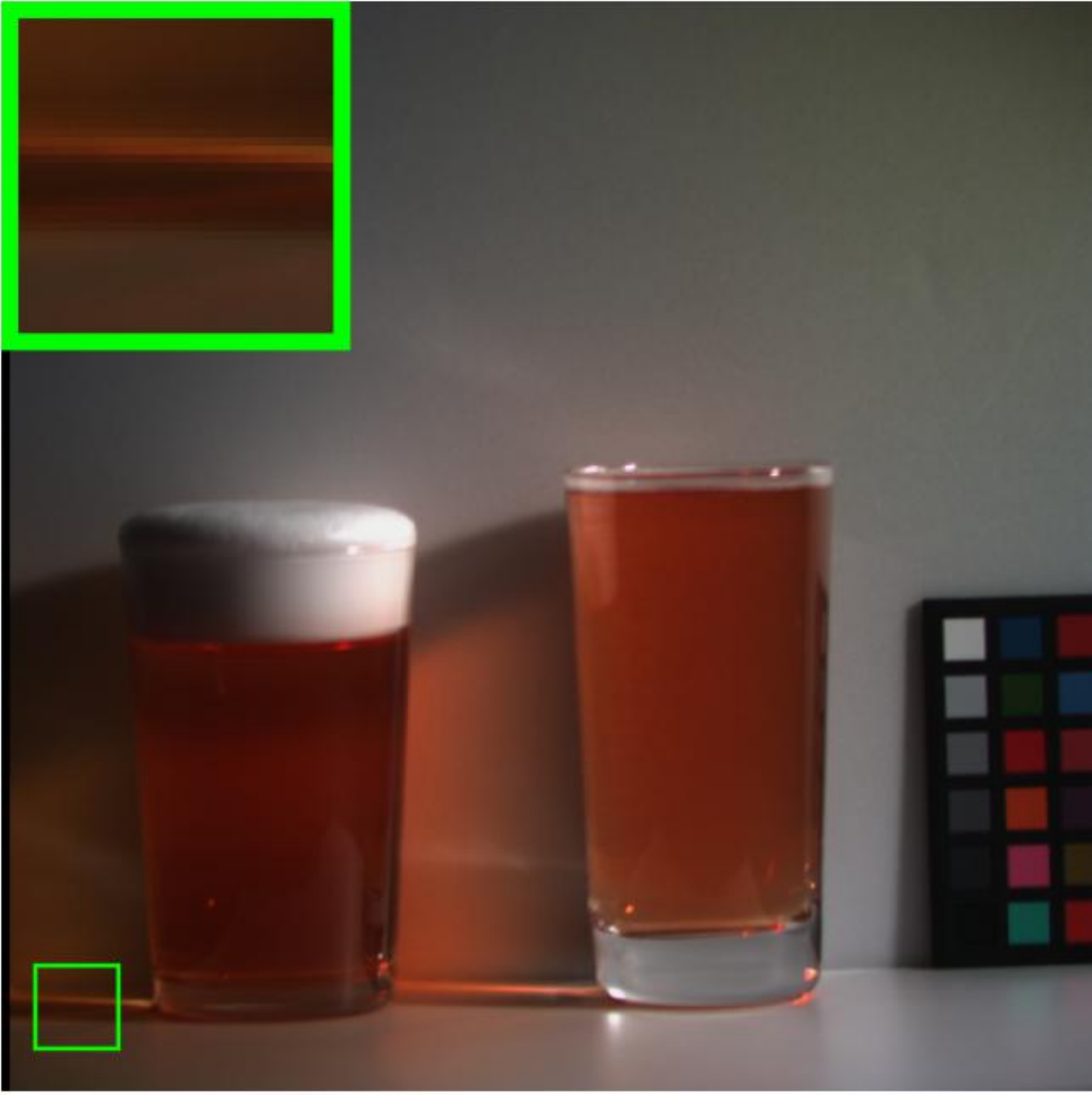}}
							{\includegraphics[width=1\linewidth]{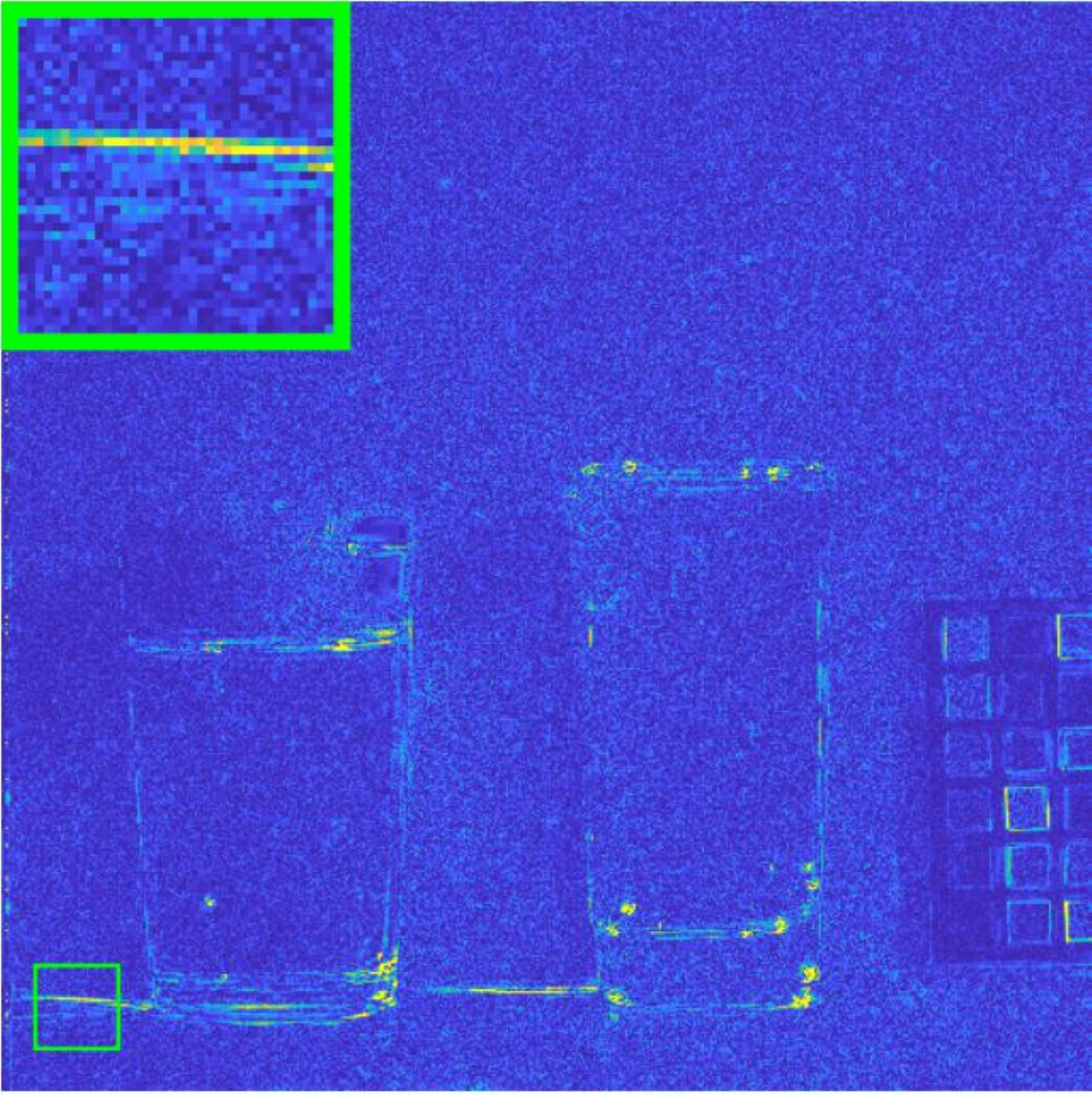}}
							\vspace{2pt}
							\scriptsize{Fusformer}
							\centering
							
						\end{minipage}
					\begin{minipage}[t]{0.155\linewidth}
							{\includegraphics[width=1\linewidth]{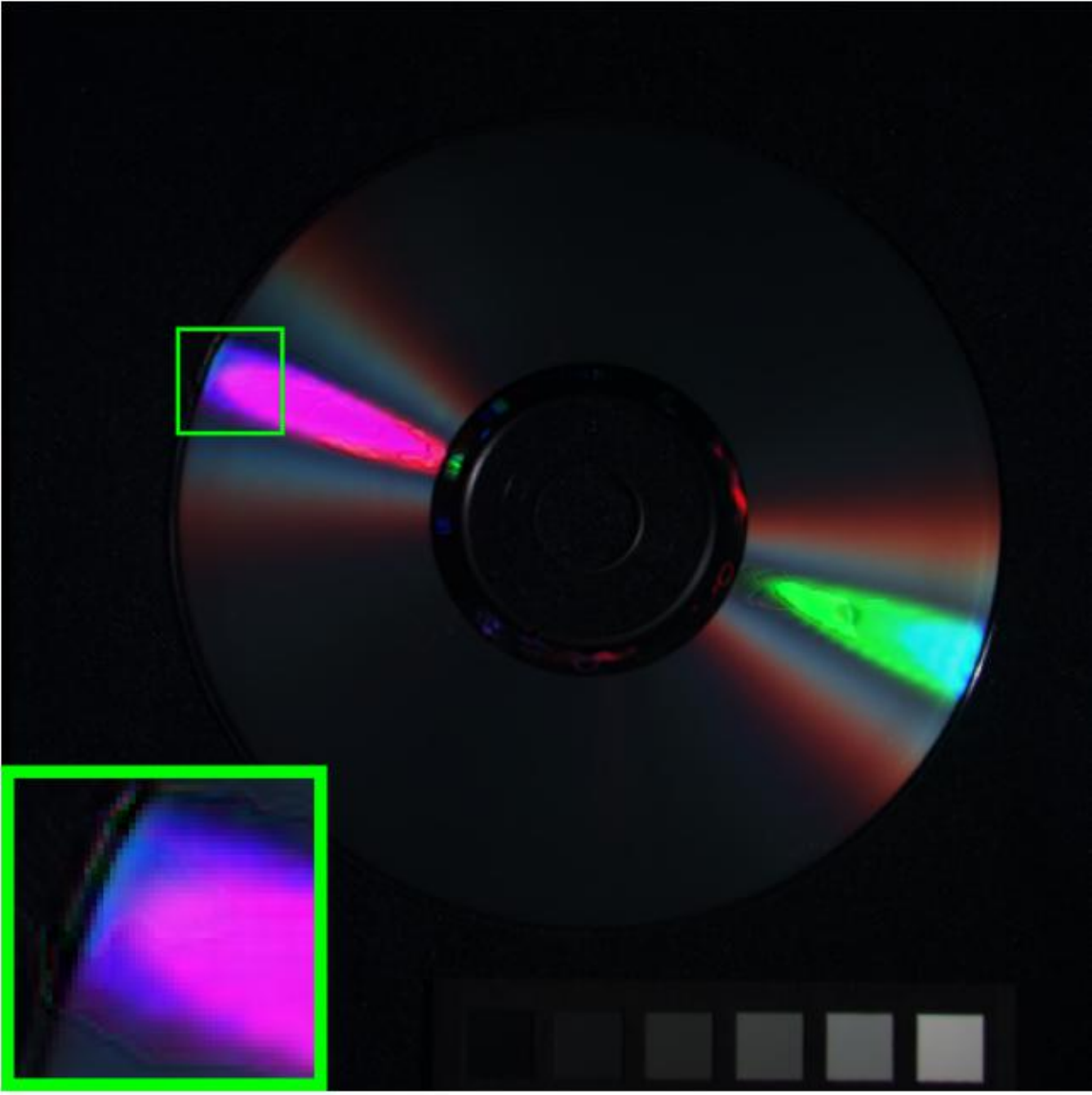}}
							\vspace{2pt}
							{\includegraphics[width=1\linewidth]{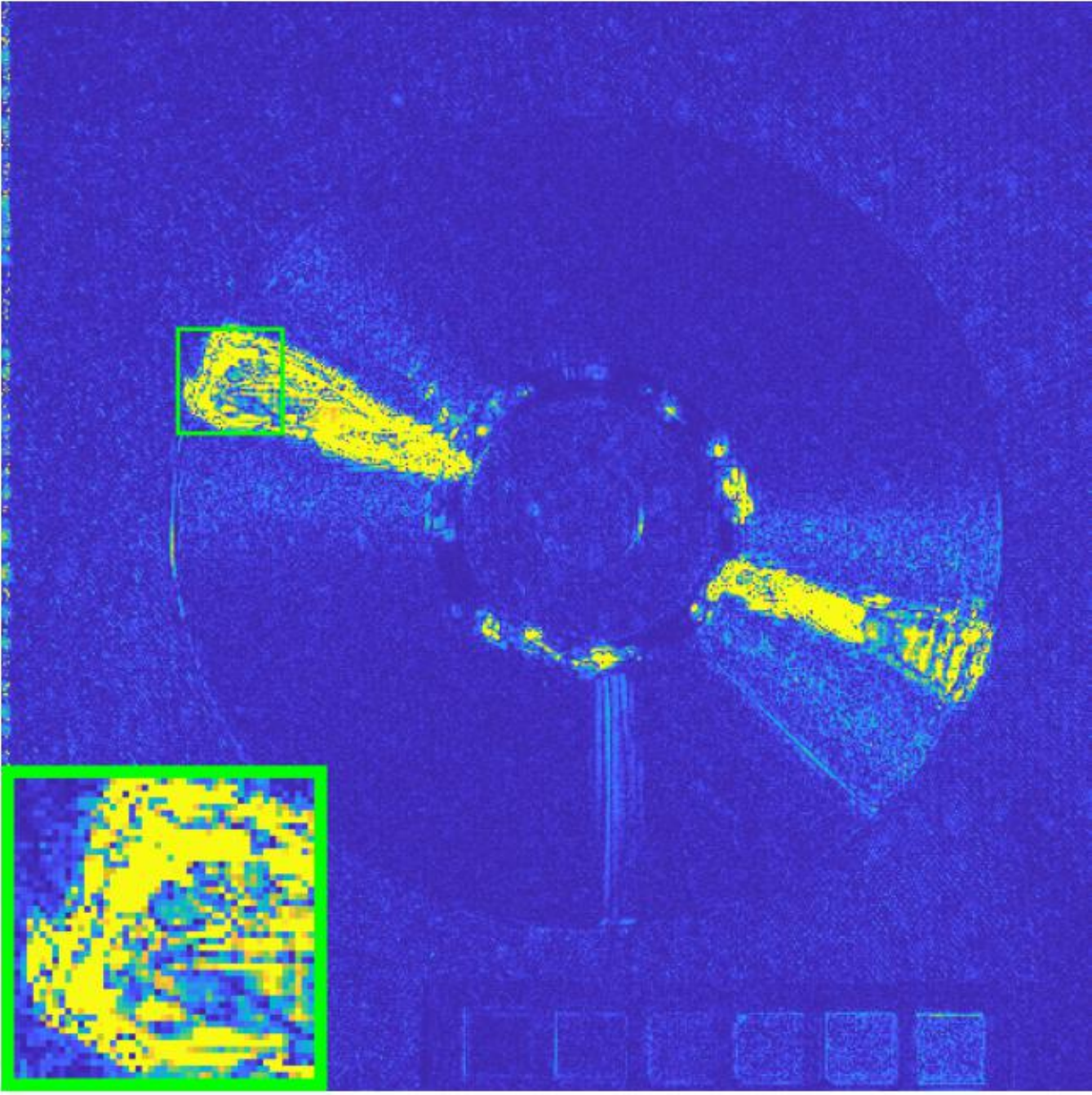}}
							{\includegraphics[width=1\linewidth]{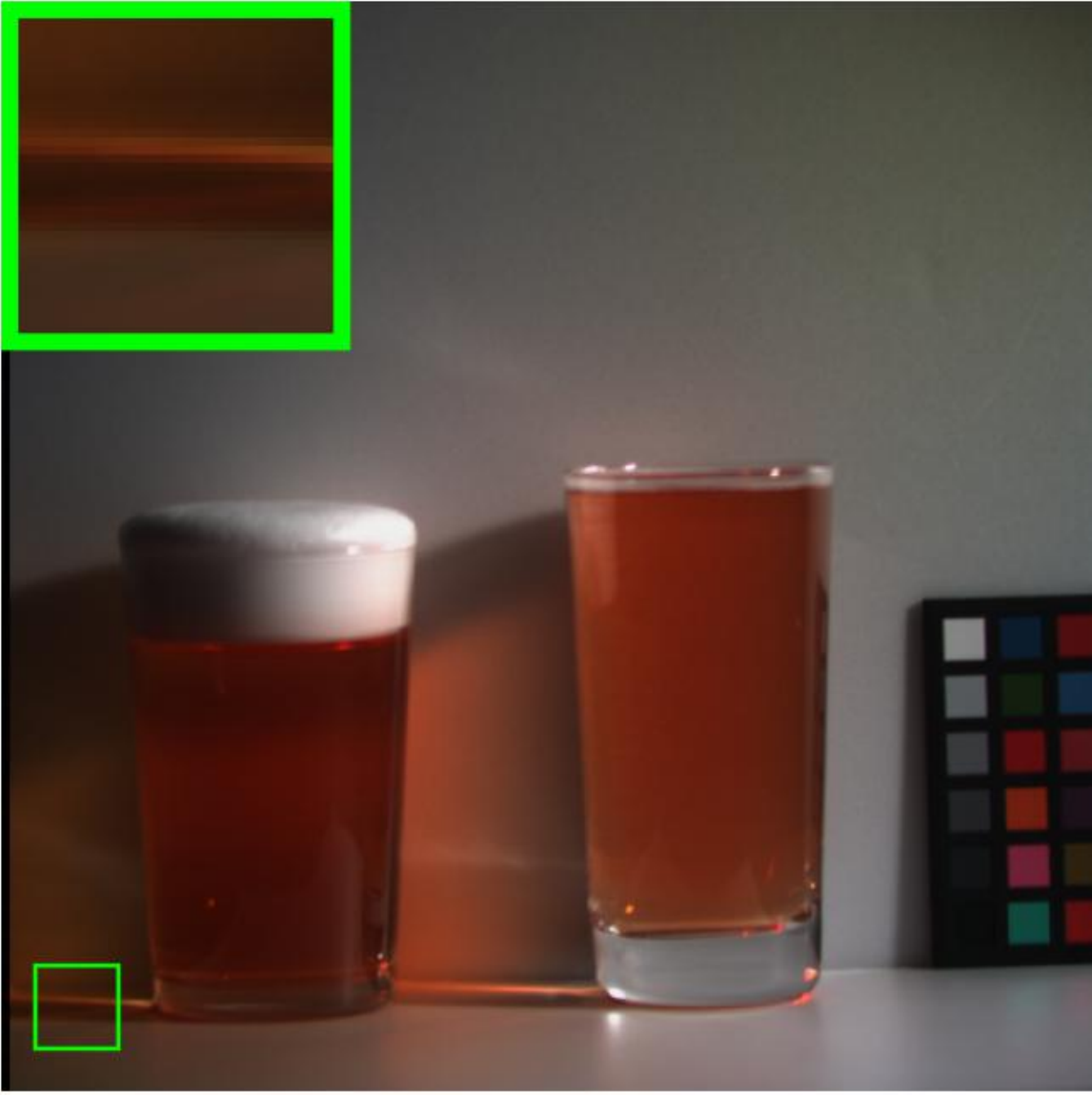}}
							{\includegraphics[width=1\linewidth]{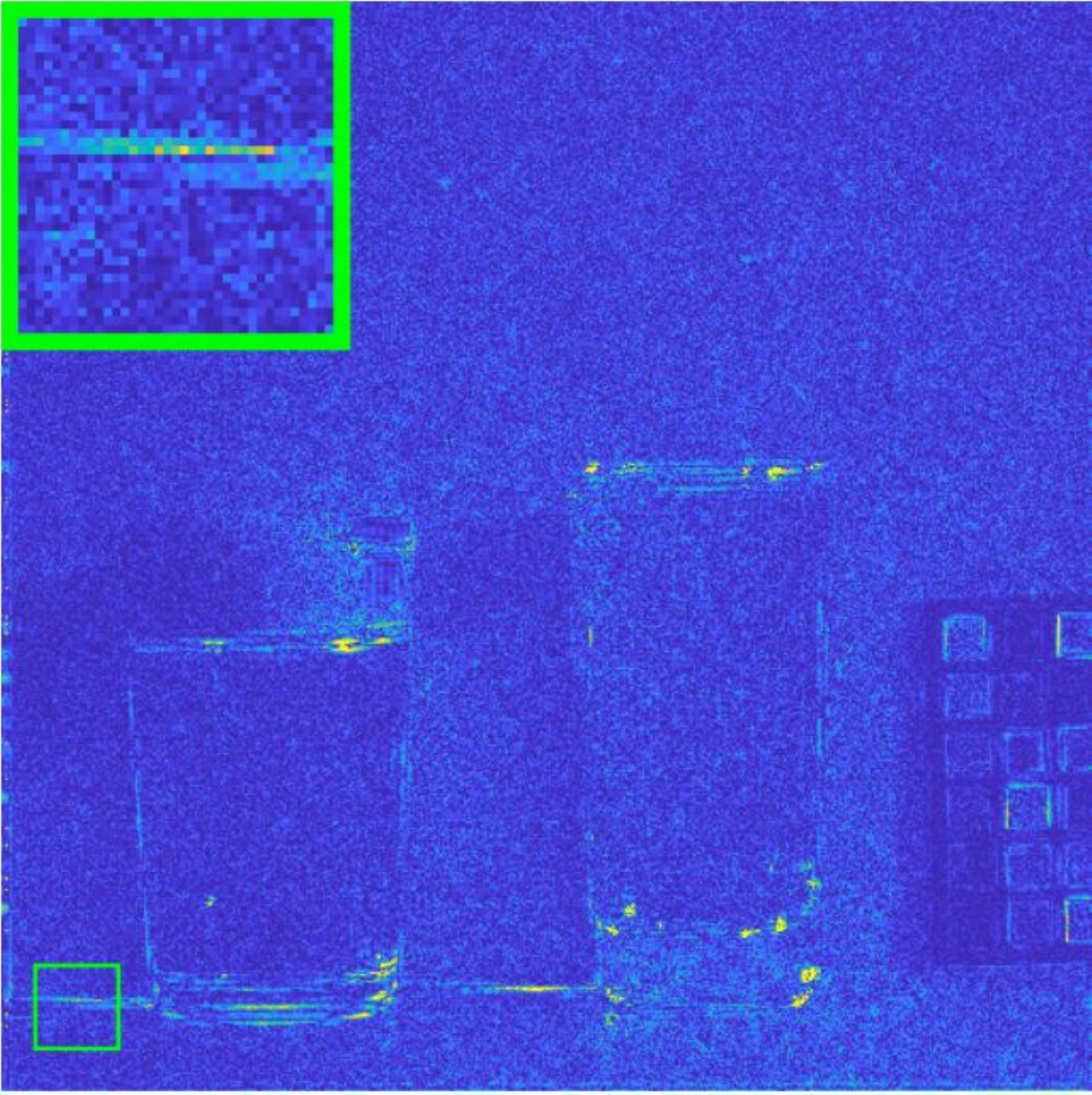}}
							\vspace{2pt}
							\scriptsize{3DT-Net}
							\centering
							
					\end{minipage}
					\begin{minipage}[t]{0.155\linewidth}
							{\includegraphics[width=1\linewidth]{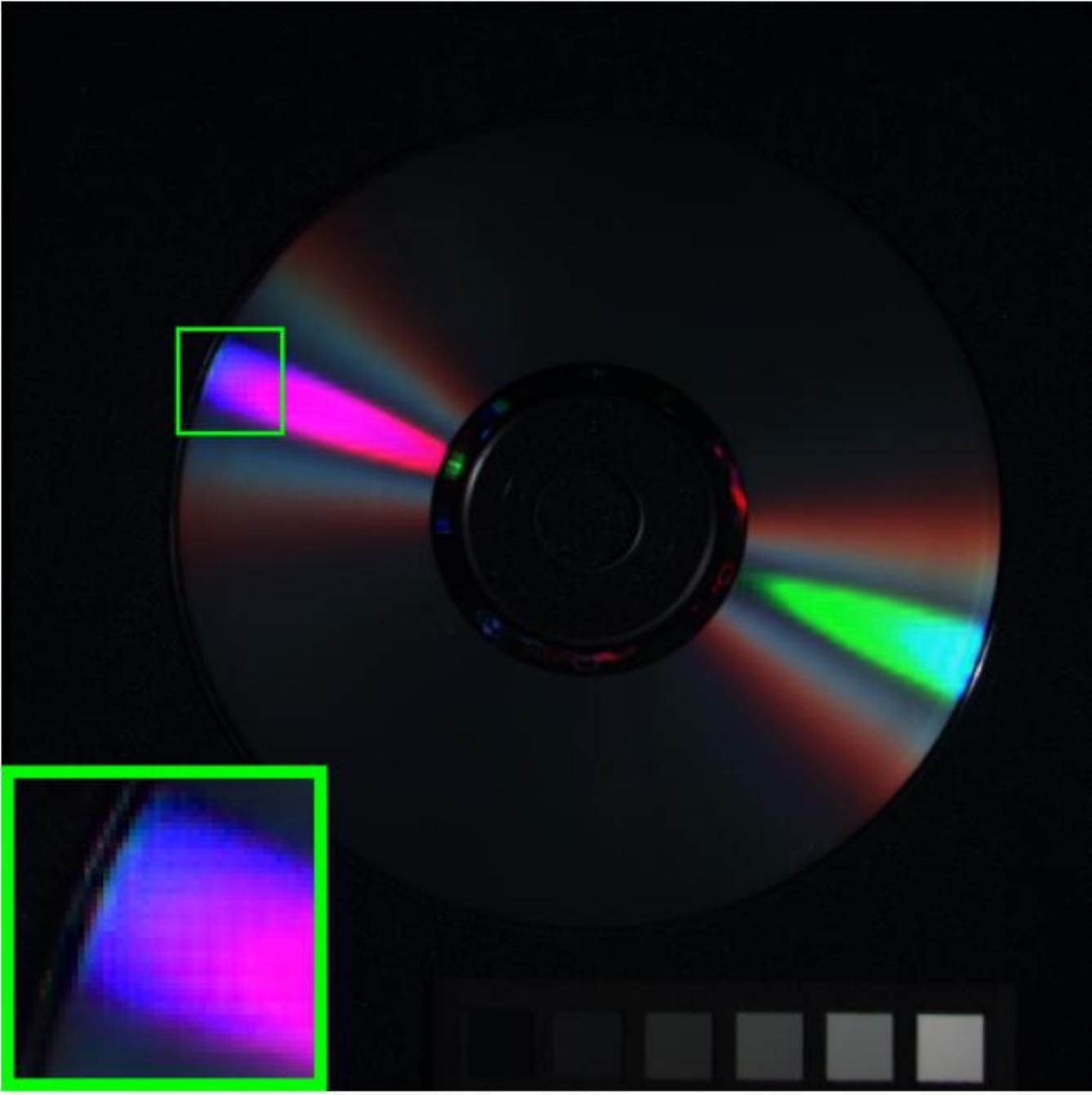}}
							\vspace{2pt}
							{\includegraphics[width=1\linewidth]{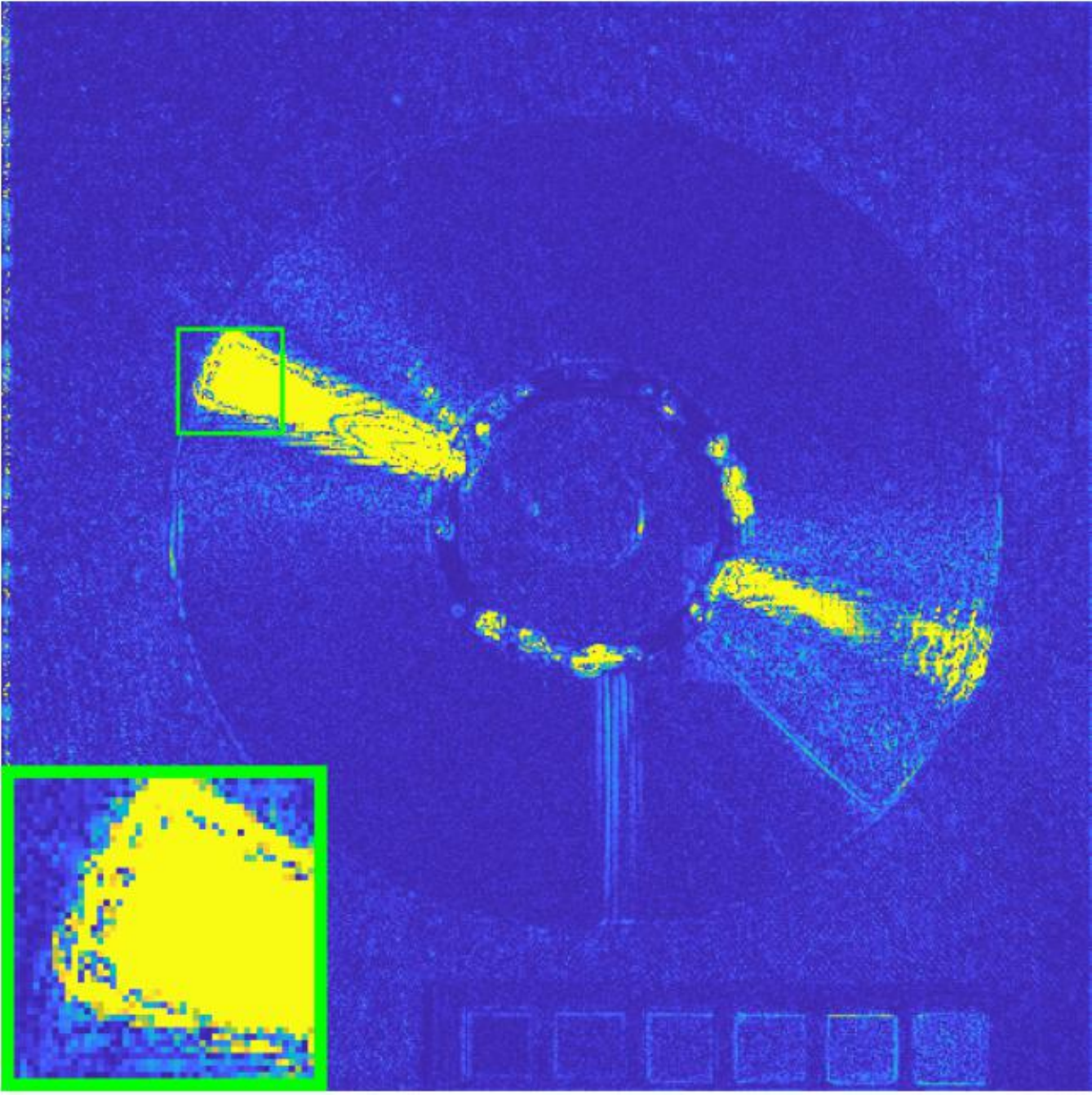}}
							{\includegraphics[width=1\linewidth]{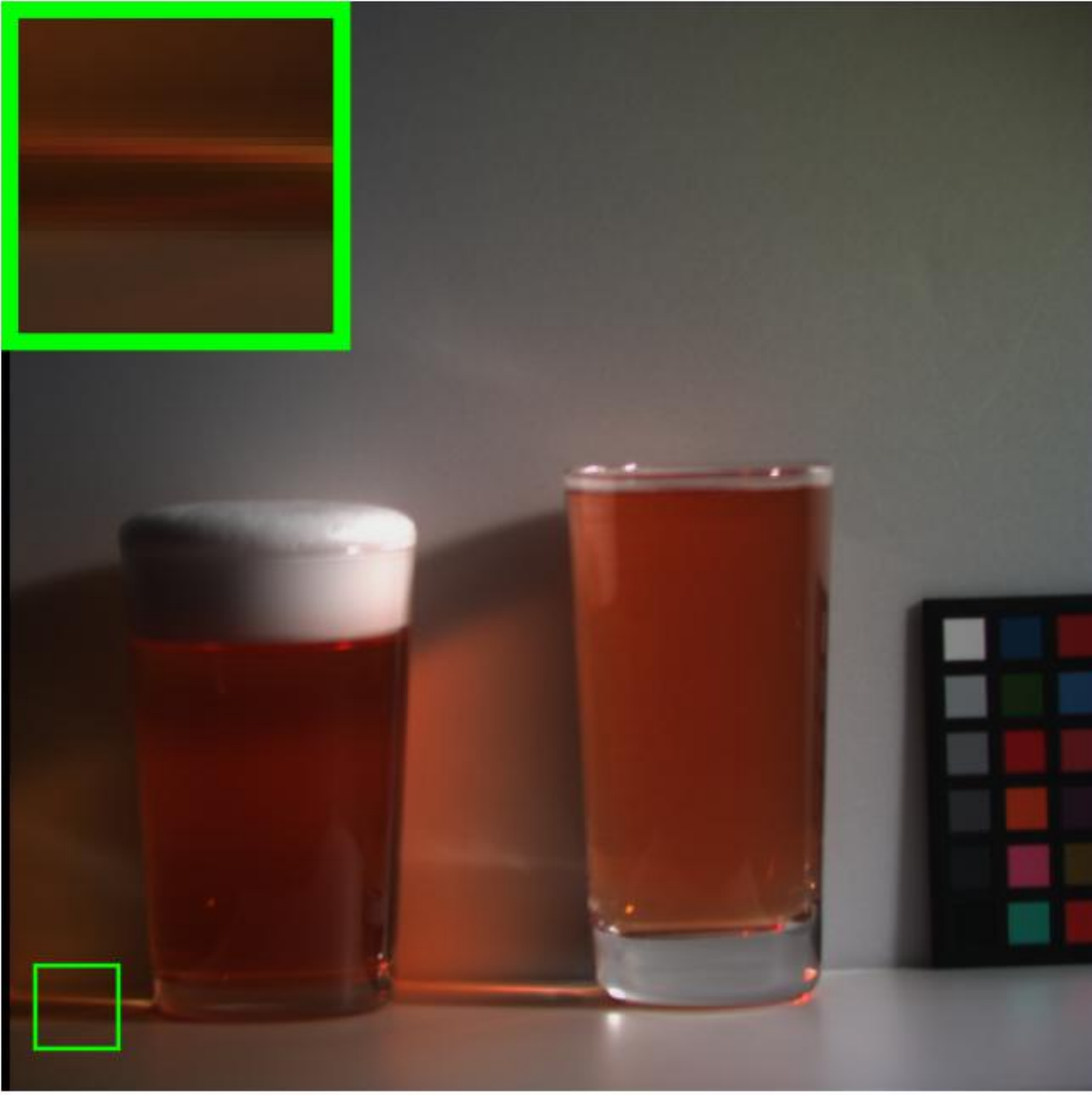}}
							{\includegraphics[width=1\linewidth]{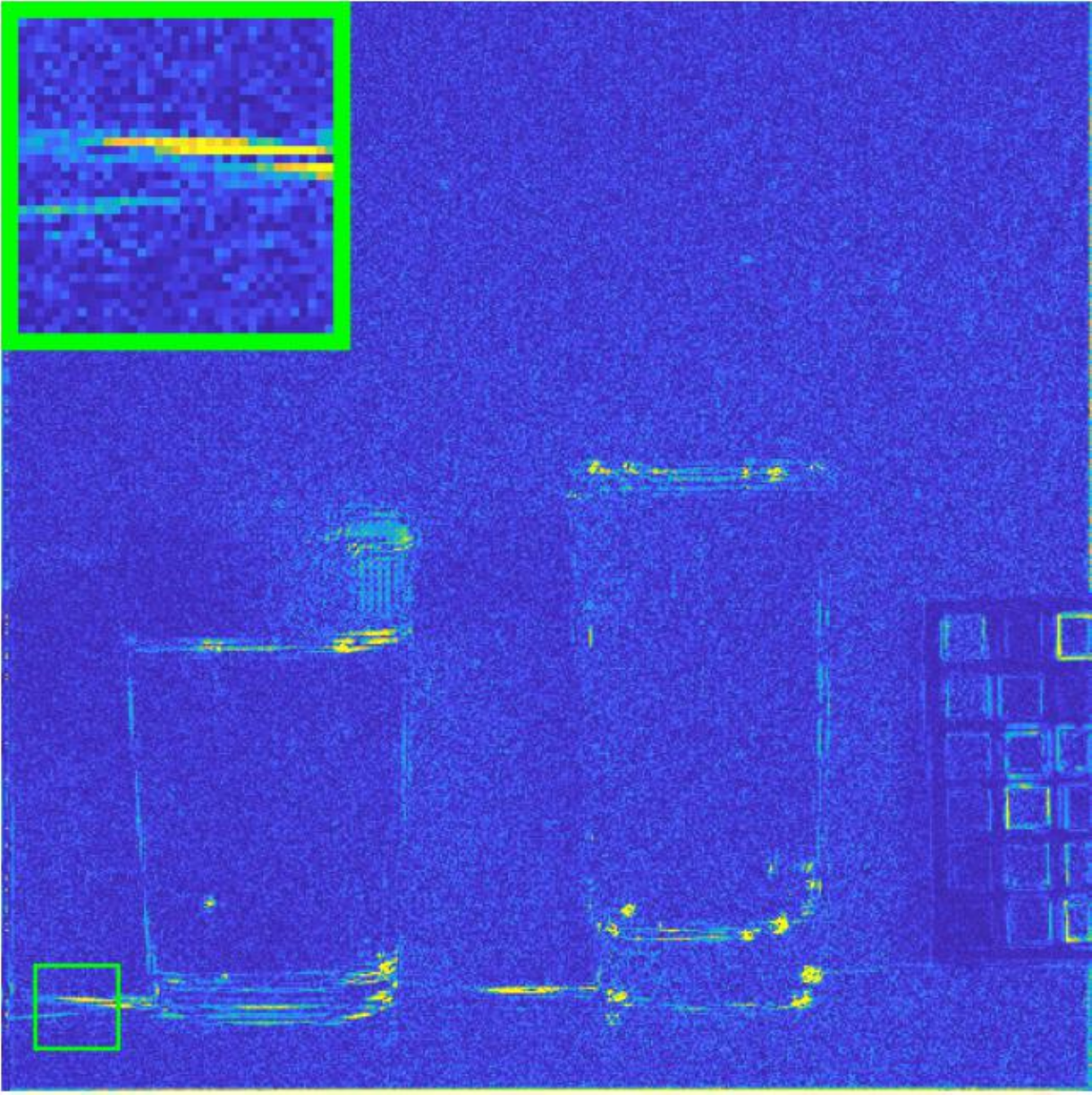}}
							\vspace{2pt}
							\scriptsize{U2Net}
							\centering
							
						\end{minipage}
					\begin{minipage}[t]{0.155\linewidth}
							{\includegraphics[width=1\linewidth]{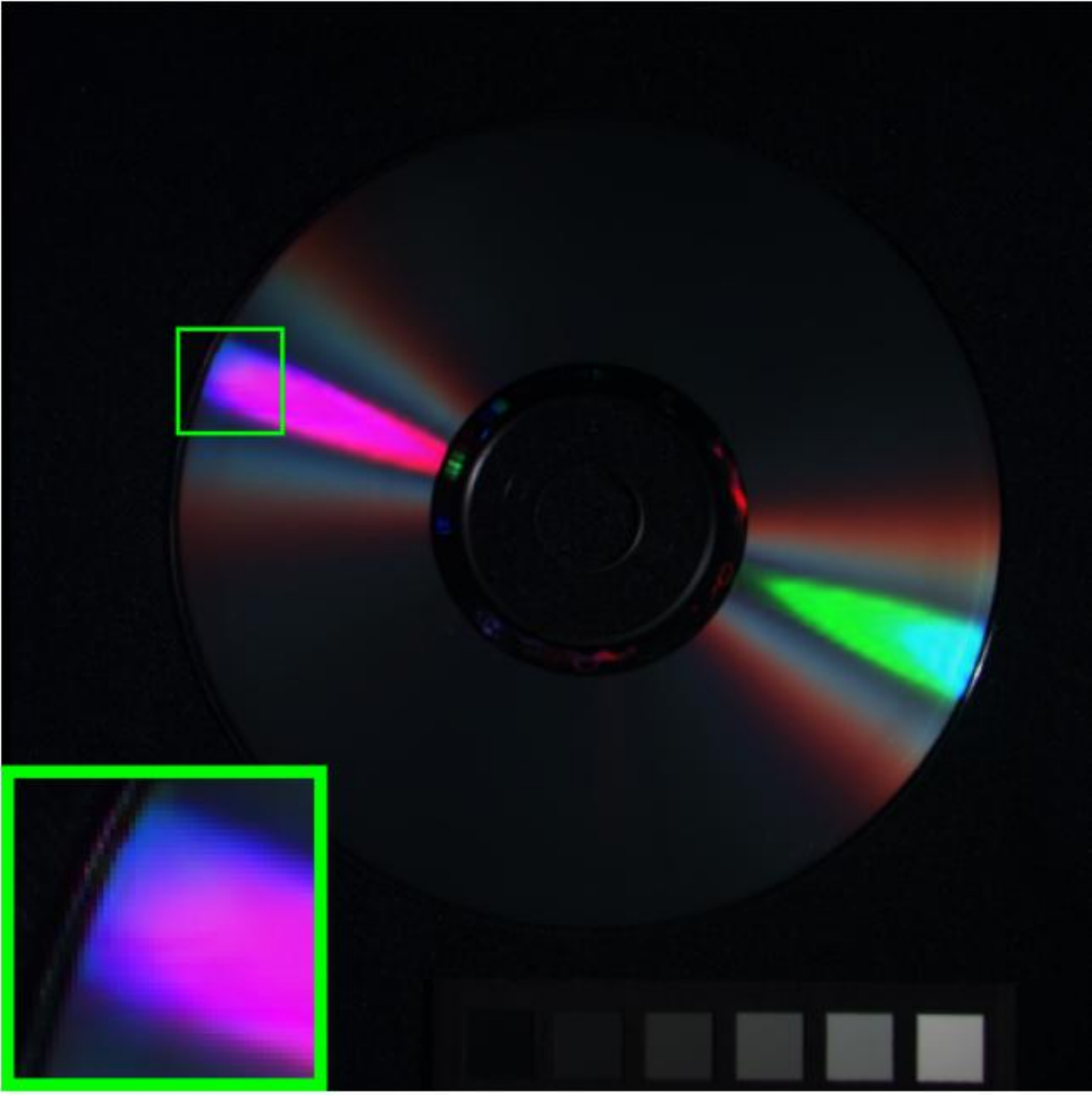}}
							\vspace{2pt}
							{\includegraphics[width=1\linewidth]{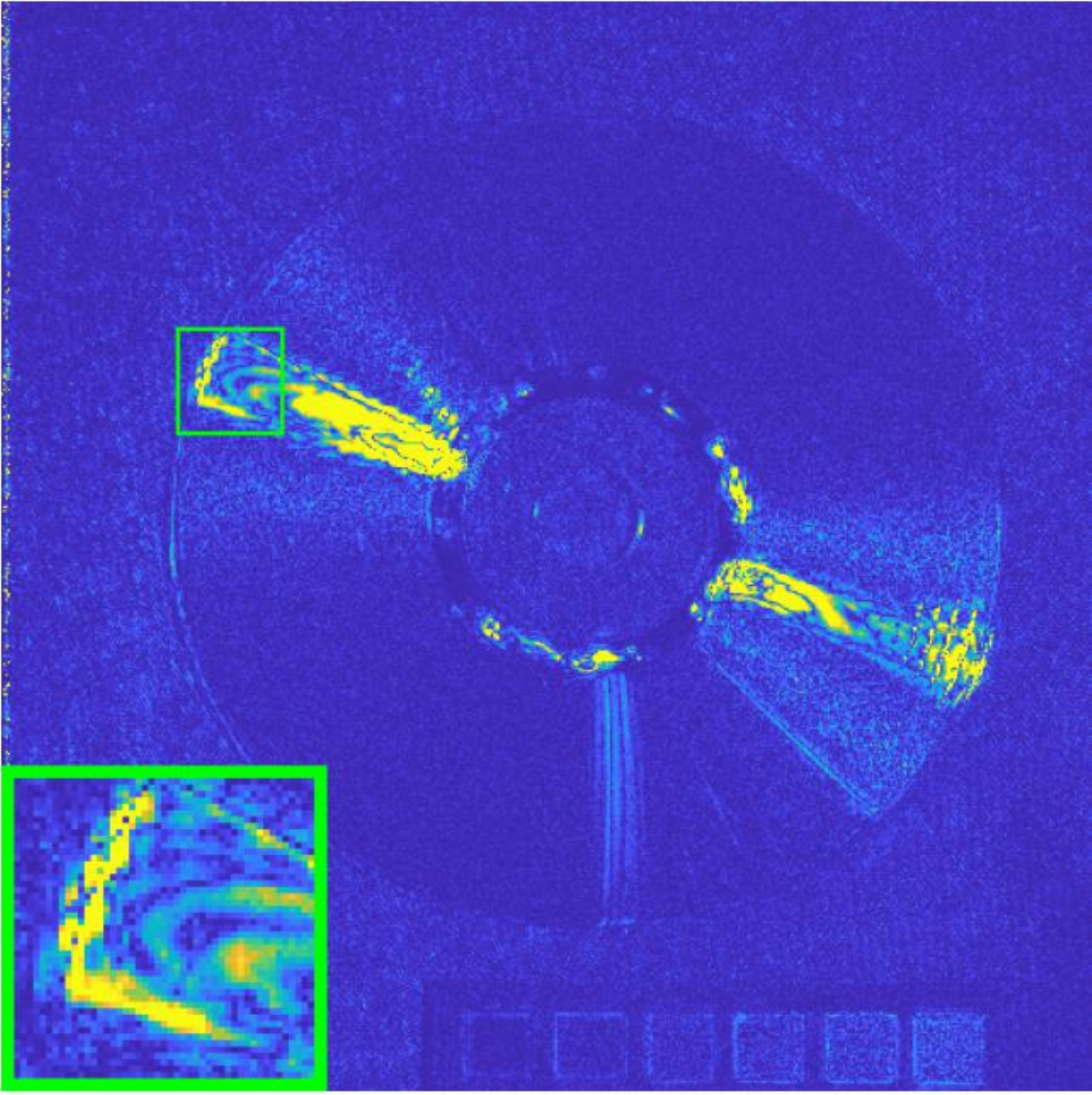}}
							{\includegraphics[width=1\linewidth]{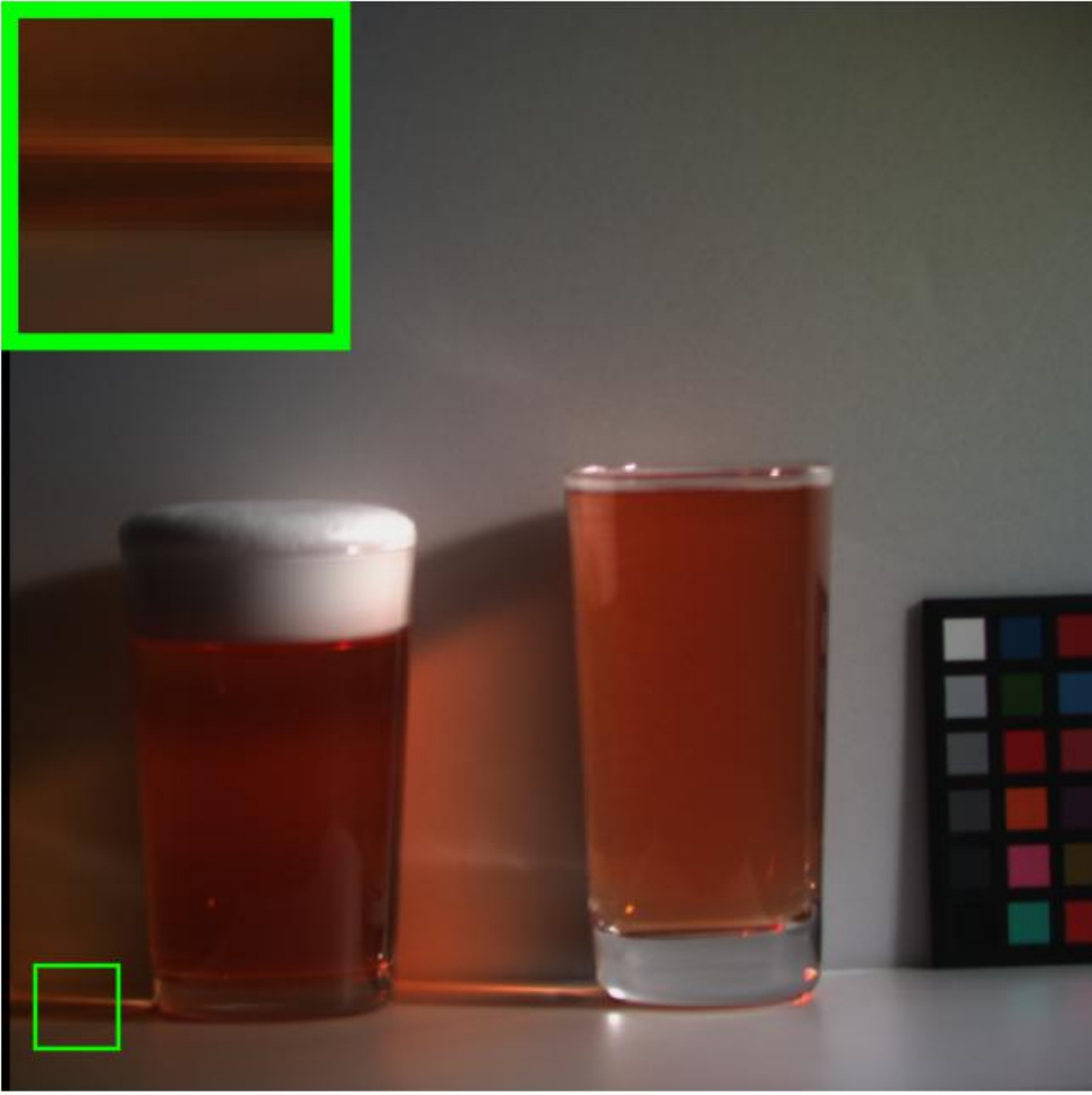}}
							{\includegraphics[width=1\linewidth]{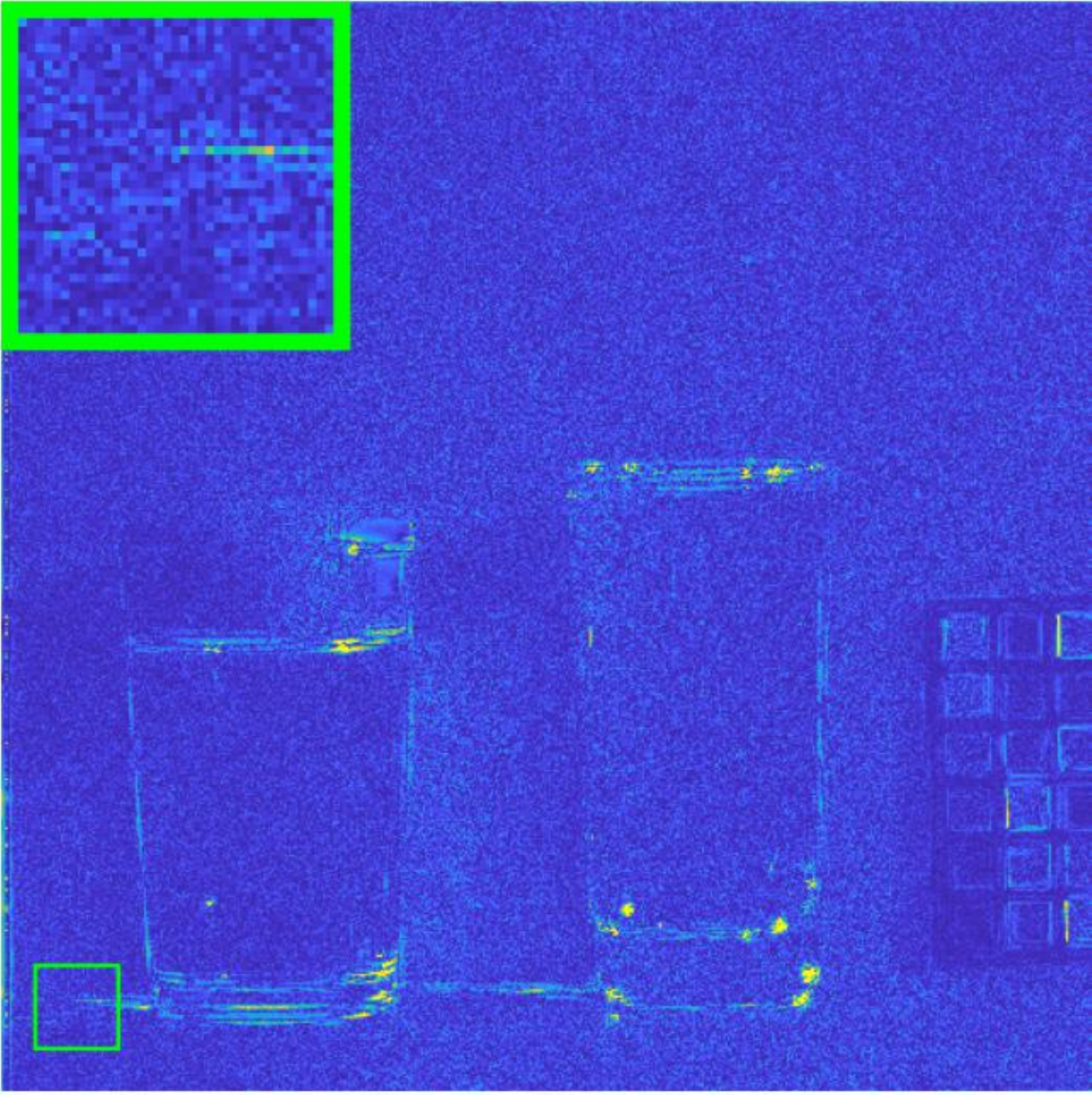}}
							\vspace{2pt}
							\scriptsize{Ada3D}
							\centering
							
						\end{minipage}
					\begin{minipage}[t]{0.155\linewidth}
							{\includegraphics[width=1\linewidth]{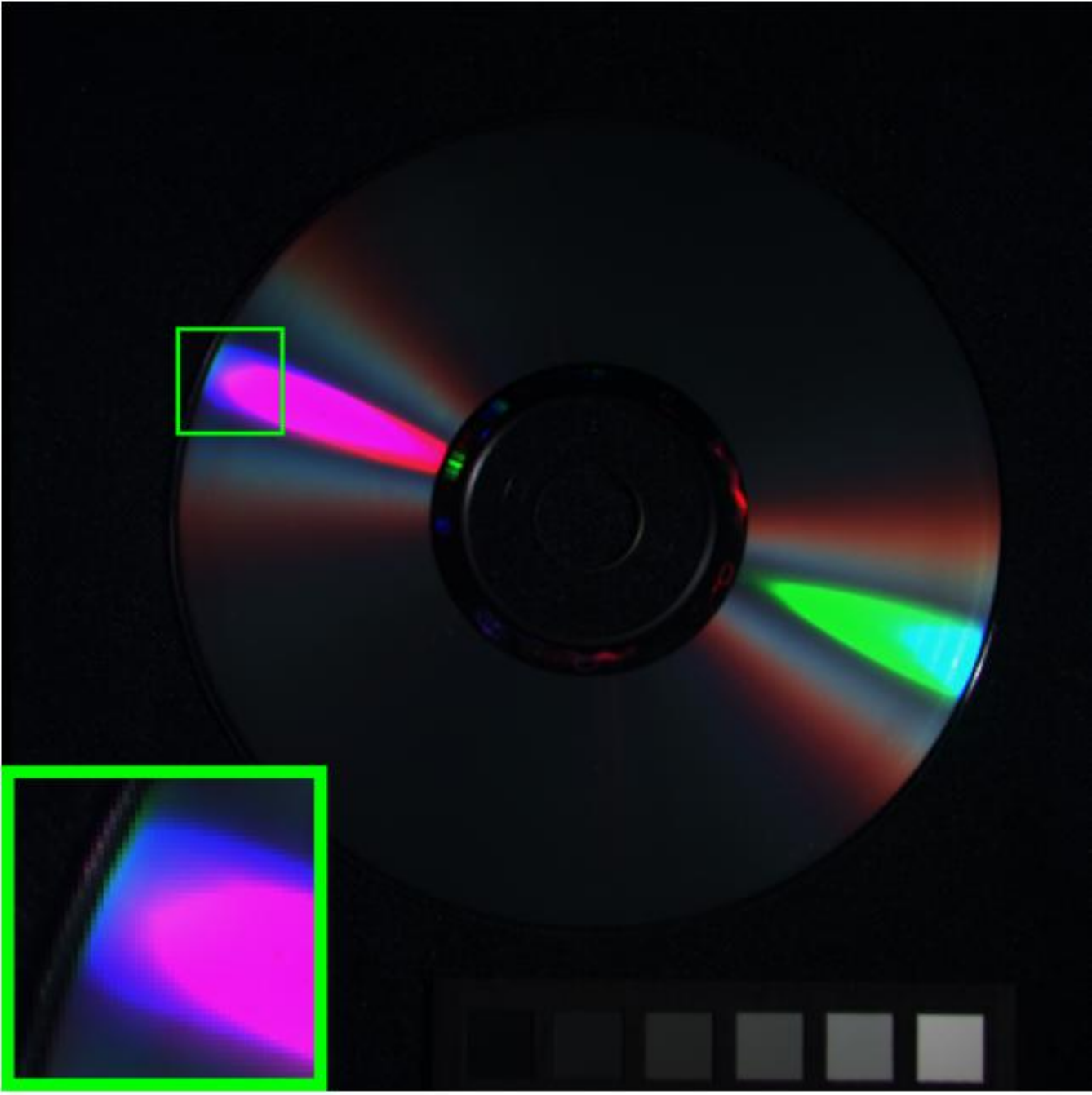}}
							\vspace{2pt}
							{\includegraphics[width=1\linewidth]{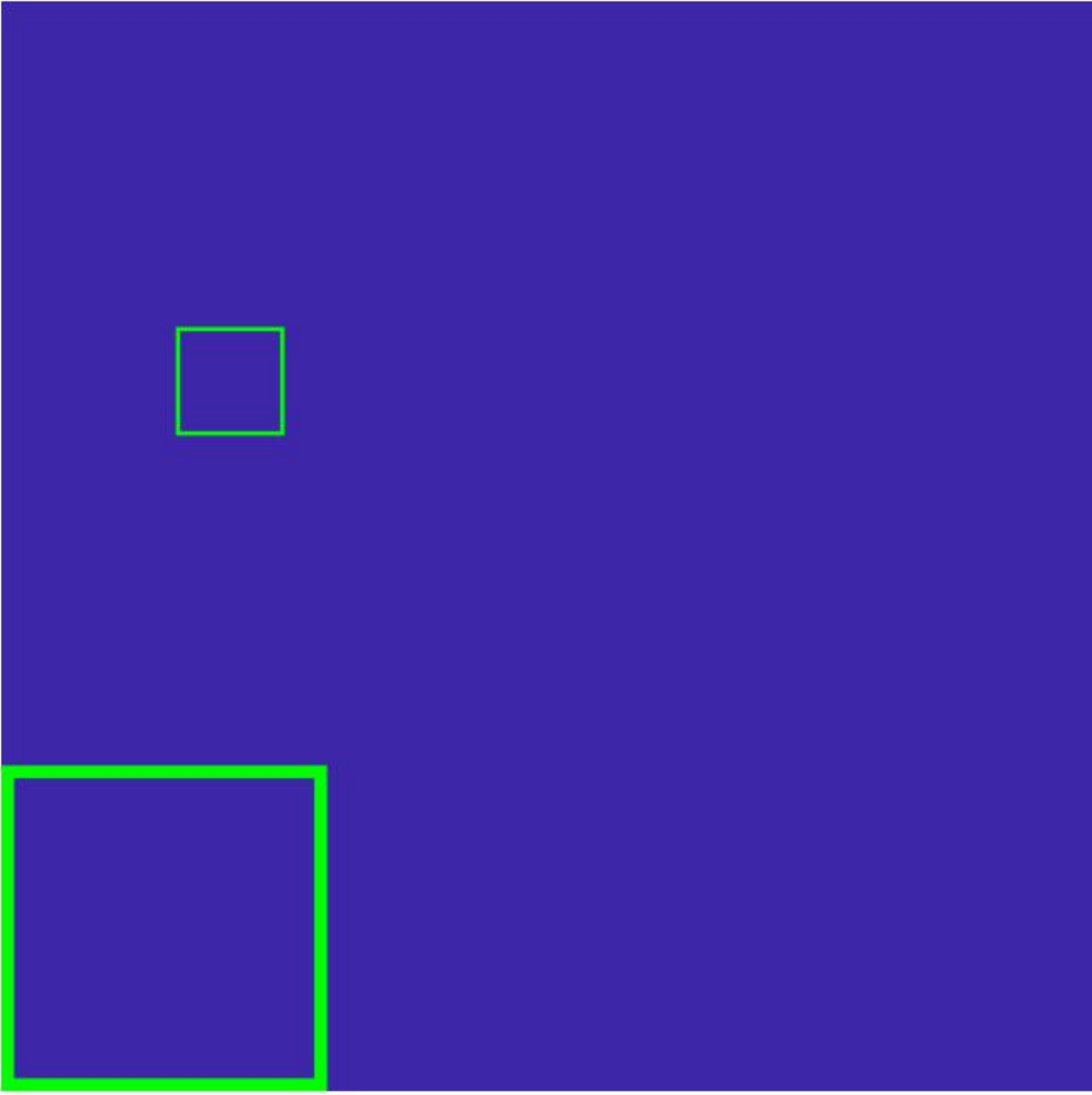}}
							{\includegraphics[width=1\linewidth]{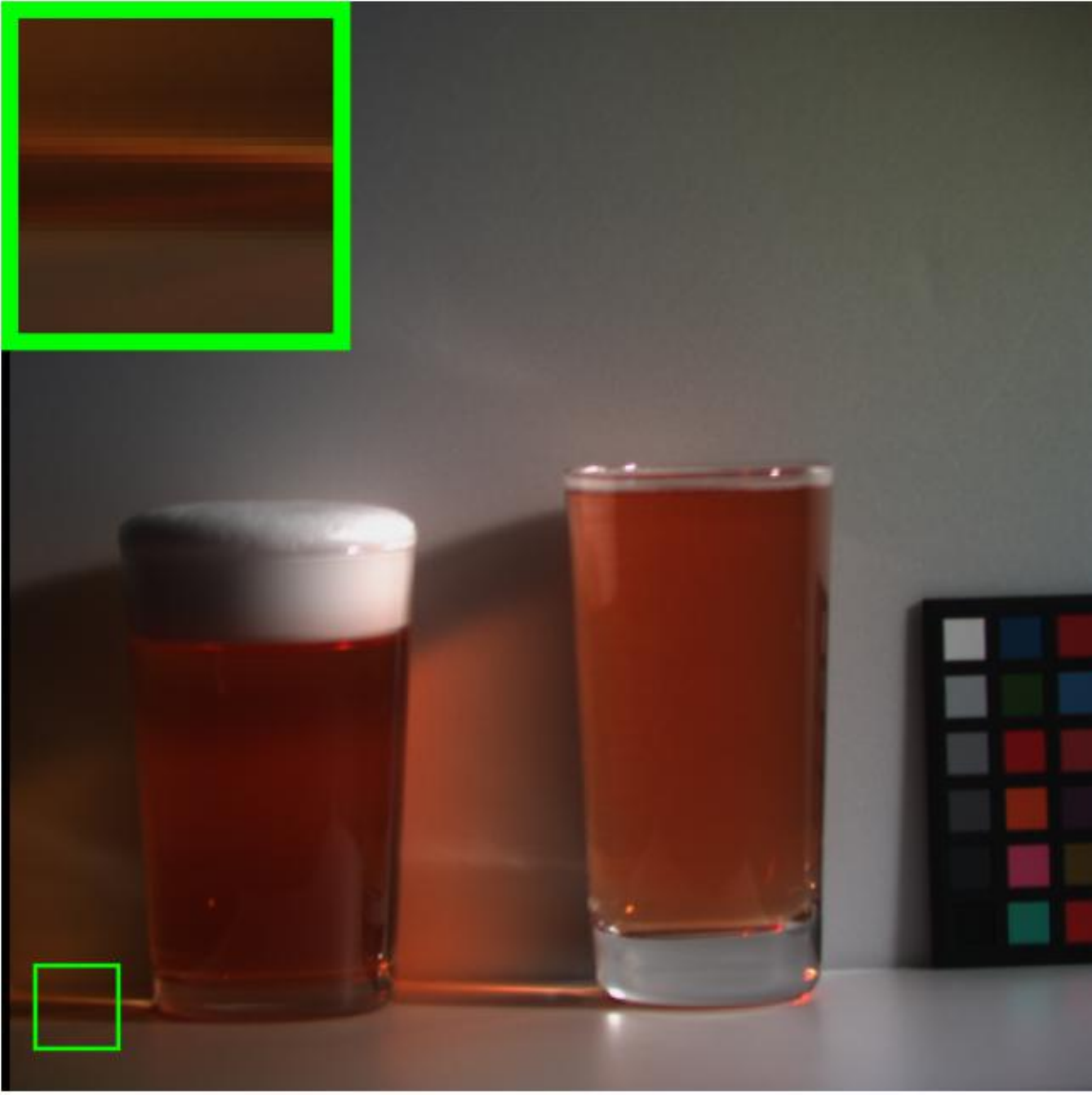}}
							{\includegraphics[width=1\linewidth]{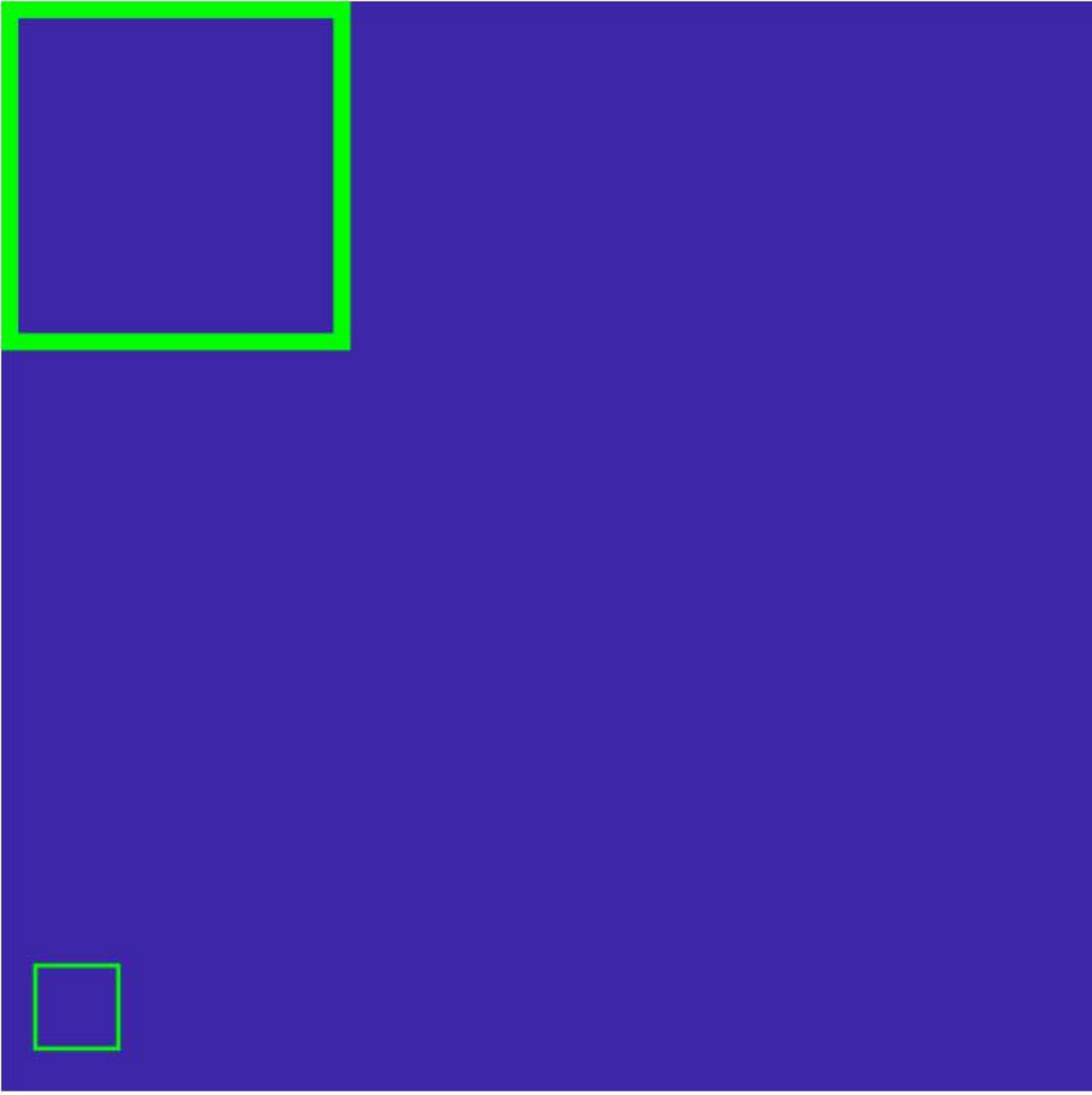}}
							\vspace{2pt}
							\scriptsize{GT}
							\centering
							
						\end{minipage}
				\end{minipage}
			\begin{minipage}[t]{0.97\linewidth}
					{\includegraphics[width=1\linewidth]{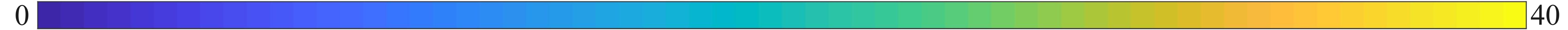}}
				\end{minipage}
		\end{center}
	\vspace{-8pt}
	\caption{Qualitative evaluation results of SOTA HISR methods on two testing samples from the CAVE dataset. Row 1: Pseudo-color images of spectral bands 6, 10, and 26 for the testing example \emph{cd}. Row 2: AEMs of spectral band 26 for \emph{cd}. Row 3: Pseudo-color images of spectral bands 6, 13, and 26 for the testing example \emph{bears}. Row 4: AEMs of spectral band 20 for \emph{bears}. \label{cave_v}}
\end{figure}

\section{Discussions}
\label{s6}
\subsection{Experiments for HISR}
\subsubsection{Dataset}
To assess the effectiveness of Ada3D for image fusion tasks outside the realm of remote sensing, we conduct experiments using the CAVE dataset\footnote{\url{https://www.cs.columbia.edu/CAVE/databases/multispectral/}}, commonly employed for the HISR task \cite{10194239}. The CAVE dataset, initially introduced in \cite{2010Generalized}, contains 32 pairs of RGB and HRHS images, each with dimensions of $512 \times 512\times 3$ and $512 \times 512\times 31$, respectively. These images are not directly suitable for training and testing purposes. Following the data processing approach described in \cite{10.1145/3581783.3612084}, we transform this dataset into 3920 RGB/LRHS/GT training samples with sizes of $64 \times 64\times 3$, $16 \times 16\times 31$, and $64 \times 64\times 31$, respectively. Additionally, we create 11 RGB/LRHS/GT testing samples with dimensions of $512 \times 512\times 3$, $128 \times 128\times 31$, and $512 \times 512\times 31$, respectively.

\subsubsection{Benchmarks, Quality Indices, \& Settings}
We compare Ada3D with six DL-based methods for HISR: ResTFNet \cite{2018Remote}, SSRNet \cite{9186332}, MoG-DCN \cite{9429905}, Fusformer \cite{9841513}, 3DT-Net \cite{ma2023learning}, and U2Net \cite{10.1145/3581783.3612084}. To align with the research standards of HISR, we select four quality indices for evaluation, namely PSNR, SSIM, SAM, and ERGAS. The ideal values for these indices are $+\infty$, 1, 0, and 0. For detailed network configurations and training settings, please refer to the supplementary material.

\subsubsection{Results} 
The quantitative evaluation results in Table~\ref{cave} demonstrate that our method consistently surpasses other approaches across all quality indices. Additionally, the qualitative assessment shown in Fig.~\ref{cave_v} illustrates that Ada3D produces fusion outputs that most closely align with the GT. These findings provide compelling evidence of Ada3D's effectiveness in image fusion tasks beyond the domain of remote sensing.

\begin{table}[t]
	\centering\renewcommand\arraystretch{1.}\setlength{\tabcolsep}{2.6pt}
	\belowrulesep=0pt\aboverulesep=0pt
	\caption{Comparison of strengths across different convolutions. \label{strength}}
	\begin{tabular}{c|c|c|c}
		\toprule
		{Convolutions} & {Spectral Preservation} & {Content Awareness} & {High Efficiency} \\  
		\midrule
		{Standard 2D} & \ding{55} & \ding{55} & \ding{51} \\
		{Adaptive 2D} & \ding{55} & \ding{51} & \ding{51} \\
		{Standard 3D} & \ding{51} & \ding{55} & \ding{55} \\
		{PAC3D} & \ding{51} & \ding{51} & \ding{55} \\
		{Ada3D (Ours)} & \ding{51} & \ding{51} & \ding{51} \\
		\bottomrule
	\end{tabular}
\end{table}

\subsection{Strengths \& Limitations}
\subsubsection{Strengths}
As shown in Table \ref{strength}, our method surpasses other convolutional paradigms by simultaneously achieving spectral preservation, content awareness, and high efficiency. Ada3D preserves spectral information by treating image fusion as a 3D problem. Additionally, its content awareness is enhanced through the use of adaptive convolution. Moreover, Ada3D achieves high efficiency with its carefully designed kernel generator and the implementation of group convolution.

\subsubsection{Limitations}
Our method exhibits two primary limitations. First, Ada3D utilizes the group convolution technique, which constrains the output feature map to have the same number of channels as one of the input feature maps. Second, Ada3D requires a 3D input, making it unsuitable for applications involving feature extraction from a single 2D input.

\section{Conclusion}
\label{s7}
In this paper, we mathematically demonstrate that treating image fusion as a 2D problem can lead to significant spectral distortions. We further show that adaptive convolution is superior to standard convolution in data processing. Based on these insights, we propose a novel convolutional paradigm called Ada3D for remote sensing image fusion. Ada3D assigns a unique set of 3D kernels to each input voxel, enabling the capture of fine-grained details. These adaptive kernels are generated through a two-step process that integrates both spatial and spectral information. In parallel, we produce adaptive biases to refine the convolutional outcome. Additionally, group convolution is employed to reduce computational complexity. We assess the performance of Ada3D on five datasets related to three image fusion tasks. Our method achieves SOTA results, highlighting the superiority of Ada3D for image fusion.

\section*{Acknowledgments}
This research is supported by National Natural Science Foundation of China (12271083), and the Project of the Department of Science and Technology of Sichuan Province (Grant No. 2025YFNH0001).

\bibliographystyle{IEEEtran}
\bibliography{reference}

\newpage
\appendices

\twocolumn[
  \begin{center}
    \Huge Supplementary Material
    \vspace{1em}
  \end{center}
]

\section*{I. Supplementary Explanation for Remark~\ref{th1}}
First, we extend Definition~\ref{d2} into matrix space as follows. 
\begin{myLem}	
	\label{l1} 
	Let $\varphi$ be a linear map from an $n$-dimensional matrix space $V$ to an $m$-dimensional matrix space $U$. Suppose $(\boldsymbol{A}_1, \boldsymbol{A}_2, \cdots, \boldsymbol{A}_n)$ is a basis for $V$ and $(\boldsymbol{B}_1, \boldsymbol{B}_2, \cdots, \boldsymbol{B}_m)$ is a basis for $U$. For each basis matrix $\boldsymbol{A}_i$ in $V$, its image under $\varphi$, denoted by $\varphi(\boldsymbol{A}_i)$, can be written as a unique linear combination of the basis matrices in $U$. Explicitly, we have:
	\begin{equation}
		\label{le1}
		\footnotesize
		\setlength{\arraycolsep}{0.5pt}
		(\varphi(\boldsymbol{A}_1),\varphi(\boldsymbol{A}_2),\cdots,\varphi(\boldsymbol{A}_n)) = (\boldsymbol{B}_1, \boldsymbol{B}_2, \cdots, \boldsymbol{B}_m)
		\begin{pmatrix}
			t_{11} & t_{12} & \cdots & t_{1n} \\ 
			t_{21} & t_{22} & \cdots & t_{2n} \\ 
			\vdots & \vdots &        & \vdots \\ 
			t_{m1} & t_{m2} & \cdots & t_{mn}
		\end{pmatrix}.
	\end{equation}
	We define the $m\times n$ matrix on the right as $\boldsymbol{T}$, representing $\varphi$ in terms of two given bases. Let $\boldsymbol{\lambda}={(\lambda_1,\lambda_2,\cdots,\lambda_n)}^\intercal$ be a coordinate vector based on $(\boldsymbol{A}_1, \boldsymbol{A}_2, \cdots, \boldsymbol{A}_n)$. The corresponding coordinate vector $\boldsymbol{\mu}={(\mu_1,\mu_2,\cdots,\mu_m)}^\intercal$ based on $(\boldsymbol{B}_1, \boldsymbol{B}_2, \cdots, \boldsymbol{B}_m)$ after the linear map is expressed as:
	\begin{equation}
		\label{le2}
		\boldsymbol{\mu} = \boldsymbol{T}\boldsymbol{\lambda}.
	\end{equation}
\end{myLem}

\noindent\emph{{Proof.}} To start, we rewrite Equation \ref{le1} in the following form:
\begin{equation}
	\small
	\begin{cases}
		\varphi(\boldsymbol{A}_1)= t_{11}\boldsymbol{B}_1+t_{21}\boldsymbol{B}_2+\cdots+t_{m1}\boldsymbol{B}_m=\sum\limits_{{i=1}}^{m}t_{i1}\boldsymbol{B}_i \\
		\varphi(\boldsymbol{A}_2)= t_{12}\boldsymbol{B}_1+t_{22}\boldsymbol{B}_2+\cdots+t_{m2}\boldsymbol{B}_m=\sum\limits_{{i=1}}^{m}t_{i2}\boldsymbol{B}_i \\
		\qquad\qquad\qquad\qquad\qquad\cdots\cdots \\
		\varphi(\boldsymbol{A}_n)= t_{1n}\boldsymbol{B}_1+t_{2n}\boldsymbol{B}_2+\cdots+t_{mn}\boldsymbol{B}_m=\sum\limits_{{i=1}}^{m}t_{in}\boldsymbol{B}_i \\
	\end{cases}.
\end{equation} 
Therefore, the image of the matrix $\boldsymbol{A}=\lambda_1\boldsymbol{A}_1+\lambda_2\boldsymbol{A}_2+\cdots+\lambda_n\boldsymbol{A}_n$ can be expressed as follows:
\begin{equation}
	\footnotesize
	\begin{aligned}
		\varphi(\boldsymbol{A}) &= \lambda_1\varphi(\boldsymbol{A}_1)+\lambda_2\varphi(\boldsymbol{A}_2)+\cdots+\lambda_n\varphi(\boldsymbol{A}_n) \\
		&= \lambda_1(\sum\limits_{{i=1}}^{m}t_{i1}\boldsymbol{B}_i)+\lambda_2(\sum\limits_{{i=1}}^{m}t_{i2}\boldsymbol{B}_i)+\cdots+\lambda_n(\sum\limits_{{i=1}}^{m}t_{in}\boldsymbol{B}_i) \\
		&=(\sum\limits_{{j=1}}^{n}\lambda_jt_{1j})\boldsymbol{B}_1+(\sum\limits_{{j=1}}^{n}\lambda_jt_{2j})\boldsymbol{B}_2+\cdots+(\sum\limits_{{j=1}}^{n}\lambda_jt_{mj})\boldsymbol{B}_m.
	\end{aligned}
\end{equation}
Consequently, the corresponding coordinate vector $\boldsymbol{\mu}$ with respect to the basis $(\boldsymbol{B}_1, \boldsymbol{B}_2, \cdots, \boldsymbol{B}_m)$ is given by:
\begin{equation}
	\begin{pmatrix}
		\mu_{1} \\ 
		\mu_{2} \\ 
		\vdots \\ 
		\mu_{m}
	\end{pmatrix}=
	\begin{pmatrix}
		t_{11} & t_{12} & \cdots & t_{1n} \\ 
		t_{21} & t_{22} & \cdots & t_{2n} \\ 
		\vdots & \vdots &        & \vdots \\ 
		t_{m1} & t_{m2} & \cdots & t_{mn}
	\end{pmatrix}
	\begin{pmatrix}
		\lambda_{1} \\ 
		\lambda_{2} \\ 
		\vdots \\ 
		\lambda_{n}
	\end{pmatrix}.
\end{equation}
This equation represents the expanded form of Equation \ref{le2}. Therefore, Lemma \ref{l1} is mathematically verified.

Recall that we denote the spectral data and feature map as $\mathcal{X}\in\mathbb{R}^{H\times W\times L}$ and $\mathcal{Y}\in\mathbb{R}^{H\times W\times C}$, respectively. Let the kernel size of $f(\cdot)$ be $k\times k$. To formulate the function $f(\cdot)$ mathematically, we undertake the following transformation. According to the properties of 2D convolution, $\mathcal{Y}(i, j, \cdot)$ is determined by the vectors within a $k\times k$ spatial window centered at $\mathcal{X}(i, j, \cdot)$. For each window, we first organize these vectors into a $k\times k\times L$ tensor. We then apply mode-$3$ unfolding to this tensor, which produces an $L\times k^2$ matrix. This matrix is further flattened row-wise into an $Lk^2$ vector. Collecting these vectors from all windows results in an $H\times W\times (Lk^2)$ tensor, denoted as $\mathcal{\hat{X}}$. Finally, we conduct mode-$3$ unfolding on both $\mathcal{\hat{X}}$ and $\mathcal{Y}$, generating $\boldsymbol{X}\in\mathbb{R}^{(Lk^2)\times(HW)}$ and $\boldsymbol{Y}\in\mathbb{R}^{C\times(HW)}$, respectively. The columns of $\boldsymbol{X}$ can be viewed as coordinate vectors with respect to a basis $(\boldsymbol{A}_{1}, \boldsymbol{A}_{2}, \cdots, \boldsymbol{A}_{Lk^2})$, where each $\boldsymbol{A}_i\in\mathbb{R}^{L\times k^2}$. This basis spans a matrix space of dimension $Lk^2$. Besides, the columns of $\boldsymbol{Y}$ can be seen as coordinate vectors related to the standard basis. Here, we represent this basis in the form of column matrices as $(\boldsymbol{E}_{1}, \boldsymbol{E}_{2}, \cdots, \boldsymbol{E}_{C})$, where $\boldsymbol{E}_i\in\mathbb{R}^{C\times 1}$. 
Let $\mathcal{T}\in\mathbb{R}^{C\times L\times k\times k}$ denote the weights of $f(\cdot)$. We perform mode-$1$ unfolding on $\mathcal{T}$, resulting in a matrix represented as $\boldsymbol{T}\in\mathbb{R}^{C\times (Lk^2)}$. Therefore, Equation~\ref{pe1} in the main text can be reformulated as follows:
\begin{equation}
	\label{pe2}
	\boldsymbol{Y}=\boldsymbol{T}\boldsymbol{X}.
\end{equation}
According to Lemma \ref{l1}, $\boldsymbol{T}$ can be interpreted as the matrix representation of the linear map $\psi$, which is the abstract form of $f(\cdot)$. $\psi$ maps spectral information into feature map channels with respect to $(\boldsymbol{A}_1,\boldsymbol{A}_2,\cdots,\boldsymbol{A}_{Lk^2})$ and $(\boldsymbol{E}_1,\boldsymbol{E}_2,\cdots,\boldsymbol{E}_C)$. Therefore, $\psi(\boldsymbol{A}_{i})$ is the column matrix that corresponds to the $i$-th column of $\boldsymbol{T}$. According to Definition~\ref{d3}, the dimension of the matrix space spanned by ($\boldsymbol{E}_{1}, \boldsymbol{E}_{2}, \cdots, \boldsymbol{E}_{C}$) is equal to ${\rm{rank}}(\boldsymbol{T})$. 
To accurately capture spectral information without distortion, it is necessary to use at least $L$ base matrices from $(\boldsymbol{A}_1,\boldsymbol{A}_2,\cdots,\boldsymbol{A}_{Lk^2})$. Therefore, if ${\rm{rank}}(\boldsymbol{T})$ is less than $L$, a loss of spectral information is unavoidable.

To further support Remark~\ref{th1}, we consider a more practical scenario by introducing a normalization layer and a non-linear activation function. Thus, Equation~\ref{pe2} can be rewritten as:
\begin{equation}
	\label{pe3}
	\boldsymbol{Y}_{final}=\sigma({\rm{Norm}}(\boldsymbol{Y}))=\sigma({\rm{Norm}}(\boldsymbol{T}\boldsymbol{X})),
\end{equation}
where ${\rm{Norm}}(\cdot)$ denotes the normalization layer, and $\sigma(\cdot)$ represents the non-linear activation function. Here, $\boldsymbol{Y}_{final}\in\mathbb{R}^{C\times (HW)}$ is the final output. Importantly, ${\rm{Norm}}(\cdot)$ is an invertible affine transformation that rescales and centers $\boldsymbol{Y}$ without changing its size. As a result, it does not recover any spectral information lost during the convolution process. Additionally, $\sigma(\cdot)$ operates element-wise and introduces non-linear distortions or stretching to the input data. Consequently, it not only fails to recover the spectral information lost by the convolutional layer, but it may also exacerbate spectral distortions. Thus, neither the normalization layer nor the non-linear activation function can recover the spectral information lost by the convolution operation, and therefore, these components do not alter the conclusion drawn in Remark~\ref{th1}.

\section*{II. Supplementary Explanations for Remark~\ref{th2}}
Here, we provide detailed explanations of Remark~\ref{th2} under two additional conditions: (\emph{i}) when the input data and target output have sizes $H\times W\times C$ and $H\times W\times C^\prime$, respectively, where $C^\prime$ denotes the number of output channels; and (\emph{ii}) when the input data and target output have dimensions of $H\times W\times L\times C$ and $H\times W\times L\times C^\prime$, respectively.

In the first condition, we denote the input data and target output as $\mathcal{A}\in\mathbb{R}^{H\times W\times C}$ and $\mathcal{B}\in\mathbb{R}^{H\times W\times C^\prime}$, respectively. Similar to the supplementary explanation for Remark~\ref{th1}, we first generate a third-order tensor, denoted as $\mathcal{A}^\prime\in\mathbb{R}^{H\times W\times (Ck^2)}$, which encompasses all convolution windows in $\mathcal{A}$. Next, we perform mode-$3$ unfolding and transposition on $\mathcal{A}^\prime$ and $\mathcal{B}$, resulting in matrices $\boldsymbol{A}\in\mathbb{R}^{(HW)\times(Ck^2)}$ and $\boldsymbol{B}\in\mathbb{R}^{(HW)\times C^\prime}$, respectively. With the kernel weights directly represented as $\boldsymbol{X}\in\mathbb{R}^{(Ck^2)\times C^\prime}$, the standard 2D convolution can be mathematically expressed as follows:
\begin{equation}
\label{pe5}
\boldsymbol{A}{\boldsymbol{X}{(\cdot,i)}}=\boldsymbol{B}{(\cdot,i)},\quad i=1, 2, \cdots, C^\prime.
\end{equation}
According to Definition~\ref{d4}, Equation \ref{pe5} can be interpreted as $C^\prime$ non-homogeneous systems. Let $\boldsymbol{\overline{A}}_i=(\boldsymbol{A}|\boldsymbol{B}(\cdot,i))\in\mathbb{R}^{(HW)\times(Ck^2+1)}$, where $i=1,2,\cdots,C^\prime$, be a sequence of augmented matrices. Since $HW$ significantly exceeds $Ck^2+1$, the standard convolution fails to yield the desired output whenever there exists an $i$ such that ${\rm{rank}}(\boldsymbol{A})\neq{\rm{rank}}(\boldsymbol{\overline{A}}_i)$. With the kernel weights directly represented as $\mathcal{X}\in\mathbb{R}^{(Ck^2)\times(HW)\times C^\prime}$, the adaptive 2D convolution can be expressed as follows:
\begin{equation}
\footnotesize
\label{pe6}
\boldsymbol{A}(j,\cdot){\mathcal{X}{(\cdot,j,i)}}=\boldsymbol{B}{(\cdot,i)},\quad i=1, 2, \cdots, C^\prime,\; j=1, 2, \cdots, HW.
\end{equation}
If we consider $\boldsymbol{A}{(j, \cdot)}$ and $\boldsymbol{\overline{A}}_i{(j, \cdot)}$ as row matrices, then ${\rm{rank}}(\boldsymbol{A}{(j, \cdot)})={\rm{rank}}(\boldsymbol{\overline{A}}_i{(j, \cdot)})\leq Ck^2$, which means the adaptive convolution can generate the intended output.

In the second condition, we denote the input data and target output as $\mathcal{C}\in\mathbb{R}^{H\times W\times L\times C}$ and $\mathcal{D}\in\mathbb{R}^{H\times W\times L\times C^\prime}$. We begin by creating a fourth-order tensor, represented as $\mathcal{C}^\prime \in\mathbb{R}^{H\times W\times L\times (Ck^3)}$, that contains all $k\times k\times k$ windows in $\mathcal{C}$. Next, we perform mode-$4$ unfolding and transposition on $\mathcal{C}^\prime$ and $\mathcal{D}$, generating $\boldsymbol{C}\in\mathbb{R}^{(HWL)\times(Ck^3)}$ and $\boldsymbol{D}\in\mathbb{R}^{(HWL)\times C^\prime}$. With the kernel weights defined as $\boldsymbol{W}\in\mathbb{R}^{(Ck^3)\times C^\prime}$, the standard 3D convolution can be expressed as follows:
\begin{equation}
\label{pe7}
\boldsymbol{C}{\boldsymbol{W}{(\cdot,i)}}=\boldsymbol{D}{(\cdot,i)},\quad i=1, 2, \cdots, C^\prime.
\end{equation}
According to Definition~\ref{d4}, Equation \ref{pe7} can be also seen as $C^\prime$ non-homogeneous systems. Let $\boldsymbol{\overline{C}}_i=(\boldsymbol{C}|\boldsymbol{D}(\cdot,i))\in\mathbb{R}^{(HWL)\times(Ck^3+1)}$, where $i=1,2,\cdots,C^\prime$, be a sequence of augmented matrices. Since $HWL$ is much larger than $Ck^3+1$, the standard convolution cannot produce the desired output whenever there exists an $i$ such that ${\rm{rank}}(\boldsymbol{C})\neq{\rm{rank}}(\boldsymbol{\overline{C}}_i)$. With the kernel weights denoted as $\mathcal{W}\in\mathbb{R}^{(Ck^3)\times(HWL)\times C^\prime}$, the adaptive 3D convolution can be expressed as follows:
\begin{equation}
\footnotesize
\label{pe8}
\boldsymbol{C}(j,\cdot){\mathcal{W}{(\cdot,j,i)}}=\boldsymbol{D}{(\cdot,i)},\quad i=1, 2, \cdots, C^\prime,\; j=1, 2, \cdots, HWL.
\end{equation}
If $\boldsymbol{C}{(j, \cdot)}$ and $\boldsymbol{\overline{C}}_i{(j, \cdot)}$ are considered row matrices, then ${\rm{rank}}(\boldsymbol{C}{(j, \cdot)})={\rm{rank}}(\boldsymbol{\overline{C}}_i{(j, \cdot)})\leq Ck^3$, which suggests that the adaptive convolution can produce the desired output.

\begin{table}[t]
	\centering\renewcommand\arraystretch{1.}\setlength{\tabcolsep}{9.7pt}
	\belowrulesep=0pt\aboverulesep=0pt
	\caption{Practical efficiency evaluation of various 3D convolutional paradigms. Our method yields the best performance.}\label{efficency}
	\begin{tabular}{l|cc}
		\toprule
		\textbf{Methods} & Time (ms) & Energy (J) \\  
		\midrule  
		\textbf{Conv3D} & 36.3 & 8.97 \\
		\textbf{DW3D} & 85.6 & 21.0 \\
		\textbf{GPAC3D} & 71.2 & 17.6 \\
		\textbf{D2Conv3D} & 289 & 31.9 \\
		\textbf{ST-A3DNet} & 43.1 & 10.4 \\
		\textbf{The Proposed} & \textbf{17.8} & \textbf{4.42} \\
		\bottomrule
	\end{tabular}
\end{table}

\begin{table}[t]
	\centering\renewcommand\arraystretch{1.}\setlength{\tabcolsep}{12pt}
	\belowrulesep=0pt\aboverulesep=0pt
	\caption{Inference time for different components in the Ada3D block.}\label{bottleneck}
		\begin{tabular}{l|c}
			\toprule
			\textbf{Modules} & Time (ms) \\  
			\midrule  
			\textbf{Spatial Kernel Generator} & 0.34 \\
			\textbf{Spectral Kernel Generator} & 0.37  \\
			\textbf{Combination Section} & 0.03 \\
			\textbf{Bias Generator} & 0.54 \\
			\textbf{Ada3D Operation} & 0.12  \\
			\midrule
			\textbf{Ada3D Block} & 1.40 \\
			\bottomrule
		\end{tabular}
\end{table}

\section*{III. Practical Efficiency Analysis}
To measure the practical efficiency of our method, we first compare Ada3D with other 3D convolutional paradigms in terms of inference time and energy consumption. As shown in Table~\ref{efficency}, the proposed method attains the lowest inference time and energy consumption, demonstrating the superior efficiency of Ada3D. Next, we analyze computational bottlenecks within the Ada3D block by measuring per-component inference time. As summarized in Table~\ref{bottleneck}, the spatial kernel generator, spectral kernel generator, and bias generator dominate the runtime. All tests are run on an Nvidia A100 GPU.

\section*{IV. Network \& Training Settings for HISR}
For the HISR task, we configure the spatial and spectral feature maps with 64 and 32 channels. The values of $k$, $\alpha$, and $\beta$ are set to $3$, $0.375$, and $0.5$, respectively. Additionally, we apply Bicubic for up-sampling. Furthermore, we utilize four ResBlocks to capture spatial details and leverage eight Ada3D blocks to integrate different types of information. All learnable weights are initialized using the Kaiming uniform distribution. During training, we set the number of epochs, batch size, initial learning rate, and ${\lambda}_{ergas}$ to 500, 4, $1\times 10^{-4}$, and $1\times 10^{-4}$. In addition, we use the Adam optimizer and reduce the learning rate by half at the $300$-th epoch.

\vfill

\end{document}